\documentclass[lettersize,journal]{IEEEtran}

\usepackage{algorithm}
\usepackage{algpseudocode}
\usepackage{amsmath, amssymb}
\usepackage{array}
\usepackage[caption=false,font=normalsize,labelfont=sf,textfont=sf]{subfig}
\usepackage{textcomp}
\usepackage{stfloats}
\usepackage{url}
\usepackage{verbatim}
\usepackage{graphicx}
\usepackage{cite}
\usepackage[utf8]{inputenc} 
\usepackage[T1]{fontenc} 
\usepackage[]{hyperref} 
\usepackage{url} 
\usepackage{booktabs} 
\usepackage{amsfonts} 
\usepackage{nicefrac} 
\usepackage{microtype} 
\usepackage{graphicx}
\usepackage{times}
\usepackage{soul}
\usepackage{amsmath}
\usepackage{amsthm}
\usepackage{siunitx}
\usepackage{booktabs}
\usepackage{multirow}
\usepackage{enumitem}
\usepackage{color}
\usepackage{bm}
\usepackage{cite}
\usepackage{bm}
\usepackage{makecell}
\usepackage{amsthm,amsmath,amssymb}
\usepackage{mathrsfs}

\newcommand{\av}[0]{\ensuremath{\boldsymbol{a}} }
\newcommand{\bv}[0]{\ensuremath{\boldsymbol{b}} }

\newcommand{\hv}[0]{\ensuremath{\boldsymbol{h}} }

\newcommand{\kv}[0]{\ensuremath{\boldsymbol{k}} }

\newcommand{\pv}[0]{\ensuremath{\boldsymbol{p}} }

\newcommand{\xv}[0]{\ensuremath{\boldsymbol{x}} }
\newcommand{\yv}[0]{\ensuremath{\boldsymbol{y}} }

\newcommand{\Phiv}[0]{\ensuremath{\boldsymbol{\Phi}} }

\newcommand{\Omegav}[0]{\ensuremath{\boldsymbol{\Omega}} }

\newcommand{\given}{\,|\,}

\theoremstyle{plain}
\newtheorem{theorem}{Theorem}
\newtheorem{definition}{Definition}
\newtheorem{assumption}{Assumption}
\newtheorem{proposition}{Proposition}

\usepackage{amsthm}

\newcommand{\eg}{\textit{e.g.}}
 
\theoremstyle{remark}
\newtheorem{remark}{Remark}

\begin{document}

\title{BaRA: Bayesian Adaptive Rank Allocation for Parameter-Efficient Fine-Tuning}

\author{
Zhibin~Duan, Yuhong~Wang, Jiahong~Fu,
Zongsheng~Yue\\ 
Bo~Chen,~\IEEEmembership{Senior Member,~IEEE,}
Zongben Xu,

\IEEEcompsocitemizethanks{\IEEEcompsocthanksitem Z. Duan, Y. Wang, B. Chen are with the National Key Lab of Radar Signal Processing, Xidian University, Xi'an, Shaanxi 710071, China;  (Corresponding author:  B. Chen)\protect
E-mail: bchen@mail.xidian.edu.cn;
\IEEEcompsocthanksitem J.Fu, Z. Yue, Z. Xu are with School of Mathematics and Statistics, Xi’an Jiaotong University, Xi’an, Shaanxi, 710049, China.} 
\thanks{This paper was produced by the IEEE Publication Technology Group. They are in Piscataway, NJ.}
\thanks{Manuscript received April 19, 2021; revised August 16, 2021.}}
\markboth{Journal of \LaTeX\ Class Files,~Vol.~14, No.~8, August~2021}%
{Shell \MakeLowercase{\textit{et al.}}: A Sample Article Using IEEEtran.cls for IEEE Journals}



\maketitle

\begin{abstract}
While Low-rank adaptation (LoRA) enables highly efficient fine-tuning by constraining task-specific updates to fixed low-rank subspaces, this rigid design limits representational flexibility and often results in overconfident predictions and miscalibrated uncertainty, especially in low-data regimes.
Recent Bayesian LoRA variants improve uncertainty estimation by modeling posterior distributions over adaptation parameters. However, these approaches typically rely on fixed or heuristically determined ranks, overlooking the inherently context-dependent nature of adaptation capacity.
In this paper,  we propose BaRA, a Bayesian Adaptive Rank Allocation framework for parameter-efficient fine-tuning. 
Drawing inspiration from probabilistic topic models, BaRA dynamically allocates adaptation capacity by activating a sparse, context-dependent subset of disentangled latent factors, enabling instance-wise variation in effective rank.
This Bayesian formulation provides principled, data-driven capacity control, mitigating over-parameterization while preserving expressiveness.
Beyond the modeling contribution, we provide a complexity-theoretic generalization analysis showing that the generalization gap of BaRA depends on the learned joint effective rank $\bar{s}_{\Phi,\theta}$ induced by the global-local gate, rather than the maximum rank $r$. This result explains why sparse adaptive rank allocation can reduce the effective hypothesis complexity while preserving input-dependent expressiveness. 
Extensive experiments on diverse natural language benchmarks demonstrate that BaRA consistently improves predictive performance, robustness, and uncertainty calibration compared to standard LoRA and existing Bayesian LoRA variants.
\end{abstract}

\begin{IEEEkeywords}
Low-Rank Adaptation, Parameter-Effective Fine-Tuning, Sparsity-aware Bayesian Model, Variational Inference, PAC-Bayesian Generlization Bound
\end{IEEEkeywords}

\section{Introduction}
Large language models (LLMs) have achieved strong performance across a wide range of natural language tasks \cite{brown2020language}, yet their massive parameter sizes render full fine-tuning prohibitively expensive for many downstream applications \cite{kaplan2020scaling}.
Parameter-efficient fine-tuning (PEFT) addresses this challenge by adapting pretrained models through updating only a small subset of parameters \cite{houlsby2019parameter, han2024parameter}.
Among PEFT methods, low-rank adaptation (LoRA) \cite{hu2022lora} is particularly effective and widely adopted.
LoRA is motivated by the observation that task-specific weight updates often lie in a low-dimensional subspace \cite{li2018measuring, aghajanyan2021intrinsic}, and parameterizes these updates using low-rank matrices.
This design substantially reduces the number of trainable parameters and associated memory overhead while preserving competitive performance. 

Despite its efficiency, LoRA-based fine-tuning with limited data often produces overconfident and poorly calibrated predictions, a phenomenon closely linked to hallucination and entropy collapse in LLMs \cite{guo2017calibration, yin2023large, xiong2023can, kim2024knowledge}.
This issue has motivated Bayesian formulations of LoRA that explicitly model uncertainty over low-rank adaptation parameters to improve generalization and predictive calibration \cite{yang2024bayesian, liu2025uncertainty}.
Existing Bayesian LoRA methods largely inherit techniques from Bayesian neural networks \cite{wang2016towards}, including computationally intensive ensemble strategies \cite{lakshminarayanan2017simple, wang2023ensemble,yang2023bayesian} and post-hoc approximations such as Laplace inference.
More recently, mean-field variational inference has emerged as a scalable alternative, jointly optimizing posterior means and variances of LoRA parameters to enable principled and tractable uncertainty-aware fine-tuning \cite{wang2024blob, samplawski2025scalable, kumar2025latent}.

Nevertheless, Bayesian LoRA methods inherit a structural limitation of standard LoRA: the adaptation rank is fixed a priori and uniformly applied in layers and input contexts \cite{hu2022lora}.
This assumption contradicts empirical evidence of pronounced layer-wise heterogeneity in transferability, where task-relevant adaptations are often concentrated in a small subset of layers \cite{yosinski2014transferable, pan2024lisa, son2025not}.
Furthermore, adaptation requirements can vary significantly across input contexts, yet existing methods lack mechanisms to dynamically modulate the effective rank in a context-dependent manner \cite{liu2023deja}.
Although recent approaches introduce layer-wise rank allocation techniques \cite{zhang2023adalora, ding2023sparse}, rank remains a deterministic architectural choice rather than a stochastic, input-adaptive variable.
As a result, overly large ranks can degrade generalization \cite{kumar2022fine} and exacerbate overconfidence \cite{kotha2023understanding}, whereas insufficient ranks restrict expressiveness and induce representational bottlenecks \cite{zhang2023adalora}.

To overcome these limitations, we revisit sparsity-aware Bayesian models \cite{wipf2004sparse, wipf2011latent}, which induce structured sparsity through principled prior design and naturally support automatic capacity determination.
Probabilistic topic models (PTMs), such as Latent Dirichlet Allocation and Poisson factor analysis \cite{blei2003latent, zhou2012beta}, exemplify this paradigm by disentangling sparsity across hierarchical levels. 
Specifically, local latent variables capture input-dependent sparse activations at the context level, while shared latent variables equipped with sparsity-promoting priors yield compact and parsimonious representations at the global level.
This hierarchical Bayesian structure provides a theoretically grounded mechanism for adaptive capacity allocation via sparse and context-dependent latent activation.

Motivated by these insights, we propose Bayesian Adaptive Rank Allocation (BaRA), a sparse Bayesian PEFT framework that dynamically allocates low-rank adaptation capacity in both a context-dependent and layer-wise manner.
BaRA introduces context-aware local latent variables and layer-specific global latent variables with sparsity-inducing priors to jointly model context-adaptive and layer-adaptive ranks.
Local latent variables selectively activate the adaptation capacity conditioned on input, capturing context-dependent uncertainty, while global latent variables encode uncertainty over adaptation parameters and promote structured sparsity across layers.
This hierarchical Bayesian formulation enables a unified and principled modeling of both the data uncertainty and the model uncertainty within a single probabilistic framework \cite{kendall2017uncertainties}.
To ensure scalability to modern LLMs, we further develop an amortized variational inference network that tightly integrates probabilistic latent-variable modeling with large-scale pretrained architectures \cite{kingma2013auto}.
In addition to the intuitive modeling motivation, we further analyze the proposed model through complexity-based generalization bounds, which theoretically verifies its effectiveness in reducing model complexity and improving generalization.
Our contributions are summarized as follows:

\begin{itemize}
\setlength{\itemsep}{3pt}

\item \textbf{Bayesian adaptive rank modeling for PEFT.}
We introduce BaRA, a principled sparse Bayesian extension of LoRA that enables automatic and context-dependent rank allocation through hierarchical prior design.

\item \textbf{Scalable variational inference.}
We develop efficient variational inference schemes that scale to modern LLMs while preserving Bayesian uncertainty over adaptive low-rank parameters.

\item \textbf{Complexity-theoretic generalization analysis.} We provide a complexity-theoretic generalization analysis showing that BaRA's generalization gap depends on the learned joint effective rank $\bar{s}_{\Phi,\theta}$ rather than the maximum rank $r$, thereby explaining its advantage over fixed-rank LoRA.

\item \textbf{Sparsity-driven generalization and calibration.}
By integrating structured sparsity into Bayesian low-rank adaptation, BaRA improves uncertainty calibration, robustness, and generalization under limited data.

\item \textbf{Comprehensive empirical evaluation.}
Extensive experiments on standard LLM fine-tuning benchmarks demonstrate that BaRA consistently outperforms deterministic and existing Bayesian LoRA methods in accuracy, calibration, and robustness.

\end{itemize}

\section{Related Work}

\subsection{Parameter-Efﬁcient Fine-Tuning Methods.}
Low-Rank Adaptation (LoRA) \cite{hu2022lora} is a widely adopted parameter-efficient fine-tuning method for large pretrained models.
By inserting trainable low-rank adapters into frozen weight matrices, LoRA enables efficient adaptation of billion-parameter LLMs with minimal computational and memory overhead.
Following its success, numerous extensions have been proposed to enhance parameter efficiency and flexibility, including structured adapter designs \cite{kopiczko2023vera, he2022sparseadapter, zhang2023lora} and dynamic adaptation strategies \cite{valipour2023dylora, lialin2023relora}.
A prominent line of work focuses on adaptive rank allocation to mitigate the limitations of fixed-rank LoRA.
Methods such as AdaLoRA \cite{zhang2023adalora}, and AutoLoRA \cite{zhang2024autolora} adjust the rank across layers or training stages to better balance expressiveness and efficiency.
Other approaches incorporate sparsity into LoRA modules, including SparseLoRA \cite{ding2023sparse} and RoseLoRA \cite{wang2024roselora}, to reduce redundancy and improve generalization.
Despite these advances, existing rank-adaptive methods remain largely deterministic and static, and do not condition rank allocation on individual input contexts.

\subsection{Bayesian Low-Rank Adaptation.} 
Bayesian learning offers a principled framework for uncertainty quantification by placing prior distributions over model parameters and performing posterior inference \cite{gal2016dropout}.
Recent work has explored Bayesian formulations of LoRA to improve calibration and robustness in fine-tuning large models.
Early approaches often incur substantial computational overhead: SWAG-LoRA \cite{onal2024gaussian} requires additional passes to estimate posterior statistics, LoRA ensembles \cite{wang2023lora} rely on training multiple instances, and Laplace-based methods such as LaplaceLoRA \cite{yang2023bayesian} perform post-hoc posterior approximation using second-order information.
To improve scalability, more recent methods restrict Bayesian inference to the low-rank adaptation subspace.
For example, Yang et al. \cite{yang2023bayesian} apply Kronecker-factored Laplace approximations to LoRA parameters; BLoB \cite{NEURIPS2024_7d535754} jointly estimates posterior means and covariances within a single fine-tuning stage, and ScalaBL \cite{samplawski2025scalable} employs stochastic variational inference in a low-dimensional LoRA subspace.
In addition, Rahmati et al. \cite{rahmati2025c} introduce contextualized Bayesian LoRA modules to adapt uncertainty estimates across inputs.
While these Bayesian LoRA methods substantially improve uncertainty estimation, they generally assume a fixed or pre-specified adaptation rank and focus on modeling parameter uncertainty rather than context-dependent capacity allocation.
In contrast, our work addresses this limitation by introducing a Bayesian framework that dynamically allocates adaptation rank conditioned on the input context

\begin{figure*}[t]
\centering
\subfloat[Low-Rank Adaptation (LoRA)]{
\includegraphics[width=0.32\linewidth]{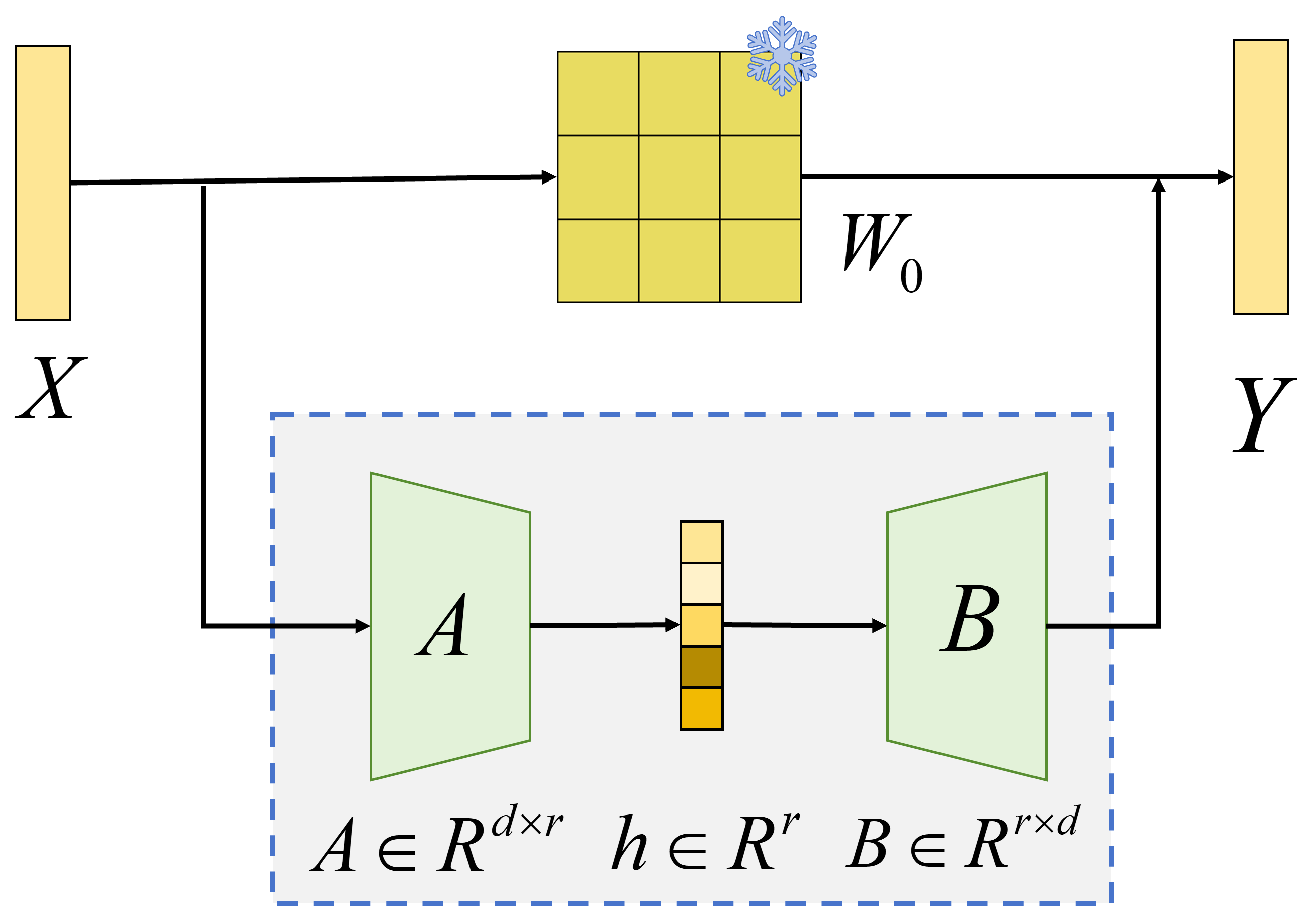}
\label{fig_lora_model}
}\quad\quad\quad
\subfloat[Bayesian Adaptive Rank Allocation(BaRA)]{
\includegraphics[width=0.40\linewidth]{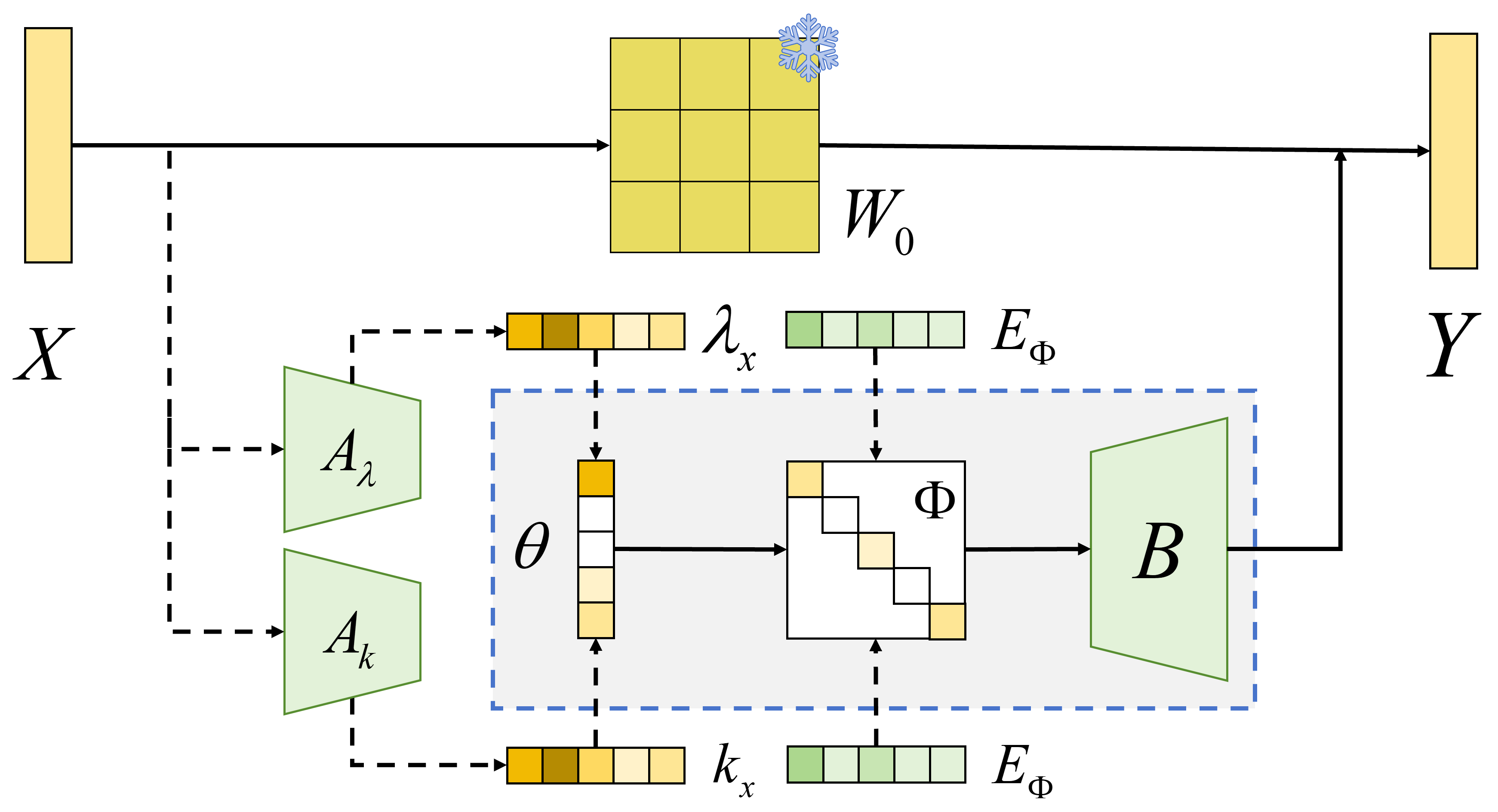}
\label{fig_model_bara}
}
\caption{Illustration of the standard LoRA (left) and the proposed BaRA (right). Green blocks represent newly introduced trainable parameters, dashed lines denote the variational inference process, and solid lines indicate the generative process.}
\label{fig_app: uncertainty}
\end{figure*} 

\section{Preliminaries} 
In this section, we briefly review the standard LoRA framework and its Bayesian extensions, which serve as the foundation of our method.

\subsection{Low-Rank Adaptation }
LoRA \cite{hu2022lora} is motivated by the empirical observation that task-specific parameter updates often lie in a low intrinsic-dimensional subspace \cite{houlsby2019parameter, han2024parameter}. 
Rather than fine-tuning the full parameters, LoRA freezes the backbone weights and parameterizes the update via a learnable low-rank decomposition, thereby achieving parameter-efficient adaptation.
Formally, as shown in Fig.~\ref{fig_lora_model}, given an input vector $\xv \in \mathbb{R}^{d}$ and a frozen weight matrix ${W}_0 \in \mathbb{R}^{d \times d}$, the adapted forward pass is defined as
\begin{equation}
\label{eq_lora}
\yv = ({W}_0 + \Delta {W}) \xv
= ({W}_0 + {B}{A}) \xv,
\end{equation}
where ${B} \in \mathbb{R}^{d \times r}$ and ${A} \in \mathbb{R}^{r \times d}$ are trainable matrices with rank $r \ll d$. 
This factorization reduces the number of trainable parameters from $d^2$ to $2rd$, significantly lowering memory usage and computational cost while maintaining competitive performance to full fine-tuning. 
More generally, the low-rank update can be expressed as
\begin{equation}
\label{eq_lora_e}
\Delta {W} = {B}{E}{A},
\end{equation}
where ${E} \in \mathbb{R}^{r \times r}$ is a diagonal scaling matrix that modulates the contribution of each rank component.
This formulation provides a convenient abstraction for analyzing and extending low-rank adaptation mechanisms \cite{zhang2023adalora}.

\subsection{Bayesian LoRA}
\label{sec_bayesian_lora}

Let $\mathcal{D} = \{(x_i, y_i)\}_{i=1}^N$ denote a dataset of $N$ i.i.d.\ input--output pairs. Under the Bayesian paradigm, model parameters are treated as random variables, and prediction marginalizes over their posterior distribution:
\begin{equation}
p(\yv \mid \xv, \mathcal{D})
= \int p(\yv \mid \xv, {W}) \, p({W} \mid \mathcal{D}) \, \mathrm{d}{W}.
\end{equation}
Leveraging the low-rank assumption, the Bayesian prediction further becomes
\begin{equation}
\label{eq_lora_bayesian}
p(\yv \mid \xv, \mathcal{D})
=
\int p(\yv \mid \xv, {W}_0, {A}, {B})
\, p({A}, {B} \mid \mathcal{D})
\, \mathrm{d}{A} \, \mathrm{d}{B}.  
\end{equation}
Within this framework, various Bayesian LoRA methods differ mainly in which components of the low-rank decomposition are treated as random variables.

 \textbf{1. BLoB \cite{wang2024blob}.}
    BLoB places a variational posterior over one low-rank factor (typically ${A}$), while keeping the other factor deterministic:
    \begin{equation}
    q({A}) = \mathcal{N}({A} \mid \boldsymbol{\mu}_{{A}}, \boldsymbol{\Omega}_{{A}}).
    \end{equation}
    This formulation captures epistemic uncertainty in a restricted subspace but leaves the effective rank fixed.

 \textbf{2. ScalaBL \cite{samplawski2025scalable}.}
    Most recent methods generalize the update as $\Delta {W} = {B}{E}{A}$ by introducing a scaling matrix ${E}$ and performing inference over ${E}$:
    \begin{equation}
    q({E}) = \mathcal{N}({E} \mid \boldsymbol{\mu}_{{E}}, \boldsymbol{\Omega}_{{E}}),
    \end{equation}
    as explored in scalable Bayesian LoRA methods \cite{samplawski2025scalable}.

\subsection{Limitations of Bayesian LoRA Methods.}
Despite their advances in uncertainty-aware fine-tuning, these Bayesian LoRA methods still suffer from two fundamental limitations:
$i)$ \textbf{Incomplete uncertainty Modeling.}
Current methods primarily model \emph{epistemic uncertainty} by placing distributions over adaptation parameters, while treating latent representations deterministically \cite{wang2024blob, samplawski2025scalable}. 
As a result, \emph{aleatoric uncertainty}, which arises from intrinsic noise, ambiguity, or label variability of the data, is not explicitly captured \cite{rahmati2025c}.
This prevents a unified modeling of uncertainty across parameter and representation levels,
which is essential for reliable decision-making in deep models \cite{kendall2017uncertainties}.
$ii)$ \textbf{Fixed-rank adaptation.}
More importantly, existing Bayesian LoRA methods assume a fixed or pre-specified low-rank structure. 
Even though parameter uncertainty is introduced, the rank remains dense and entangled, limiting the model's ability to adaptively allocate capacity in a data- or context-dependent manner.
While adaptive-ranking approaches such as AdaLoRA \cite{zhang2023adalora, ding2023sparse} partially mitigate this issue via dynamical rank adjustment, they lack a principled formulation and do not provide uncertainty-aware rank inference.

\section{Bayesian Adaptive Rank Allocation}
\label{sec:bnlora}
This section presents BaRA, a sparse Bayesian low-rank adaptation framework that unifies uncertainty modeling and adaptive capacity allocation. We first present the generative model of BaRA (Sec.~\ref{sec_cara}), followed by an efficient amortized variational inference algorithm (Sec.~\ref{sec_avi}). Then, we discuss the differences between BaRA and standard LoRA (Sec.~\ref{Bara_Lora}).
\subsection{Generative Model of BaRA}
\label{sec_cara}

To introduce BaRA, we first rewrite LoRA in an equivalent latent-variable form:
\begin{equation}
\yv
= W_0 \xv + B E A \xv
= W_0 \xv + B E \hv,
\qquad
\hv = A \xv,
\end{equation}
where $\hv$ denotes the low-dimensional latent representation induced by the frozen projection matrix $A$. In this view, the low-rank adaptation is no longer treated as a deterministic matrix product, but as a structured latent transformation operating in a compact subspace.

From a Bayesian perspective, the predictive distribution can be written as
\begin{equation}
\begin{split}
p(\yv & \mid \xv, \mathcal{D})
= \\
&\int p(\yv \mid \xv, W_0, \hv, E, B)\,
p(\hv, E, B \mid \mathcal{D})
\, \mathrm{d}\hv \, \mathrm{d}E \, \mathrm{d}B,
\end{split}
\end{equation}
which explicitly marginalizes over both latent representations and adaptation parameters. This formulation highlights that uncertainty in prediction arises not only from the observation model, but also from the uncertainty of the adapted low-rank parameters and the context-dependent latent features.

To jointly capture \emph{aleatoric uncertainty}, which stems from data ambiguity and input noise, and \emph{epistemic uncertainty}, which reflects uncertainty in the learned adaptation mechanism \cite{kendall2017uncertainties}, we treat both the latent representation $\hv$ and the adaptation weights as stochastic variables. Specifically, we parameterize the local latent variable by $\boldsymbol{\theta}$ and the global adaptation factors by $\Phi$. This design is not only probabilistically principled, but also computationally efficient, because all stochasticity is confined to a low-dimensional adaptation subspace rather than the full model parameter space \cite{samplawski2025scalable}.

As illustrated in Fig.~\ref{fig_model_bara}, we model the low-rank update through a hierarchical generative process:
\begin{equation}
\yv = W_0 \xv + B \Phi \boldsymbol{\theta},
\end{equation}
where $\Phi$ and $\boldsymbol{\theta}$ jointly define a structured factorization of the adaptation space. In this hierarchy, $\Phi$ captures layer-wise global basis factors shared across the dataset, while $\boldsymbol{\theta}$ represents input-specific activations over these basis factors. Intuitively, $\Phi$ defines \emph{what can be adapted}, whereas $\boldsymbol{\theta}$ determines \emph{how much each factor is activated} for a given input.

To encourage automatic rank selection, we impose sparsity-inducing Gamma priors on the latent variables:
\begin{equation}
\Phi \sim \mathrm{Gamma}(\alpha_\Phi, \beta_\Phi),
\qquad
\boldsymbol{\theta} \sim \mathrm{Gamma}(\alpha_\theta, \beta_\theta).
\end{equation}
By choosing shape parameters smaller than one, the Gamma prior places substantial mass near zero, thereby promoting shrinkage of unnecessary factors and yielding a sparse posterior representation. In our setting, this mechanism enables the model to activate only a subset of latent factors for each input, leading to data-dependent rank allocation without manual tuning.

The resulting predictive distribution marginalizes over both parameter-level and representation-level uncertainty:
\begin{equation}
\begin{split}
p(\yv \mid \xv, \mathcal{D})
=
\int p(\yv \mid W_0, \Phi, \boldsymbol{\theta})\,
p(\Phi \mid \mathcal{D})\,
p(\boldsymbol{\theta} \mid \xv)
\, \mathrm{d}\Phi \, \mathrm{d}\boldsymbol{\theta}.
\end{split}
\end{equation}
This Bayesian marginalization provides two complementary benefits: it improves predictive robustness by averaging over plausible adaptation configurations, and it yields an interpretable decomposition of uncertainty into dataset-level structural uncertainty and instance-level activation uncertainty.

\begin{remark}
The proposed hierarchical construction enables multi-level adaptivity:
\begin{enumerate}
\item \textbf{Context-wise adaptive rank.} The local latent variable $\boldsymbol{\theta}$ induces input- or token-dependent activation over adaptation factors, so the effective rank can vary across contexts.
\item \textbf{Layer-wise adaptive rank.} The global latent variable $\Phi$ defines a layer-specific dictionary of adaptation factors, allowing different layers to exhibit different sparsity patterns and rank capacities.
\end{enumerate}
Together, these two levels of stochasticity endow parameter-efficient fine-tuning with principled uncertainty quantification, sparse factor selection, and flexible rank adaptation.
\end{remark}

\subsection{ Variational Inference Network}
\label{sec_avi}

Exact posterior inference over the latent variables $(\boldsymbol{\theta}, \Phi)$ is intractable, primarily due to the nonlinear dependence between the latent factors and the observed data, as well as the hierarchical coupling induced by the global--local structure. To enable scalable learning, we adopt a mean-field variational approximation of the form
\begin{equation}
q(\boldsymbol{\theta}, \Phi)
=
q(\Phi)\prod_i q(\boldsymbol{\theta}_i),
\end{equation}
where $\Phi$ denotes the global layer-wise latent factors shared across all instances, and $\boldsymbol{\theta}_i$ represents the instance-specific latent activations associated with the $i$-th input. This factorization decouples global structural learning from sample-specific inference, thereby making optimization tractable in large-scale settings.

Although the Gamma distribution is a natural prior for modeling nonnegative and sparse latent variables, direct reparameterization of Gamma random variables typically leads to high-variance gradient estimators and unstable optimization \cite{ruiz2016generalized, naesseth2017reparameterization, zhang2020deep}. To address this issue, we follow the Weibull approximation strategy in \cite{zhang2020deep} and use a Weibull variational family to approximate the Gamma posterior. This choice is particularly appealing because the Weibull distribution shares the same positive support as the Gamma distribution, admits efficient pathwise reparameterization, and can closely match the shape of sparse nonnegative posteriors. In particular, the Weibull shape parameter controls the tail behavior and sparsity level of the posterior, making it well suited for adaptive rank learning.

\noindent\textbf{Local Variables.}
For each instance $i$, we employ amortized inference to parameterize the variational posterior of the local latent variable $\boldsymbol{\theta}_i$ conditioned on the observed input $\xv_i$:
\begin{equation}
\begin{split}
&q(\boldsymbol{\theta}_i \mid \xv_i)
= \mathrm{Weibull}(\kv_i, \boldsymbol{\lambda}_i),\\
&\kv_i
= \mathrm{Softplus}({A_k}\xv_i), \\
&\boldsymbol{\lambda}_i
= \mathrm{Softplus}({A_\lambda}\xv_i) / \exp(1 + 1/\kv_i),
\end{split}
\label{eq_local_vi}
\end{equation}
where $A_k, A_\lambda \in \mathbb{R}^{D \times r}$ are learnable amortization parameters. The Softplus transformation guarantees positivity of the variational parameters while remaining smooth and fully differentiable, which is crucial for stable gradient-based optimization. The normalization term $\exp(1 + 1/\kv_i)$ follows the Weibull parameterization used to align its moments with those of the target Gamma family \cite{zhang2020deep}. Intuitively, this construction allows the inference network to map each input $\xv_i$ to a sparse nonnegative posterior over latent activations, thereby enabling instance-adaptive factor selection.

\noindent\textbf{Global Variables.}
The global latent variable $\Phi$ is modeled in an analogous manner:
\begin{equation}
\begin{split}
&q(\Phi)
= \mathrm{Weibull}(\kv_{\Phi}, \boldsymbol{\lambda}_{\Phi}),\\
&\kv_{\Phi}
= \mathrm{Softplus}(E_k), \\
&\boldsymbol{\lambda}_{\Phi}
= \mathrm{Softplus}(E_\lambda) / \exp(1 + 1/\kv_{\Phi}),
\end{split}
\label{eq_global_vi}
\end{equation}
where $E_k, E_\lambda \in \mathbb{R}^{r}$ are free variational parameters optimized directly during training. In contrast to the local variables, $\Phi$ is shared across all data instances and therefore captures layer-wise structural regularities, such as common sparsity patterns and global rank allocation tendencies. This separation between local adaptivity and global structural learning provides a compact yet expressive posterior approximation.

From a Bayesian perspective, the local posteriors $\{q(\boldsymbol{\theta}_i \mid \xv_i)\}$ account for sample-specific uncertainty and activation strength, whereas $q(\Phi)$ summarizes the dataset-level inductive bias. Jointly optimizing these two components yields a scalable variational inference scheme that can flexibly balance data fit and structural regularization.

\subsection{Amortized Variational Inference}

Let $\Omegav$ denote all trainable parameters spanning both the generative model and the inference networks within BaRA framework. The marginal likelihood of the full dataset $(X, Y)$ is defined by integrating out the model parameters:
\begin{equation}
p \left( { {Y} \given {X} } \right) = \int \int {  {p\left( {Y \: | \: \boldsymbol{\theta}, \Phiv} \right)} }
{ { { {p\left( {\boldsymbol{\theta}_i} \right)} }}}\pv(\Phiv) d\boldsymbol{\theta} d \Phiv .
\end{equation}
Direct computation of this high-dimensional integral is generally intractable. Following the variational inference (VI) paradigm, we instead maximize the Evidence Lower Bound (ELBO) on the log-likelihood, $\log \pv(Y \given X)$.
Generalizing the summation over individual data points into expectations over the full dataset $X$, the ELBO $\mathcal{L}(Y \given X)$ is given by:
\begin{equation} \label{eq: BNDL_ELBO}
\begin{split}
\mathcal{L}(Y\given X) &= {\mathbb{E}_{q(\boldsymbol{\theta} \given X)q(\bm{\Phi})}}\left[ {\ln p\left({Y} \: | \: \boldsymbol{\theta}, \Phiv \right)} \right] \\
&-{{\mathbb{E}_{q(\boldsymbol{\theta} \given X)}}\left[ {\ln \frac{{q\left(\boldsymbol{\theta} \: | \:  X \right)}}{{p\left(\boldsymbol{\theta}\right) }}} \right]}  - {{\mathbb{E}_{q(\Phiv)}}\left[ {\ln \frac{{q\left(\bm{\Phi} \right)}}{{p\left(\bm{\Phi} \right) }}} \right]} .
\end{split}
\end{equation}
The ELBO is composed of three distinct terms: $i)$ The first term represents the expected log-likelihood (reconstruction term). It drives the generative model to accurately fit the data. $ii)$  The second and third terms are Kullback–Leibler (KL) divergences. 
These act as regularizers, ensuring that the variational posterior distributions, $q(\boldsymbol{\theta} \given X)$ and $q(\bm{\Phi})$, remain close to their respective prior distributions, $p(\boldsymbol{\theta})$ and $p(\bm{\Phi})$.
The KL regularizers constrain the variational posteriors to remain close to the sparse Gamma priors. This Bayesian regularization promotes automatic shrinkage of redundant components, enabling adaptive rank allocation and mitigating over-parameterization. As a result, the model can retain only the latent factors that are strongly supported by the data, while suppressing unnecessary or weakly activated dimensions.
The complete set of BNDL parameters, $\Omegav$, is directly optimized by maximizing $\mathcal{L}(Y \given X)$ using standard gradient-based methods, such as stochastic gradient descent (SGD) \cite{DBLP:journals/corr/KingmaB14}.


\begin{algorithm*}
\caption{BaRA Training Procedure}
\begin{algorithmic}[1]
\Require Pretrained weights $\mathbf{W}_0\in\mathbb{R}^{n\times d}$, fine-tuning dataset $\mathcal{D}$, prior distribution $P(\boldsymbol{\theta})$, $P(\boldsymbol{\Phi})$
\Require Number of training epochs $E$, batch size $B$, learning rate $\eta$
\Require KL divergence weights $\beta_{\theta}$, $\beta_{\Phi}$
\State $\mathbf{A}\leftarrow\text{Orthogonal}(\mathbb{R}^{d\times 2r})$ 
\Comment{Initialize down-projection}
\State $\mathbf{B}\leftarrow\mathbf{0}$ \Comment{Initialize as in LoRA}
\State $\boldsymbol{\Phi}\leftarrow\mathbf{0}\in\mathbb{R}^{2r}$ \Comment{Initialize global Weibull parameters}
\For{epoch $e\gets 1\ldots E$}
    \For{batch $\mathcal{D}_t\sim\mathcal{D}$} 
        \State $\mathbf{u}_{\theta}\sim\mathcal{U}(0,1)$, $\mathbf{u}_{\Phi}\sim\mathcal{U}(0,1)$  \Comment{Sample uniform noise}
        \State $[\mathbf{k}_{\theta},\boldsymbol{\lambda}_{\theta}]\leftarrow\text{softplus}(\mathbf{A}(\mathcal{D}_t))$ \Comment{Input-dependent Weibull parameters}
        \State $[\mathbf{k}_{\Phi},\boldsymbol{\lambda}_{\Phi}]\leftarrow\text{softplus}(\boldsymbol{\Phi})$ \Comment{Global Weibull parameters}
        \State $\boldsymbol{\theta}_t\leftarrow\boldsymbol{\lambda}_{\theta}\odot(-\log(1-\mathbf{u}_{\theta}))^{\mathbf{k}_{\theta}}$ \Comment{Reparameterization trick}
        \State $\boldsymbol{\Phi}_t\leftarrow\boldsymbol{\lambda}_{\Phi}\odot(-\log(1-\mathbf{u}_{\Phi}))^{\mathbf{k}_{\Phi}}$ \Comment{Reparameterization trick}
        \State $\mathbf{W}_t\leftarrow\mathbf{W}_0+\mathbf{B}(\boldsymbol{\theta}_t\odot\boldsymbol{\Phi}_t)$
        \State $\mathcal{L}_t\leftarrow-\frac{1}{B}\log P(\mathcal{D}_t|\mathbf{W}_t)+\beta_{\theta}D_{\mathrm{KL}}(q(\boldsymbol{\theta})\|P(\boldsymbol{\theta}))+\beta_{\Phi}D_{\mathrm{KL}}(q(\boldsymbol{\Phi})\|P(\boldsymbol{\Phi}))$ 
        \Comment{Compute ELBO}
        \State $\boldsymbol{\psi}\leftarrow[\mathbf{A},\boldsymbol{\Phi},\mathbf{B}]$ \Comment{Collect trainable parameters}
        \State $\boldsymbol{\psi}\leftarrow\boldsymbol{\psi}-\eta\frac{\partial\mathcal{L}_t}{\partial\boldsymbol{\psi}}$ \Comment{Compute gradient and update parameters}
    \EndFor
\EndFor
\end{algorithmic}
\label{bara_Algor}
\end{algorithm*}

Algorithm \ref{bara_Algor} details the training pipeline of BaRA. This method takes the fine-tuning dataset $\mathcal{D}$ as input and adopts a variational inference framework with two Weibull-distributed latent variables: the input-dependent posterior $q(\boldsymbol{\theta})$ and the global posterior $q({\Phi})$. Specifically, the input is first projected through the linear down-projection $\mathbf{A} \in \mathbb{R}^{d \times 2r}$ to a $2r$-dimensional space, encoding the shape parameter $\mathbf{k}_{\theta}$ and scale parameter $\boldsymbol{\lambda}_{\theta}$ of the Weibull distribution, respectively. The softplus activation function ensures the positivity of these parameters. The global latent variable ${\Phi} \in \mathbb{R}^{2r}$ is directly parameterized as a learnable vector, where the first $r$ dimensions encode $\mathbf{k}_{\Phi}$ and the remaining $r$ dimensions encode $\boldsymbol{\lambda}_{\Phi}$. Both latent variables are sampled via the Weibull reparameterization trick using uniform noise $\mathbf{u} \sim \mathcal{U}(0,1)$, thus enabling gradient-based optimization. The sampled latent variables are combined through element-wise multiplication and then passed to the up-projection $\mathbf{B}$ to form the low-rank weight adaptation. The training objective is the evidence lower bound (ELBO), which consists of the negative log-likelihood and two KL divergence terms; the latter regularize $q(\boldsymbol{\theta})$ and $q({\Phi})$ towards their respective Gamma priors $P(\boldsymbol{\theta})$ and $P({\Phi})$.

\subsection{Complexity Analysis} \label{subsec:complexity}
Let $d$ denote the hidden dimension and $r$ the maximum rank.  
BaRA retains the same order of forward computational complexity as LoRA, $\mathcal{O}(dr)$, since only a sparse subset of latent factors is activated per input. In practice, the \emph{effective} rank is often much smaller than $r$ due to the sparsity prior, leading to reduced average computational cost during inference.  
In terms of parameter complexity, BaRA introduces additional variational parameters for ${\Phi}$ and the inference network, resulting in $\mathcal{O}(dr + r)$ parameters, which remains linear in $r$ and comparable to existing Bayesian LoRA variants. Importantly, BaRA provides adaptive rank selection and uncertainty quantification without incurring the quadratic or ensemble-level costs of alternative Bayesian fine-tuning approaches.

\section{Theory Analysis}
In contrast to prior intuition-driven studies, our framework enables the principled derivation of theoretical guarantees on generalization. 
These guarantees are derived directly from the probabilistic structure of BaRA, thereby providing rigorous support for its empirical performance.

\subsection{Connections and Differences Between BaRA and LoRA}
\label{Bara_Lora}

From Eqs.~\ref{eq_local_vi} and \ref{eq_global_vi}, the variational posteriors satisfy
${\mathbb{E}}[\boldsymbol{\theta}_i] = \mathrm{Softplus}(A_\lambda \xv_i)$ and
${\mathbb{E}}[\boldsymbol{\Phi}] = \mathrm{Softplus}(E_\lambda)$.
Replacing the stochastic latent variables $\boldsymbol{\theta}_j$ and $\boldsymbol{\Phi}$ with their expectations therefore yields a deterministic mapping equivalent to a LoRA-style update:
\begin{equation}
\label{eq_nonlora}
\yv
= {W}_0 \xv
+ {B}\,\mathrm{Softplus}(E_\lambda)\,
\mathrm{Softplus}(A_\lambda \xv).
\end{equation}
Equivalently, as the Weibull shape parameter $\kv_{\cdot} \rightarrow \infty$, the posterior variance vanishes and the distribution collapses to a point mass at its mean.
In this limit, the proposed stochastic adaptation layer degenerates to a deterministic \emph{non-negative LoRA} (NonLoRA) formulation.
Importantly, BaRA constitutes a general modeling framework rather than a Softplus-specific design.
Alternative activation or gating mechanisms that induce sparsity, such as $\mathrm{ReLU}(\cdot)$ or $\mathrm{Top}\text{-}k(\cdot)$, can be readily incorporated to enable adaptive rank allocation under different inductive biases.

The essential distinction between NonLoRA and standard LoRA lies in its \emph{input-dependent nonlinear adaptation} mechanism.
Standard LoRA applies a linear update within a fixed, globally shared low-rank subspace.
In contrast, NonLoRA induces data-dependent activation over latent factors, yielding
\begin{equation}
\Delta{W} \xv
= {B}_{\mathcal{I}(\xv)} \,
{E}_{\mathcal{I}(\xv)}(\xv),
\end{equation}
where $\mathcal{I}(\xv)$ denotes the index set of latent dimensions activated by the input $\xv$.
As $\mathcal{I}(\xv)$ varies across inputs, NonLoRA effectively implements a mixture of low-rank projections, dynamically routing different inputs to distinct adaptation subspaces.
This mechanism substantially enhances representational expressivity relative to LoRA, while preserving sparsity, non-negativity, and parameter efficiency.

\subsection{Generalization Analysis via Joint Global--Local Complexity}
\label{subsec:joint_complexity_main}

We provide a complexity-theoretic analysis to explain why the proposed BaRA framework improves generalization. 
The key insight is that BaRA does not use all rank components uniformly. 
Instead, the global latent variable $\boldsymbol{\Phi}$ and the local latent variable $\boldsymbol{\theta}(\xv)$ jointly induce an input-dependent sparse rank support through the multiplicative gate
\begin{equation}
\boldsymbol{g}(\xv)
=
\boldsymbol{\Phi}\boldsymbol{\theta}(\xv).
\end{equation}
Therefore, the effective complexity of BaRA is governed by the number of activated rank components rather than the pre-specified maximum rank $r$.

\begin{definition}[BaRA hypothesis class]
\label{def:bara_hypothesis}
Consider a BaRA-adapted linear layer with frozen weight $W_0$ and low-rank adaptation
\begin{equation}
\Delta W(\xv)
=
B
\operatorname{diag}
\left(
\boldsymbol{\Phi}\odot\boldsymbol{\theta}(\xv)
\right)
A,
\end{equation}
where $A\in\mathbb{R}^{r\times d_{\mathrm{in}}}$, 
$B\in\mathbb{R}^{d_{\mathrm{out}}\times r}$, and $r$ is the maximum rank.
The corresponding BaRA hypothesis class is defined as
\begin{equation}
\mathcal{H}_{\Phi,\theta}
=
\left\{
\xv\mapsto
B
\operatorname{diag}
\left(
\boldsymbol{\Phi}\odot\boldsymbol{\theta}(\xv)
\right)
A\xv
\right\}.
\end{equation}
\end{definition}

\begin{definition}[Joint global--local effective rank]
\label{def:joint_effective_rank}
For a threshold $\tau>0$, define the joint active support of BaRA as
\begin{equation}
\mathcal{S}_{\Phi,\theta}(\xv)
=
\left\{
k\in\{1,\ldots,r\}:
|\Phi_k\theta_k(\xv)|>\tau
\right\}.
\end{equation}
The input-dependent effective rank is
\begin{equation}
s_{\Phi,\theta}(\xv)
=
\left|
\mathcal{S}_{\Phi,\theta}(\xv)
\right|.
\end{equation}
Given a training set $\mathcal{D}=\{(\xv_i,y_i)\}_{i=1}^{n}$, the empirical average effective rank is
\begin{equation}
\bar{s}_{\Phi,\theta}
=
\frac{1}{n}
\sum_{i=1}^{n}
s_{\Phi,\theta}(\xv_i).
\end{equation}
\end{definition}

\begin{assumption}[Boundedness and Lipschitz continuity]
\label{assump:bounded_lipschitz}
We assume that the inputs and adaptation parameters are bounded:
\begin{equation}
\|\xv\|_2\le R_x,\quad
\|A\|_F\le R_A,\quad
\|B\|_F\le R_B,
\end{equation}
and
\begin{equation}
\|\boldsymbol{\Phi}\|_{\infty}\le R_{\Phi},
\qquad
\|\boldsymbol{\theta}(\xv)\|_{\infty}\le R_{\theta}.
\end{equation}
The loss function $\ell(h(\xv),y)$ is assumed to be bounded by $M$ and $L$-Lipschitz with respect to the prediction $h(\xv)$.
\end{assumption}

\begin{theorem}[Generalization bound of BaRA]
\label{thm:bara_generalization_main}
Under Assumption~\ref{assump:bounded_lipschitz}, with probability at least $1-\delta$ over the draw of the training set $\mathcal{D}$, the following inequality holds uniformly for all $h\in\mathcal{H}_{\Phi,\theta}$:
\begin{equation}
\begin{split}
\label{eq:bara_generalization_main}
\mathcal{R}(h)
\le
\widehat{\mathcal{R}}_{\mathcal{D}}(h)
+
&\mathcal{O}
\left(
L R_A R_B R_{\Phi} R_{\theta} R_x
\sqrt{
\frac{
\bar{s}_{\Phi,\theta}\log r
}{n}
} \right.
+\\
&
\left.M
\sqrt{
\frac{
\log(1/\delta)
}{n}
}
\right),
\end{split}
\end{equation}
where $\mathcal{R}(h)$ and $\widehat{\mathcal{R}}_{\mathcal{D}}(h)$ denote the population risk and empirical risk, respectively.
\end{theorem}

Theorem~\ref{thm:bara_generalization_main} shows that the generalization gap of BaRA is controlled by the joint effective rank $\bar{s}_{\Phi,\theta}$ rather than the maximum rank $r$. 
This is fundamentally different from standard LoRA, where all rank components are always activated. 
For standard LoRA, one has
\begin{equation}
s_{\mathrm{LoRA}}(\xv)=r,
\qquad
\bar{s}_{\mathrm{LoRA}}=r,
\end{equation}
which leads to the complexity-dependent generalization gap
\begin{equation}
\mathcal{R}(h)-\widehat{\mathcal{R}}_{\mathcal{D}}(h)
=
\mathcal{O}
\left(
\sqrt{
\frac{
r\log r
}{n}
}
\right).
\end{equation}
In contrast, BaRA satisfies
\begin{equation}
\bar{s}_{\Phi,\theta}\ll r
\end{equation}
when the Gamma--Weibull latent structure induces sparse global and local gates. 
Therefore, the BaRA generalization gap becomes
\begin{equation}
\mathcal{R}(h)-\widehat{\mathcal{R}}_{\mathcal{D}}(h)
=
\mathcal{O}
\left(
\sqrt{
\frac{
\bar{s}_{\Phi,\theta}\log r
}{n}
}
\right),
\end{equation}
which is tighter than the bound of dense fixed-rank LoRA.

\begin{proposition}[Synergistic effect of global and local sparsity]
\label{prop:joint_sparsity_main}
Define the global active support and local active support as
\begin{equation}
\mathcal{S}_{\Phi}
=
\left\{
k:
|\Phi_k|>\tau_{\Phi}
\right\},
\qquad
s_{\Phi}
=
|\mathcal{S}_{\Phi}|,
\end{equation}
and
\begin{equation}
\mathcal{S}_{\theta}(\xv)
=
\left\{
k:
|\theta_k(\xv)|>\tau_{\theta}
\right\},
\qquad
s_{\theta}(\xv)
=
|\mathcal{S}_{\theta}(\xv)|.
\end{equation}
Then the joint support satisfies
\begin{equation}
\mathcal{S}_{\Phi,\theta}(\xv)
\subseteq
\mathcal{S}_{\Phi}
\cap
\mathcal{S}_{\theta}(\xv),
\end{equation}
and therefore
\begin{equation}
s_{\Phi,\theta}(\xv)
\le
\min
\left\{
s_{\Phi},
s_{\theta}(\xv)
\right\}.
\end{equation}
Consequently,
\begin{equation}
\bar{s}_{\Phi,\theta}
\le
\min
\left\{
s_{\Phi},
\bar{s}_{\theta}
\right\},
\qquad
\bar{s}_{\theta}
=
\frac{1}{n}\sum_{i=1}^{n}s_{\theta}(\xv_i).
\end{equation}
\end{proposition}

Proposition~\ref{prop:joint_sparsity_main} explains the hierarchical advantage of BaRA. 
The global latent variable $\boldsymbol{\Phi}$ performs dataset-level rank selection by suppressing rank components that are not consistently useful across the training data. 
The local latent variable $\boldsymbol{\theta}(\xv)$ further performs input-dependent activation within the globally selected rank dictionary. 
Their multiplicative interaction produces a joint effective rank that is no larger than either the global active rank or the average local active rank. 
As a result, BaRA obtains a tighter complexity control than using only a global gate or only a local gate.

\begin{remark}[Interpretation]
The global variable $\boldsymbol{\Phi}$ reduces variance by shrinking redundant rank components at the dataset level, while the local variable $\boldsymbol{\theta}(\xv)$ reduces bias by allowing input-dependent adaptation. 
Thus, the joint gate $\boldsymbol{\Phi}\odot\boldsymbol{\theta}(\xv)$ implements an adaptive bias--variance trade-off: simple inputs use fewer rank components, whereas difficult or ambiguous inputs can activate more components when necessary. 
This explains why BaRA improves generalization, uncertainty calibration, and robustness under distribution shift.
\end{remark}

\section{Experiments
}
In this section, we empirically validate the effectiveness of BaRA across a wide range of settings.
Consistent with the modeling motivations in Sec.~\ref{sec:bnlora}, the results demonstrate that BaRA provides three clear advantages over existing baselines:
\textit{(1)} accurate and well-calibrated uncertainty quantification,
\textit{(2)} stronger output diversity that translates into improved test-time scaling behavior, and
\textit{(3)} effective mitigation of catastrophic forgetting, substantially reducing the alignment tax.

\begin{table*}[t]
\centering
\caption{In-distribution experiment using Qwen2.5-7B. We report the mean and standard deviation of test set performance across six commonsense reasoning tasks.Bold and underlined results denote the best and second best mean performance on each metric/dataset. ACC ($\uparrow$), ECE ($\downarrow$), NLL ($\downarrow$).}
\renewcommand{\arraystretch}{1.25} 
{
\scalebox{1.00}{
\begin{tabular}{@{}llcccccccc@{}}
\toprule
\textbf{Metric} & \textbf{Method} & \textbf{Params (M)} & \textbf{WG-S} & \textbf{ARC-C} & \textbf{ARC-E} & \textbf{WG-M} & \textbf{OBQA} & \textbf{BoolQ} & \textbf{Average} \\
\midrule
\multirow{10}{*}{ACC ($\uparrow$)} 
& MLE & 3.768 & 78.86$\pm$0.8  & 89.53$\pm$0.4 & 95.60$\pm$0.2 & 82.30$\pm$0.7 & \underline{92.25}$\pm$0.9 & 89.06$\pm$0.2 & 87.93 \\
& MAP & 3.768 & 78.94$\pm$0.8 & 88.98$\pm$0.8 & 95.73$\pm$0.3 & 82.09$\pm$0.7 & 91.72$\pm$0.7 & 89.04$\pm$0.1 & 87.75 \\
& MC-Dropout & 3.768 & 78.57$\pm$0.4 & 89.44$\pm$0.3 & 95.77$\pm$0.4 & 82.72$\pm$1.0 & 91.80$\pm$0.6 & 88.99$\pm$0.1 & 87.88 \\
& Ensemble & 11.305 & 79.09$\pm$0.4 & 89.44$\pm$0.5 & 95.73$\pm$0.1 & \textbf{83.23}$\pm$0.4 & \textbf{92.70}$\pm$0.6 & \underline{89.13}$\pm$0.1 & \underline{88.22} \\
& Laplace & 3.768 & 77.28$\pm$0.6 & 85.25$\pm$1.3 & 95.34$\pm$0.5 & 81.99$\pm$0.7 & 91.68$\pm$0.4 & 87.77$\pm$0.1 & 86.55 \\
& BLoB & 5.403 & 78.66$\pm$0.7 & 89.53$\pm$0.8 & \underline{96.54}$\pm$0.3 & 82.30$\pm$0.3 & 91.72$\pm$0.7 & 89.05$\pm$0.2 & 87.97 \\
& ScalaBL & 3.769 & 78.64$\pm$0.4 & 90.16$\pm$0.8 & 96.26$\pm$0.1 & 81.42$\pm$0.3 & 90.90$\pm$0.5 & 88.48$\pm$0.1 & 87.64 \\
& C-LoRA & 4.275 & \textbf{80.04}$\pm$0.2 & 89.52$\pm$0.4 & 96.35$\pm$0.1 & 81.83$\pm$0.5 & 89.90$\pm$0.7 & 88.73$\pm$0.3 & 87.73 \\
& FVAE-LoRA & 33.53 & 77.06$\pm$0.7 & \textbf{93.18}$\pm$0.6 & 96.25$\pm$0.4 & 78.87$\pm$0.4 & 91.21$\pm$0.8 & 88.99$\pm$0.3 & 87.59 \\
\cmidrule{2-10}
& \textbf{BaRA} & 5.608 & \underline{79.29}$\pm$0.6 & \underline{92.23}$\pm$0.6 & \textbf{96.56}$\pm$0.4 & \underline{82.96}$\pm$0.3 & 91.25$\pm$0.7 & \textbf{89.45}$\pm$0.2 & \textbf{88.62} \\
\midrule
\multirow{10}{*}{ECE ($\downarrow$)} 
& MLE & 3.768 & 20.14$\pm$0.9 & 10.11$\pm$0.5 & 4.17$\pm$0.2 & 16.10$\pm$0.6 & 6.40$\pm$0.8 & 3.79$\pm$0.1 & 10.12 \\
& MAP & 3.768 & 19.99$\pm$0.9 & 10.54$\pm$0.7 & 4.08$\pm$0.2 & 16.42$\pm$0.8 & 6.61$\pm$0.6 & 3.81$\pm$0.1 & 10.24 \\
& MC-Dropout & 3.768 & 20.15$\pm$0.3 & 10.06$\pm$0.4 & 4.01$\pm$0.3 & 15.46$\pm$0.8 & 6.60$\pm$0.4 & 3.88$\pm$0.1 & 10.03 \\
& Ensemble & 11.305 & 19.06$\pm$0.4 & 10.13$\pm$0.4 & 3.75$\pm$0.1 & 13.65$\pm$0.8 & 4.96$\pm$0.6 & 2.61$\pm$0.1 & 9.03 \\
& Laplace & 3.768 & 13.32$\pm$3.6 & 37.90$\pm$2.1 & 33.80$\pm$3.8 & 4.81$\pm$0.7 & \textbf{1.90}$\pm$0.4 & \textbf{1.18}$\pm$0.2 & 15.49 \\
& BLoB & 5.403 & \underline{7.88}$\pm$0.3 & \underline{4.03}$\pm$1.0 & \textbf{1.60}$\pm$0.4 & 5.08$\pm$0.4 & 2.16$\pm$0.5 & \underline{1.40}$\pm$0.3 & \underline{3.69} \\
& ScalaBL & 3.769 & 8.88$\pm$0.5 & 5.03$\pm$0.9 & 1.78$\pm$0.2 & \underline{3.64}$\pm$0.2 & 2.43$\pm$0.7 & 1.96$\pm$0.3 & 3.95 \\
& C-LoRA & 4.275 & 16.4$\pm$0.6 & 9.05$\pm$0.5 & 3.70$\pm$0.1 & 11.9$\pm$0.4 & 5.23$\pm$0.8 & 2.50$\pm$0.1 & 8.14 \\
& FVAE-LoRA & 33.53 & 11.27$\pm$0.5 & 4.19$\pm$0.3 & 2.91$\pm$0.2 & 5.54$\pm$0.3 & \underline{2.08}$\pm$0.4 & 4.92$\pm$0.2 & 5.15 \\
\cmidrule{2-10}
& \textbf{BaRA} & 5.608 & \textbf{3.54}$\pm$0.5 & \textbf{3.54}$\pm$0.7 & \underline{1.67}$\pm$0.3 & \textbf{2.80}$\pm$0.3 & 2.20$\pm$0.5 & 1.74$\pm$0.4 & \textbf{2.58} \\
\midrule
\multirow{10}{*}{NLL ($\downarrow$)} 
& MLE & 3.768 & 1.94$\pm$0.3 & 1.05$\pm$0.1 & 0.44$\pm$0.0 & 1.20$\pm$0.1 & 0.38$\pm$0.1 & 0.25$\pm$0.0 & 0.88 \\
& MAP & 3.768 & 1.88$\pm$0.2 & 1.05$\pm$0.1 & 0.43$\pm$0.0 & 1.27$\pm$0.2 & 0.39$\pm$0.0 & 0.25$\pm$0.0 & 0.88 \\
& MC-Dropout & 3.768 & 1.90$\pm$0.2 & 1.02$\pm$0.1 & 0.43$\pm$0.0 & 1.07$\pm$0.0 & 0.36$\pm$0.0 & 0.25$\pm$0.1 & 0.84 \\
& Ensemble & 11.305 & 1.33$\pm$0.1 & 0.75$\pm$0.0 & 0.25$\pm$0.0 & 0.74$\pm$0.0 & 0.27$\pm$0.0 & 0.24$\pm$0.0 & 0.60 \\
& Laplace & 3.768 & 0.55$\pm$0.0 & 0.80$\pm$0.1 & 0.51$\pm$0.1 & 0.44$\pm$0.0 & 0.23$\pm$0.0 & 0.29$\pm$0.1 & 0.47 \\
& BLoB & 5.403 & \underline{0.51}$\pm$0.0 & 0.30$\pm$0.0 & \textbf{0.10}$\pm$0.0 & \underline{0.39}$\pm$0.0 & \textbf{0.21}$\pm$0.0 & \underline{0.23}$\pm$0.0 & \underline{0.29} \\
& ScalaBL & 3.769 & \underline{0.51}$\pm$0.0 & 0.31$\pm$0.0 & \underline{0.11}$\pm$0.0 & 0.40$\pm$0.0 & \underline{0.23}$\pm$0.0 & 0.24$\pm$0.0 & 0.30 \\
& C-LoRA & 4.275 & 1.18$\pm$0.0 & 0.62$\pm$0.0 & 0.21$\pm$0.0 & 0.61$\pm$0.0 & 0.31$\pm$0.0 & 0.27$\pm$0.0 & 0.53 \\
& FVAE-LoRA & 33.53 & 0.62$\pm$0.1 & \underline{0.27}$\pm$0.0 & 0.13$\pm$0.0 & 0.47$\pm$0.0 & \underline{0.23}$\pm$0.0 & 0.29$\pm$0.0 & 0.33 \\
\cmidrule{2-10}
& \textbf{BaRA} & 5.608 & \textbf{0.44}$\pm$0.1 & \textbf{0.25}$\pm$0.0 & \textbf{0.10}$\pm$0.0 & \textbf{0.36}$\pm$0.0 & \textbf{0.21}$\pm$0.0 & \textbf{0.22}$\pm$0.0 & \textbf{0.26} \\
\bottomrule
\end{tabular}
}}
\label{tab:results_with_row_average}
\end{table*}

\begin{table*}[t]
\centering
\caption{Out-of-distribution experiment using Qwen2.5-7B.We report the mean and standard deviation of test set performance for models fine-tuned on the OBQA dataset across four Distribution Shift tasks. Bold and underlined results denote the best and second best mean performance on each metric/dataset.}
\renewcommand{\arraystretch}{1.25} 
 {\scalebox{1.00}{\begin{tabular}{@{}llcccccccc@{}}
\toprule
\textbf{Metric} & \textbf{Method} & \textbf{Params (M)} & \textbf{OBQA} & \textbf{ARC-C} & \textbf{ARC-E} & \textbf{Chemistry} & \textbf{Physics} & \textbf{Average} \\
\midrule
\multirow{10}{*}{ACC ($\uparrow$)} 
& MLE & 3.768 & $\underline{92.25} \pm 0.9$ & $90.88 \pm 0.7$ & $95.64 \pm 0.5$ & $53.00 \pm 1.3$ & $53.00 \pm 1.5$ & $73.13$ \\
& MAP & 3.768 & $91.72 \pm 0.7$ & $90.20 \pm 0.9$ & $95.53 \pm 0.6$ & $53.50 \pm 0.9$ & $53.25 \pm 3.1$ & $73.12$ \\
& MC-Dropout & 3.768 & $91.80 \pm 0.6$ & $90.37 \pm 0.5$ & $95.51 \pm 0.4$ & $52.75 \pm 1.3$ & $51.00 \pm 2.1$ & $72.41$ \\
& Ensemble & 11.305 & $\textbf{92.70} \pm 0.6$ & $90.84 \pm 0.6$ & $95.71 \pm 0.5$ & $53.25 \pm 1.0$ & $\underline{53.88} \pm 1.2$ & $73.42$ \\
& Laplace & 3.768 & $91.68 \pm 0.4$ & $90.51 \pm 0.7$ & $95.61 \pm 0.4$ & $48.75 \pm 1.8$ & $50.74 \pm 2.3$ & $71.40$ \\
& BLoB & 5.403 & $91.72 \pm 0.7$ & $\textbf{92.49} \pm 0.5$ & $\textbf{96.07} \pm 0.5$ & $\underline{54.69} \pm 1.4$ & $53.65 \pm 2.8$ & $\underline{74.23}$ \\
& ScalaBL & 3.769 & $90.90 \pm 0.5$ & $91.06 \pm 1.1$ & $95.74 \pm 0.5$ & $52.60 \pm 1.8$ & $53.13 \pm 1.5$ & $73.13$ \\
& C-LoRA & 4.275 & $89.90 \pm 0.7$ & $91.74 \pm 0.7$ & $95.51 \pm 0.4$ & $54.50 \pm 1.5$ & $49.02 \pm 2.9$ & $72.69$ \\
& FVAE-LoRA & 33.53 & $91.21 \pm 0.8$ & $\underline{91.93} \pm 0.8$ & $95.89 \pm 0.5$ & $53.08 \pm 1.6$ & $51.96 \pm 2.1$ & $73.22$ \\
\cmidrule{2-9}
& BaRA & 5.608 & $91.25 \pm 0.7$ & $91.89 \pm 0.6$ & $\underline{96.04} \pm 0.3$ & $\textbf{55.73} \pm 1.3$ & $\textbf{54.17} \pm 1.8$ & $\textbf{74.46}$ \\
\midrule
\multirow{10}{*}{ECE ($\downarrow$)} 
& MLE & 3.768 & $6.40 \pm 0.8$ & $7.72 \pm 0.6$ & $3.48 \pm 0.4$ & $23.29 \pm 2.2$ & $23.22 \pm 3.2$ & $14.43$ \\
& MAP & 3.768 & $6.61 \pm 0.6$ & $7.89 \pm 0.9$ & $3.31 \pm 0.2$ & $22.90 \pm 1.9$ & $21.52 \pm 4.4$ & $13.91$ \\
& MC-Dropout & 3.768 & $6.60 \pm 0.4$ & $7.63 \pm 0.8$ & $3.38 \pm 0.2$ & $23.74 \pm 1.6$ & $21.61 \pm 2.1$ & $14.09$ \\
& Ensemble & 11.305 & $4.96 \pm 0.6$ & $6.18 \pm 0.6$ & $2.63 \pm 0.4$ & $19.49 \pm 1.4$ & $17.33 \pm 2.0$ & $11.41$ \\
& Laplace & 3.768 & $\textbf{1.90} \pm 0.4$ & $4.75 \pm 0.7$ & $1.99 \pm 0.4$ & $14.31 \pm 2.1$ & $\textbf{11.94} \pm 4.5$ & $\underline{8.25}$ \\
& BLoB & 5.403 & $2.16 \pm 0.5$ & $4.46 \pm 0.5$ & $2.35 \pm 0.4$ & $16.21 \pm 2.2$ & $16.93 \pm 2.4$ & $9.99$ \\
& ScalaBL & 3.769 & $2.43 \pm 0.7$ & $\underline{4.41} \pm 0.7$ & $1.92 \pm 0.4$ & $16.94 \pm 1.8$ & $16.29 \pm 1.8$ & $9.89$ \\
& C-LoRA & 4.275 & $5.23 \pm 0.8$ & $4.44 \pm 0.7$ & $\textbf{1.32} \pm 0.4$ & $\underline{14.10} \pm 0.8$ & $15.59 \pm 2.1$ & $8.86$ \\
& FVAE-LoRA & 33.53 & $\underline{2.08} \pm 0.4$ & $4.52 \pm 0.6$ & $1.74 \pm 0.5$ & $17.49 \pm 1.9$ & $15.50 \pm 2.3$ & $9.81$ \\
\cmidrule{2-9}
& BaRA & 5.608 & $2.20 \pm 0.4$ & $\textbf{3.75} \pm 0.3$ & $\underline{1.35} \pm 0.3$ & $\textbf{13.00} \pm 1.1$ & $\underline{14.10} \pm 2.6$ & $\textbf{8.05}$ \\
\midrule
\multirow{10}{*}{NLL ($\downarrow$)} 
& MLE & 3.768 & $0.38 \pm 0.1$ & $0.44 \pm 0.0$ & $0.23 \pm 0.0$ & $1.53 \pm 0.1$ & $1.18 \pm 0.1$ & $0.85$ \\
& MAP & 3.768 & $0.39 \pm 0.0$ & $0.46 \pm 0.0$ & $0.22 \pm 0.0$ & $1.52 \pm 0.1$ & $1.19 \pm 0.1$ & $0.85$ \\
& MC-Dropout & 3.768 & $0.36 \pm 0.0$ & $0.43 \pm 0.0$ & $0.21 \pm 0.0$ & $1.50 \pm 0.1$ & $1.19 \pm 0.0$ & $0.83$ \\
& Ensemble & 11.305 & $0.27 \pm 0.0$ & $0.33 \pm 0.0$ & $0.17 \pm 0.0$ & $1.29 \pm 0.0$ & $1.07 \pm 0.0$ & $0.72$ \\
& Laplace & 3.768 & $\underline{0.23} \pm 0.0$ & $0.32 \pm 0.0$ & $0.15 \pm 0.0$ & $\textbf{1.11} \pm 0.0$ & $1.03 \pm 0.0$ & $0.65$ \\
& BLoB & 5.403 & $\textbf{0.21} \pm 0.0$ & $0.28 \pm 0.0$ & $0.15 \pm 0.0$ & $1.32 \pm 0.1$ & $0.99 \pm 0.0$ & $0.69$ \\
& ScalaBL & 3.769 & $\underline{0.23} \pm 0.0$ & $\underline{0.27} \pm 0.0$ & $0.14 \pm 0.0$ & $1.26 \pm 0.0$ & $\underline{0.96} \pm 0.0$ & $0.66$ \\
& C-LoRA & 4.275 & $0.31 \pm 0.0$ & $0.28 \pm 0.0$ & $\textbf{0.12} \pm 0.1$ & $1.20 \pm 0.0$ & $1.02 \pm 0.0$ & $0.66$ \\
& FVAE-LoRA & 33.53 & $\underline{0.23} \pm 0.0$ & $\underline{0.27} \pm 0.0$ & $\underline{0.13} \pm 0.0$ & $\underline{1.18} \pm 0.0$ & $0.99 \pm 0.0$ & $\underline{0.64}$ \\ 
\cmidrule{2-9}
& BaRA & 5.608 & $\textbf{0.21} \pm 0.0$ & $\textbf{0.25} \pm 0.0$ & $\textbf{0.12} \pm 0.0$ & $1.19 \pm 0.0$ & $\textbf{0.93} \pm 0.0$ & $\textbf{0.62}$ \\
\bottomrule
\end{tabular}
}}
\label{tab:ood_qwen7b}
\end{table*}

\subsection{Quantitative Experiments: Commonsense Reasoning Task}
\label{sec:commonsense_reasoning}

\subsubsection{Fine-tuning and Evaluation}
We implement BaRA using the \textsc{BayesAdapt} library \cite{samplawski2025scalable} and fine-tune Qwen2.5-7B \cite{qwen2} on a suite of commonsense reasoning benchmarks.
The library provides standardized implementations of several Bayesian fine-tuning baselines.
For C-LoRA, we adopt the official implementation released by \cite{rahmati2025c}. Since no public code is available for FVAE-LoRA \cite{kumar2025latent}, we reproduce the method carefully following the original paper.

All experiments are conducted on a single NVIDIA RTX PRO 6000 GPU with 96GB memory. We fine-tune Qwen2.5-7B on six commonsense reasoning tasks: Winogrande-small (WG-S), Winogrande-medium (WG-M) \cite{sakaguchi2021winogrande}, ARC-Challenge (ARC-C), ARC-Easy (ARC-E) \cite{clark2018think}, OpenBookQA (OBQA) \cite{mihaylov1809can}, and BoolQ \cite{clark2019boolq}. These benchmarks include both binary and multiple-choice classification settings. For each task, we select the appropriate next-token logits according to the format of the target output and train the model to maximize the log-likelihood of the correct answer.

Following \cite{samplawski2025scalable} and \cite{wang2024blob}, we apply LoRA to the query and value projection matrices of each self-attention layer, as well as to the softmax output head. All models are fine-tuned for 5,000 steps using the AdamW optimizer, with a batch size of 4, a learning rate of $1\times 10^{-4}$, a LoRA rank of $r=8$, and a maximum sequence length of 300 tokens. For a fair and consistent evaluation, we uniformly adopt the final training checkpoint for all fine-tuned models. For BaRA, we place a $\mathrm{Gamma}(1,1)$ prior over the LoRA weights and model the variational posterior as a fully factorized Weibull distribution. Other Bayesian baselines follow their original prior and posterior specifications.

Following the experimental protocol of \cite{samplawski2025scalable} and \cite{wang2024blob}, 
we fine-tune and evaluate our approach using a suite of commonsense reasoning 
datasets shown in \ref{tab:commonsense_datasets}. These datasets are posed as multiple 
choice questions. Given an input prompt with a question, we elicit the LLM's 
softmax distribution over the next token. We then select the logits for each 
possible answer (e.g., A,B,C,D) and renormalize. In this way, we transform 
these commonsense reasoning tasks into a classification task. This makes it 
straightforward to compute standard uncertainty metrics.
\subsubsection{Baselines} 
We compare BaRA against a comprehensive set of strong baselines, including Deep Ensembles \cite{lakshminarayanan2017simple}, Monte Carlo Dropout \cite{gal2016dropout}, Bayesian LoRA with Backpropagation (BLoB) \cite{wang2024blob}, Scalable Bayesian Low-Rank Adaptation (ScalaBL) \cite{samplawski2025scalable}, Factorized VAE LoRA (FVAE-LoRA) \cite{kumar2025latent}, Contextual LoRA (C-LoRA) \cite{rahmati2025c}, Laplace-LoRA \cite{yang2024bayesian}, and deterministic LoRA trained with MLE and MAP objectives.

\subsubsection{Evaluation metrics}
We report accuracy (ACC), expected calibration error (ECE), and negative log-likelihood (NLL).
ACC measures predictive performance, while ECE and NLL jointly assess the quality of uncertainty estimates.
All results are averaged over three independent runs, and we report mean $\pm$ standard deviation.
For dataset of $N$ test instances $\mathbf{x}_n$ with correct label $y_n$ and a probabilistic model $P_{\boldsymbol{\theta}}$, NLL is defined as:
\begin{equation}
\text{NLL} = \frac{1}{N} \sum_{n=1}^{N} -\log P_{\boldsymbol{\theta}}(y_n|\mathbf{x}_n)
\label{eq:nll}
\end{equation}
That is, it is the expected negative log probability of the correct class under the model.

ECE measures how a model's confidence aligns with the accuracy of its predictions. It can be computed by binning the predictions by their confidence. We then compute a weighted average of the difference between the accuracy and confidence within each bin:
\begin{equation}
\text{ECE} = \sum_{k=1}^{K} \frac{|B_k|}{N} \left| \text{acc}(B_k) - \text{conf}(B_k) \right|
\label{eq:ece}
\end{equation}
where $K$ is the number of bins and $B_k$ is the set of samples in the $k$-th bin. Following \cite{wang2024blob}, we use $K=15$ in all experiments.

\begin{table*}[ht]
\centering
\caption{Commonsense Reasoning Datasets used in experiments. We note that the MMLU datasets are used only in the out-of-distribution experiments and therefore have no training samples.}
\renewcommand{\arraystretch}{1.25} 
 {\scalebox{1.00}{
\begin{tabular}{llccc}
\toprule
Dataset & Citation & \#Classes & Train Samples & Test Samples \\
\midrule
Winogrande-Small (WG-S) & \cite{sakaguchi2021winogrande} & 2 & 0.64K & 1.27K \\
Winogrande-Medium (WG-M) & \cite{sakaguchi2021winogrande} & 2 & 2.56K & 1.27K \\
ARC-Easy (ARC-E) & \cite{clark2018think} & 4 & 2.25K & 0.57K \\
ARC-Challenge (ARC-C) & \cite{clark2018think} & 4 & 1.12K & 0.30K \\
OpenBookQA (OBQA) & \cite{mihaylov1809can} & 4 & 4.96K & 0.50K \\
BoolQ & \cite{clark2019boolq} & 2 & 2.49K & 3.27K \\
MMLU-Chemistry & \cite{hendrycks2021measuring} & 4 & - & 0.10K \\
MMLU-Physics & \cite{hendrycks2021measuring} & 4 & - & 0.10K \\
\bottomrule
\end{tabular}}}
\label{tab:commonsense_datasets}
\end{table*}

\subsubsection{Results on In-distribution Commonsense Reasoning}
Table~\ref{tab:results_with_row_average} summarizes in-distribution results on six commonsense benchmarks.
BaRA achieves the highest average accuracy (\textbf{88.62}), outperforming all competing methods.
Notably, it attains state-of-the-art performance on ARC-C and ARC-E, demonstrating strong predictive capability across both challenging and relatively clean benchmarks.
Beyond accuracy, BaRA delivers the most reliable uncertainty estimates.
It achieves the lowest average NLL (\textbf{0.26}) and ECE (\textbf{2.58}), consistently improving over prior Bayesian LoRA methods such as BLoB and ScalaBL.
This reflects the benefit of jointly modeling uncertainty in both latent representations and adaptation parameters, rather than restricting uncertainty to low-rank weights alone.
From an efficiency perspective, BaRA strikes a favorable balance between performance and parameter cost.
With only 5.608M trainable parameters, it outperforms resource-intensive Deep Ensembles (11.305M parameters) in ACC and NLL while maintaining comparable calibration.
At the same time, it consistently surpasses lighter Bayesian LoRA variants (\eg, ScalaBL with 3.769M parameters), indicating that its gains stem from adaptive rank allocation and structured sparsity rather than brute-force parameter scaling.

\subsubsection{Results on Out-of-distribution Robustness}
To assess robustness under distribution shift, we fine-tune models on OpenBookQA (OBQA) and evaluate them on ARC and MMLU (Chemistry and Physics), representing small and large domain shifts, respectively.
Results are reported in Table~\ref{tab:ood_qwen7b}.
BaRA achieves the highest average OOD accuracy (\textbf{74.46}), outperforming all baselines.
The advantage is particularly pronounced on the large-shift MMLU tasks, where BaRA yields gains of 3--7\% over competing Bayesian LoRA methods (Chemistry: $55.73 \pm 1.3$, Physics: $54.17 \pm 1.8$).
These improvements highlight the effectiveness of BaRA’s context-dependent rank allocation when facing unseen domains. 
Regarding uncertainty estimation, BaRA again demonstrates superior robustness.
It attains the lowest average ECE (\textbf{8.05}) and NLL (\textbf{0.62}) under distribution shift, while maintaining stable calibration on the challenging MMLU benchmarks.
This stability suggests that the global sparsity prior prevents overconfident predictions when the data distribution changes.
Overall, the OOD results confirm that BaRA not only improves in-distribution performance but also generalizes well under distribution shift, offering a principled and parameter-efficient solution for uncertainty-aware fine-tuning of LLMs.

\begin{figure*}[t]
\centering
\subfloat[Chat.]{
\includegraphics[width=0.215\linewidth]{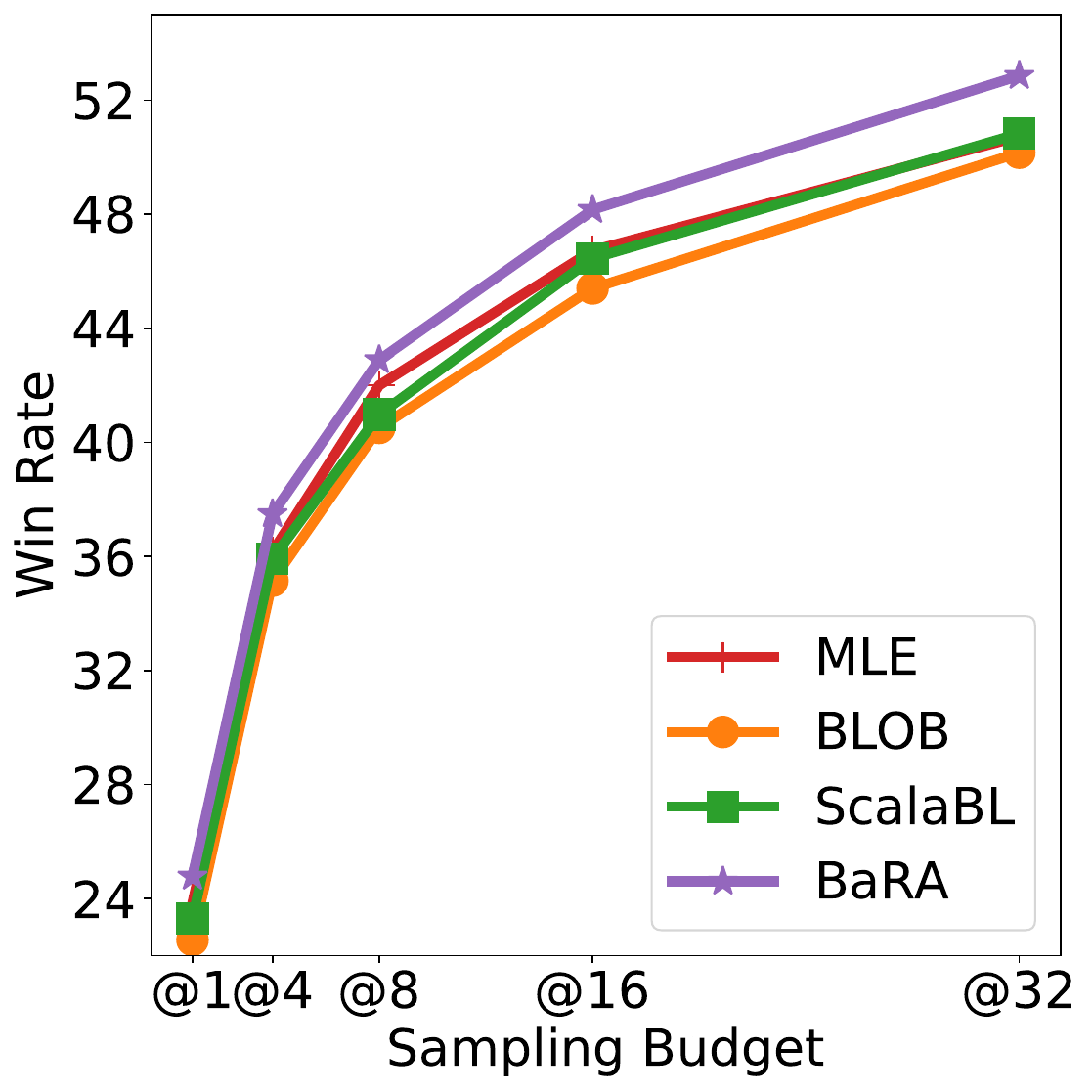}
\label{fig_chat_result}
} \quad
\subfloat[Code generation.]{
\includegraphics[width=0.213\linewidth]{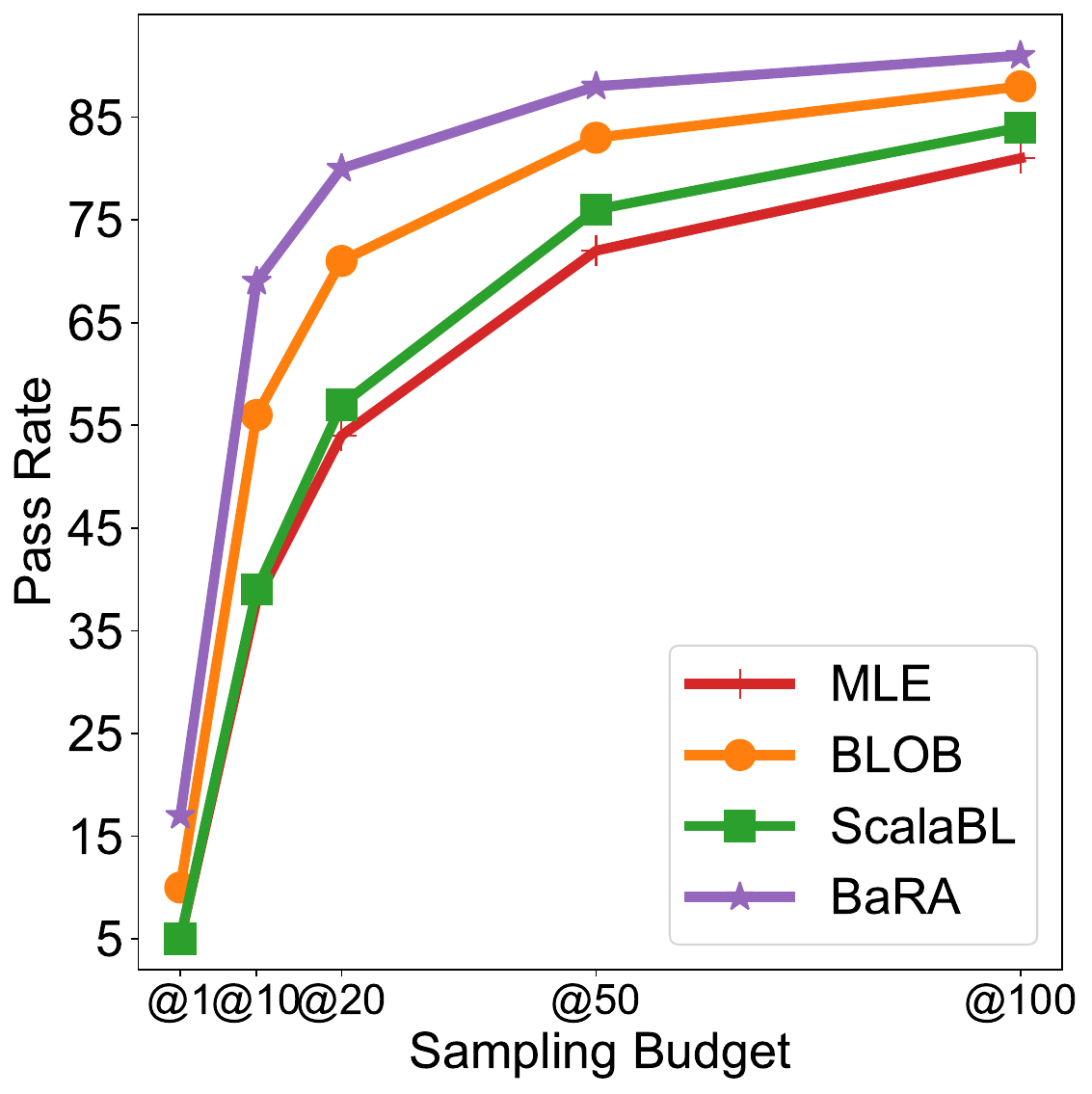}
\label{fig_code_result}
}\quad
\subfloat[Diversity
v.s. win rate.]{
\includegraphics[width=0.222\linewidth]{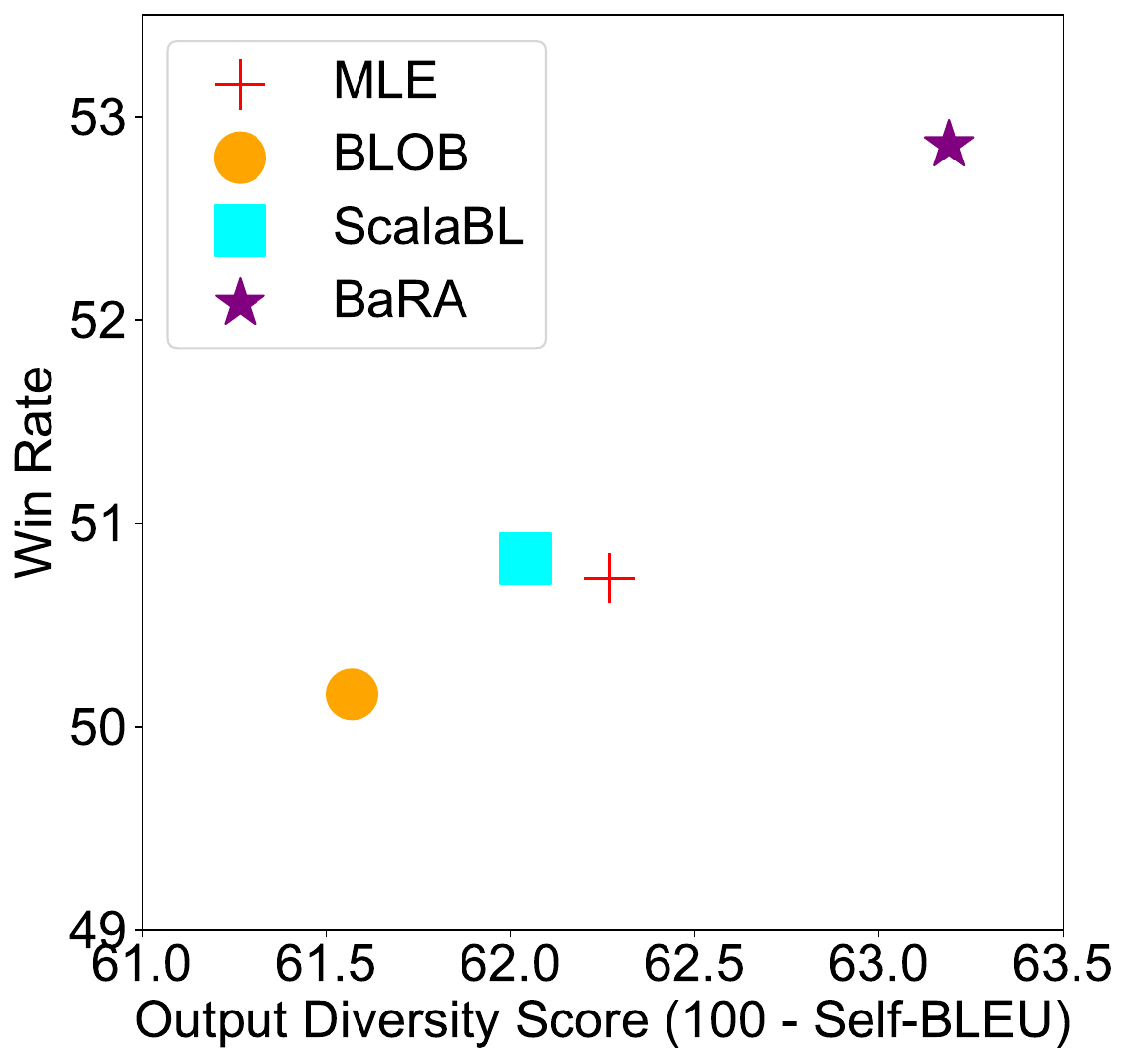}
\label{fig_chat_diverse}
}\quad
\subfloat[Entropy
v.s. Pass rate.]{
\includegraphics[width=0.215\linewidth]{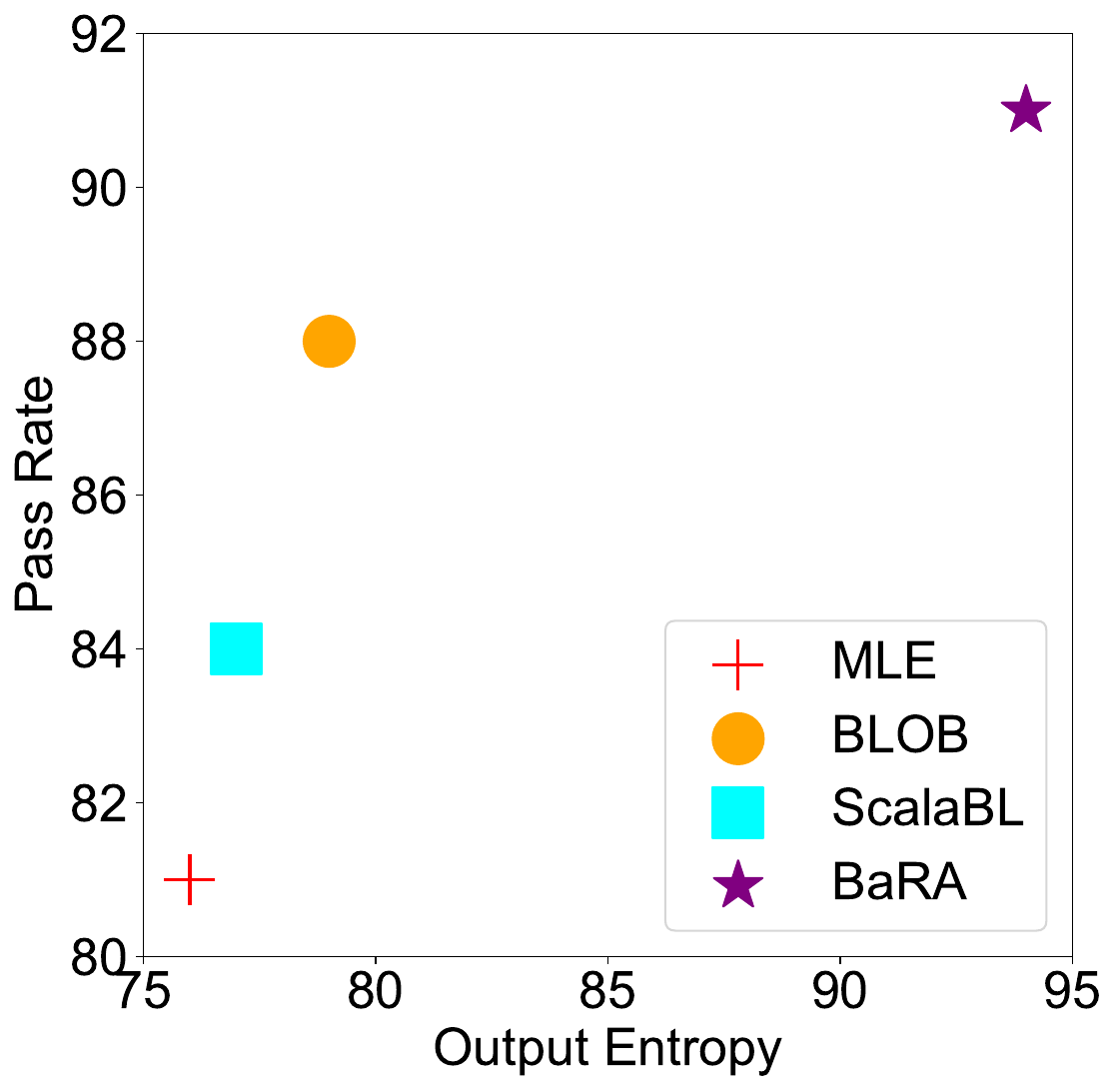}
\label{fig_code_diverse}
}
\caption{Performance of test-time scaling. The results demonstrate that BaRA achieves better performance with the same sampling budget and is more efficient in reaching comparable performance.}
\label{fig_app: uncertainty}
\end{figure*} 

\newcommand{\deltaup}[1]{\textsuperscript{\raisebox{0.25ex}{\hspace{0.10em}\textcolor{green!55!black}{\tiny$\uparrow$#1}}}}
\newcommand{\deltadown}[1]{\textsuperscript{\raisebox{0.25ex}{\hspace{0.10em}\textcolor{red!70!black}{\tiny$\downarrow$#1}}}}



\begin{figure*}
\centering
    \includegraphics[width=0.98 \linewidth]{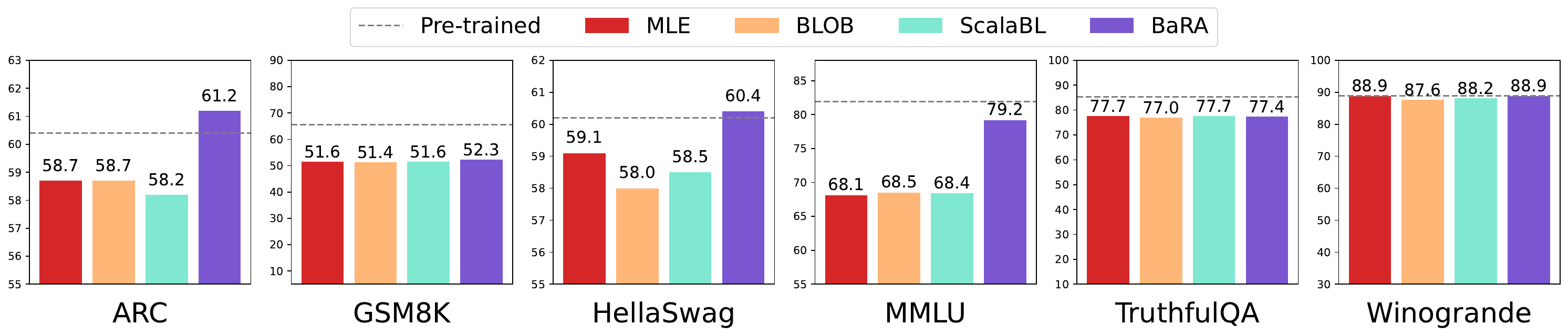}
    \centering
\caption{Performance on tasks from the OpenLLM leaderboard. The results indicate that BaRA outperforms other Bayesian LoRA, demonstrating a lower alignment tax.}
\label{fig_tax}
\end{figure*}

\subsection{Quantitative Experiments: Improving Diversity and Test Time Scaling Tasks}

\subsubsection{Fine-tuning Setup}  
We fine-tune the pretrained Qwen-2.5-7B foundation model on the UltraFeedback dataset \cite{cui2023ultrafeedback}.
Specifically, this dataset comprises diverse prompts sourced from well-established instruction-tuning datasets (\eg, Evol-Instruct and UltraChat), paired with corresponding responses generated by state-of-the-art conversational models including GPT-4 and Llama-2-7B/13B/70B-Chat. 
While we follow the core experimental framework reported in \cite{li2024preserving} to ensure reproducibility, we adjusted several hyperparameters in a principled manner to better align with the architectural characteristics of the Qwen2.5-7B model and our computational constraints: the learning rate was set to  $1 \times 10^{-4}$, a linear learning rate schedule was adopted, and using rank of $r = 8$ . The model is trained for a total of 2 epochs, and the BaRA-specific hyperparameters were set identical to those used in the Commonsense Reasoning experiment (\ref{sec:commonsense_reasoning}).
Additionally, the maximum sequence length—encompassing both the input prompt and the corresponding response—is constrained to 2,048 tokens to balance model capacity and computational efficiency.

\subsubsection{Evaluation Protocol}
To evaluate {chat generation}, we adopt a best-of-$N$ sampling protocol on AlpacaEval \cite{li2023alpacaeval}, a standard benchmark for instruction-following and open-ended dialogue.
Each fine-tuned model is prompted with 805 evaluation queries, generating 32 candidate responses per prompt.
A strong and independent reward model, \textsc{FsfairX-LLaMA3-RM-v0.16}, which achieves state-of-the-art performance on RewardBench \cite{lambert2025rewardbench}, is used to select the best response.
This protocol isolates the effect of generative diversity and uncertainty-aware sampling from reward model quality.
For {code generation}, we evaluate on the HumanEval benchmark \cite{chen2021evaluating}, which consists of 163 Python programming problems.
Performance is measured using execution-based correctness under the standard pass@$k$ metric, reflecting the model’s ability to generate functionally correct programs.

\subsubsection{Results on Chat Generation.}
For AlpacaEval, the selected responses are compared against GPT-4 outputs, and win rates are reported in Fig.~\ref{fig_chat_result}.
Under the same sampling budget, BaRA substantially outperforms all baseline methods, including MLE, BLoB, and ScalaBL.
More strikingly, BaRA achieves comparable or superior win rates while requiring only \emph{approximately half} the number of samples.
This demonstrates that BaRA is significantly more \emph{sampling-efficient}, benefiting from its uncertainty-aware and context-adaptive rank allocation, which promotes diverse yet high-quality candidate generations.

\subsubsection{Result on Code Generation.}
Results on HumanEval are summarized in Fig.~\ref{fig_code_result}.
BaRA consistently achieves higher pass@$k$ scores than competing approaches across all values of $k$.
Notably, BaRA reaches parity with strong baselines using only $50\%$ of their sampling budget, indicating that its advantages extend beyond natural language generation to structured, execution-sensitive tasks.
This improvement suggests that adaptive rank allocation enables BaRA to flexibly allocate modeling capacity to difficult inputs, yielding more reliable program synthesis.

\subsubsection{Diversity and Test-Time Scaling.}
Across both chat and code benchmarks, we observe a strong positive correlation between output diversity and downstream performance \cite{li2024preserving}.
Specifically, higher diversity (measured by $100-\text{Self-BLEU}$) correlates with improved chat win rates, while higher predictive entropy correlates with higher pass@$k$ scores in code generation.
Among all methods, BaRA consistently achieves the highest diversity and entropy, supporting the hypothesis that its sparse Bayesian formulation implicitly encourages entropy maximization.
This property enables BaRA to better exploit test-time scaling, converting additional samples into tangible performance gains more effectively than existing Bayesian and deterministic LoRA variants.

\subsubsection{Results on Alleviating Alignment Tax}

We further evaluate whether BaRA mitigates catastrophic forgetting and reduces alignment tax.
Following prior work, we assess all models on six downstream benchmarks from the OpenLLM leaderboard: ARC, GSM8K, HellaSwag, MMLU, TruthfulQA, and WinoGrande.
Results are summarized in Fig.~\ref{fig_tax}.
Across most benchmarks, baseline methods exhibit noticeable performance degradation after supervised fine-tuning, with cross-entropy (CE) suffering the most severe drops.
In contrast, BaRA consistently preserves downstream performance and avoids significant degradation, particularly on challenging benchmarks such as ARC and HellaSwag, where both deterministic LoRA and Bayesian baselines (\eg, MLE, BLoB) experience marked declines.
Quantitatively, BaRA reduces alignment tax by approximately $80\%$ relative to CE.
While CE incurs an average performance drop of 1.5 points across tasks, BaRA limits this degradation to only 0.3 points, demonstrating a substantially improved trade-off between alignment and generalization.
These results highlight the role of Bayesian sparsity and adaptive rank allocation in preventing over-specialization during fine-tuning.

\subsection{Quantitative Experiments: Ablation Study}

\subsubsection{Ablation Study on Parameter Budgets}

We first investigate the effect of parameter budget to examine whether the performance gains of BaRA are simply attributable to a larger number of trainable parameters. For fair comparison, we increase the rank of Bayesian baselines BLoB and ScalaBL to 16 and conduct evaluations on commonsense reasoning tasks with Qwen2.5-7B. As shown in Table \ref{tab:parameter_budget}, BaRA achieves superior performance with much fewer parameters even when the baseline methods are equipped with larger parameter budgets. The results indicate that the improvements stem from the proposed sparse Bayesian adaptation mechanism, adaptive rank allocation and structured sparsity, rather than from increased model capacity.

\begin{table}[htbp]
\centering
\caption{Comparison with parameter-matched baselines on Qwen2.5-7B (average over commonsense reasoning tasks)}
\label{tab:parameter_budget}
\renewcommand{\arraystretch}{1.25} 
{
\begin{tabular}{l S[table-format=2.2] S[table-format=2.2] S[table-format=2.2] S[table-format=1.2]}
\toprule
Method           & Params(M) & {ACC ($\uparrow$)} & {ECE ($\downarrow$)} & {NLL ($\downarrow$)} \\
\midrule
ScalaBL (rank=16) & 7.54   & 87.24 & 4.08  & 0.30  \\
BLoB (rank=16)    & 10.81  & 87.24 & 2.98  & 0.29  \\
BaRA (Ours)       & 5.61   & \textbf{88.62} & \textbf{2.58}  & \textbf{0.26}  \\
\bottomrule
\end{tabular}}
\end{table}

\subsubsection{Generalization beyond Qwen2.5-7B}

To evaluate the generalizability of BaRA, we further conduct experiments on \textbf{LLaMA-2-7B} in addition to Qwen2.5-7B. As shown in Table~\ref{tab:llama2_7b_results}, BaRA achieves competitive accuracy and the best uncertainty estimation quality in terms of both ECE and NLL. These results demonstrate that BaRA is not tied to a specific backbone and can generalize effectively to different large language models.

\begin{table}[htbp]
\centering
\caption{Performance comparison on LLaMA-2-7B.}
\label{tab:llama2_7b_results}
\renewcommand{\arraystretch}{1.25} 
{
\begin{tabular}{l S[table-format=2.2] S[table-format=2.2] S[table-format=1.2]}
\toprule
Method & {ACC ($\uparrow$)} & {ECE ($\downarrow$)} & {NLL ($\downarrow$)} \\
\midrule
MLE (LoRA) & \textbf{78.95} & 18.14 & 1.41 \\
ScalaBL & 78.12 & 5.43 & 0.51 \\
BLoB & 77.61 & 5.93 & 0.54 \\
BaRA (Ours) & 78.24 & \textbf{5.09} & \textbf{0.47} \\
\bottomrule
\end{tabular}}
\end{table}

\subsubsection{Efficiency and computational cost}
We also evaluate the computational overhead of BaRA in terms of peak GPU memory usage, training time, and inference time. As reported in Table~\ref{tab:efficiency_profiling}, BaRA exhibits a moderate increase in training and inference cost compared with the baselines, while maintaining comparable memory consumption. This trade-off is expected, since BaRA performs Bayesian optimization over a structured latent space to improve sparsity control and uncertainty modeling.

\begin{table*}[htbp]
\centering
\caption{Efficiency profiling on the WG-S dataset.}
\label{tab:efficiency_profiling}
\renewcommand{\arraystretch}{1.25} 
{
\begin{tabular}{lccc}
\toprule
Method & Peak GPU Memory (GB) & Training Time (s) & Inference Time (s) \\
\midrule
ScalaBL & 17.68 & 1063 & 175 \\
BLoB & 17.47 & 1039 & 167 \\
BaRA & 17.18 & 1487 & 223 \\
\bottomrule
    \end{tabular}}
\end{table*}

\subsubsection{Ablation on Hierarchical Decomposition }
We further study the effect of decoupling the global latent variable $\boldsymbol{\Phi}$ from the input-dependent latent variable $\boldsymbol{\theta}$, compared with a simplified single-latent formulation. This hierarchical decomposition provides two complementary benefits. First, $\boldsymbol{\Phi}$ captures task-level global importance and acts as a cross-input information bottleneck, which improves identifiability and regularization. Second, $\boldsymbol{\theta}$ enables token-level adaptive activation conditioned on the input context, thereby preserving instance-specific flexibility.
As shown in Table~\ref{tab:ablation_hierarchical}, the full hierarchical formulation $\boldsymbol{\Phi}\boldsymbol{\theta}$ achieves the best overall performance across accuracy, calibration, and likelihood. This confirms that combining global rank regularization with local adaptivity is beneficial for the proposed model.

The theoretical analysis suggests that the generalization gap of BaRA is controlled by the joint effective rank $\bar{s}_{\Phi,\theta}$ rather than the maximum rank $r$. 
This prediction is consistent with the sparsity observations in our experiments. 
The learned global gate $\boldsymbol{\Phi}$ suppresses rank components that are not consistently useful across the dataset, while the local gate $\boldsymbol{\theta}(\xv)$ further activates only a context-dependent subset of the remaining components. 
As a result, the final gate $\boldsymbol{\Phi}\odot\boldsymbol{\theta}(\xv)$ induces substantially sparser effective activations than either standard LoRA or single-level gating variants. 
Empirically, this reduction in effective rank correlates with improved calibration and OOD robustness, supporting the view that BaRA improves generalization by reducing the effective complexity of the adaptation class while preserving input-dependent expressiveness.

\begin{table}[htbp]
\centering
\caption{Ablation on Hierarchical Decomposition (average commonsense reasoning tasks)}
\label{tab:ablation_hierarchical}
\renewcommand{\arraystretch}{1.25} 
{
\begin{tabular}{c 
                S[table-format=2.2]
                S[table-format=1.2]
                S[table-format=1.2]}
\toprule
Variant & {ACC} & {ECE} & {NLL} \\
\midrule
$\boldsymbol{\Phi}$ & 87.58 & 6.62 & 0.42 \\
\midrule
$\boldsymbol{\theta}$ & 88.62 & 4.54 & 0.30 \\
\midrule
$\boldsymbol{\Phi}\boldsymbol{\theta}$ & \textbf{88.65} & \textbf{3.73} & \textbf{0.27} \\
\bottomrule
\end{tabular}}
\end{table}

\subsubsection{Prior Sensitivity}

We also conduct an ablation study on the shape and rate parameters of the Gamma prior used to control sparsity. The results show that BaRA is robust to reasonable variations in the prior specification. This robustness can be attributed to the fact that the KL regularization term is weighted substantially less than the NLL term during optimization, so training is primarily driven by the data likelihood rather than the exact prior choice. As a result, the model maintains stable sparsity patterns and consistent performance across different prior settings.

\begin{table}[htbp]
\centering
\caption{Gamma prior ablation results}
\label{tab:gamma_prior_ablation}
\renewcommand{\arraystretch}{1.25} 
{
\begin{tabular}{lcccc}
\toprule
Prior & Sparsity Rate & ACC & ECE & NLL \\
\midrule
$\text{Gamma}(0.5, 1.0)$ & 22.66 & 79.43 & 4.15 & 0.60 \\

$\text{Gamma}(1.0, 1.0)$ & 23.23 & 79.29 & 3.50 & 0.44 \\

$\text{Gamma}(2.0, 1.0)$ & 21.26 & 79.27 & 3.96 & 0.52 \\

$\text{Gamma}(1.0, 0.5)$ & 21.81 & 80.63 & 4.28 & 0.46 \\

$\text{Gamma}(1.0, 2.0)$ & 20.68 & 78.71 & 4.12 & 0.50 \\
\bottomrule
\end{tabular}}
\end{table}

\subsection{Qualitative Analysis of Sparsity Structures}
\label{analysis_sparse}

\subsubsection{Experimental Setup}

We further investigate the sparsity behavior induced by BaRA from both
layer-wise and token-wise perspectives. The experiments are conducted on
Qwen2.5-7B using the ARC-Challenge dataset. We focus on the value projection
modules across all 28 Transformer layers and evaluate four rank configurations,
i.e., $r \in \{8,16,32,64\}$. Unless otherwise specified, sparsity is computed
from last-token activations with a threshold $\tau=0.1$.

We consider three sparsity metrics to characterize different sources of
sparsification. First, $\boldsymbol{\Phi}$ measures the sparsity of the
Weibull-parameterized diagonal modulation matrix
$\mathbf{D}_{I(\mathbf{x})}$ in BaRA, which reflects the sparsity of global
adaptation factors. Since standard LoRA does not contain such an explicit
diagonal modulation branch, we report $\boldsymbol{\Phi}=0$ for LoRA as a
bookkeeping value rather than as an empirical sparsity measurement. Second,
$\boldsymbol{\theta}$ measures the sparsity of instance-dependent latent
activations before modulation, corresponding to
$\mathbf{E}_{I(\mathbf{x})}(\mathbf{x})$ for BaRA and $A\mathbf{x}$ for LoRA.
Third, $\boldsymbol{\Psi}$ measures the sparsity of final adapted activations,
corresponding to
$\mathbf{D}_{I(\mathbf{x})}\mathbf{E}_{I(\mathbf{x})}(\mathbf{x})$ for BaRA
and $A\mathbf{x}$ for LoRA.

For a vector or matrix $\mathbf{z}$ with $N$ elements, sparsity is defined as
the percentage of entries whose absolute values are smaller than $\tau$:
\begin{equation}
\operatorname{Sparsity}(\mathbf{z};\tau)
=
\frac{100}{N}
\sum_{i=1}^{N}
\mathbb{I}\left(|z_i|<\tau\right).
\end{equation}
The metrics $\boldsymbol{\theta}$ and $\boldsymbol{\Psi}$ are computed on
last-token representations. All values reported in Table~\ref{tab:r_value_results_bara}
are percentages.

\begin{table}[tbp]
\centering
\caption{Sparsity of $\boldsymbol{\Phi}$, $\boldsymbol{\theta}$, and
$\boldsymbol{\Psi}$ for BaRA and LoRA under different rank configurations.
All values are reported in percentage.}
\renewcommand{\arraystretch}{1.15}
{\scalebox{1.00}{
\begin{tabular}{@{}clcccc@{}}
\toprule
\textbf{Metric} & \textbf{Method} & \textbf{$r=8$} & \textbf{$r=16$} & \textbf{$r=32$} & \textbf{$r=64$} \\
\midrule
\multirow{2}{*}{$\boldsymbol{\Phi}$}
& LoRA & $0$ & $0$ & $0$ & $0$ \\
& BaRA & $5.36$ & $4.91$ & $6.14$ & $5.52$ \\
\midrule
\multirow{2}{*}{$\boldsymbol{\theta}$}
& LoRA & $4.91$ & $13.62$ & $17.14$ & $20.85$ \\
& BaRA & $16.96$ & $18.12$ & $18.93$ & $20.53$ \\
\midrule
\multirow{2}{*}{$\boldsymbol{\Psi}$}
& LoRA & $4.91$ & $13.62$ & $17.14$ & $20.85$ \\
& BaRA & $34.38$ & $35.49$ & $37.61$ & $39.56$ \\
\bottomrule
\end{tabular}
}}
\label{tab:r_value_results_bara}
\end{table}

\subsubsection{Layer-wise Sparsity Analysis}

Table~\ref{tab:r_value_results_bara} shows that BaRA consistently induces
stronger effective sparsity than LoRA across all rank configurations. For LoRA,
the final activation sparsity $\boldsymbol{\Psi}$ is identical to the latent
activation sparsity $\boldsymbol{\theta}$ because LoRA does not introduce an
additional modulation mechanism. In contrast, BaRA combines instance-dependent
latent activations with a sparse diagonal modulation matrix, leading to a much
higher final sparsity $\boldsymbol{\Psi}$.

Specifically, BaRA maintains a stable diagonal modulation sparsity
$\boldsymbol{\Phi}$ of approximately $5\%$ across different ranks, suggesting
that the learned modulation factors are not overly sensitive to the rank
configuration. Meanwhile, BaRA achieves substantially higher final activation
sparsity, ranging from $34.38\%$ to $39.56\%$, whereas LoRA reaches only
$4.91\%$ to $20.85\%$. This gap indicates that BaRA does not merely increase
the sparsity of latent activations, but further amplifies sparsification through
the interaction between latent activation selection and diagonal modulation.

\begin{figure*}[t]
\centering
\subfloat[$r=8$]{
    \includegraphics[width=0.22\linewidth]{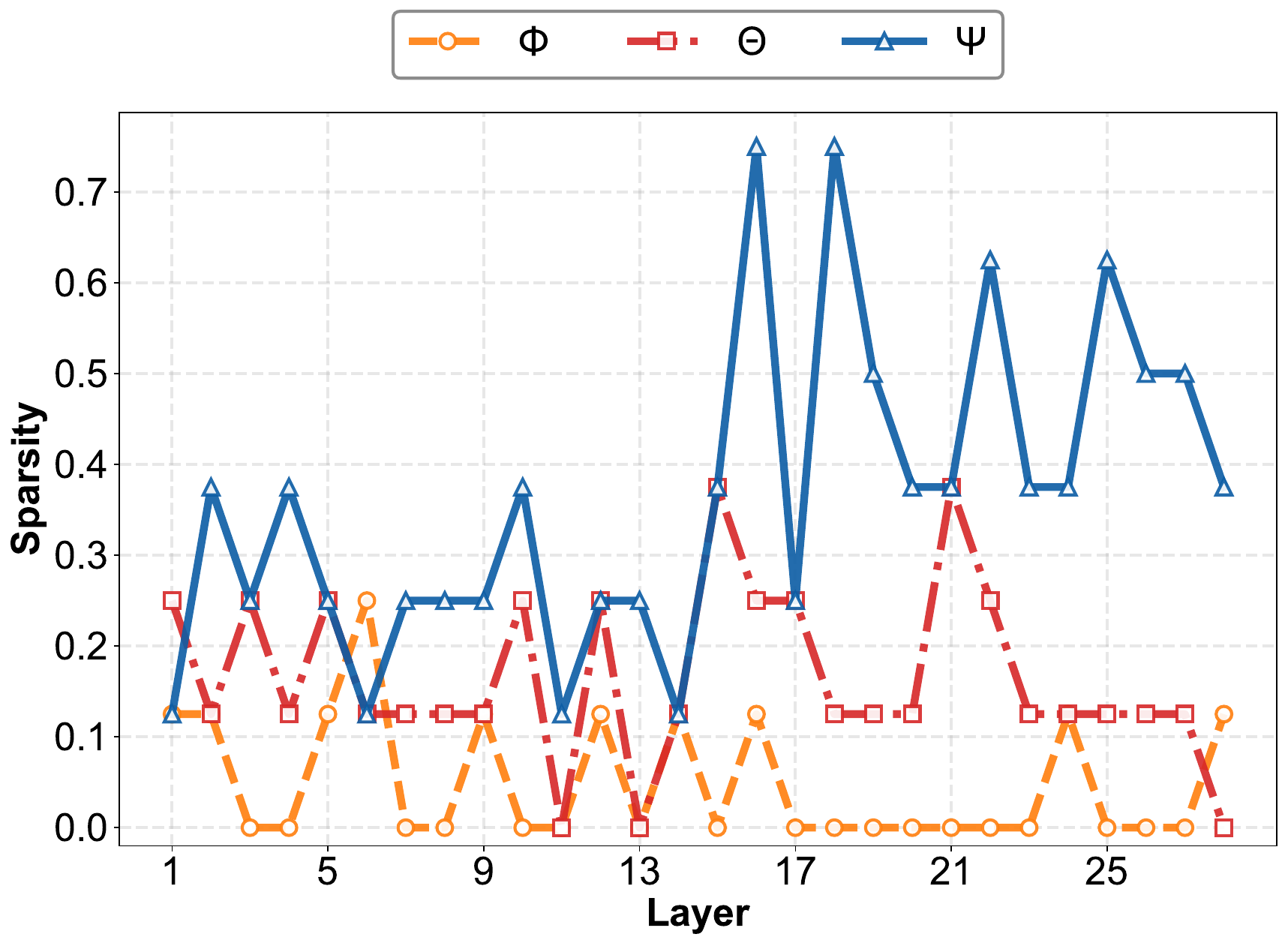}
    \label{appfig:sparsity_r8}}
\quad
\subfloat[$r=16$]{
    \includegraphics[width=0.22\linewidth]{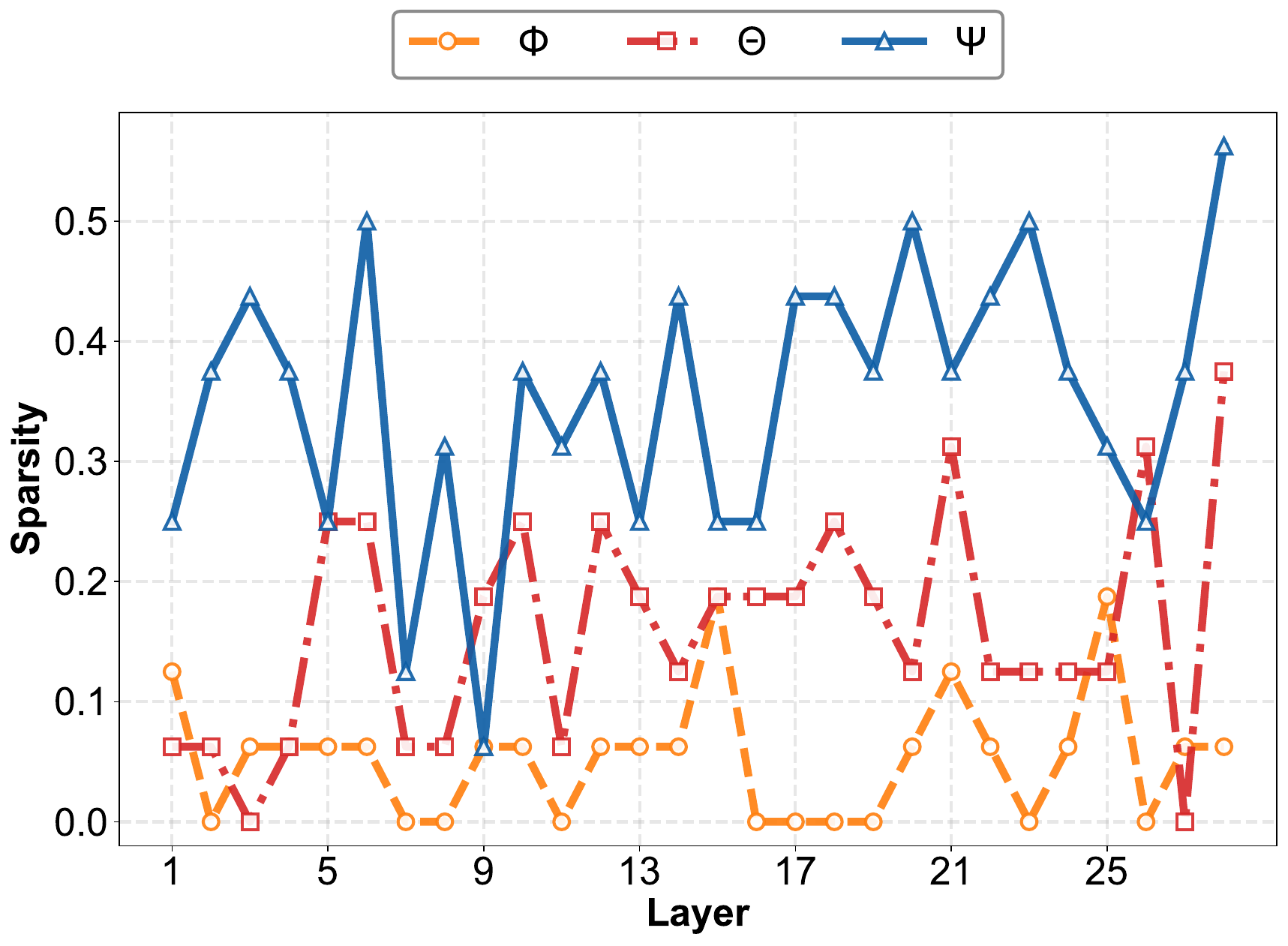}
    \label{appfig:sparsity_r16}}
\quad
\subfloat[$r=32$]{
    \includegraphics[width=0.22\linewidth]{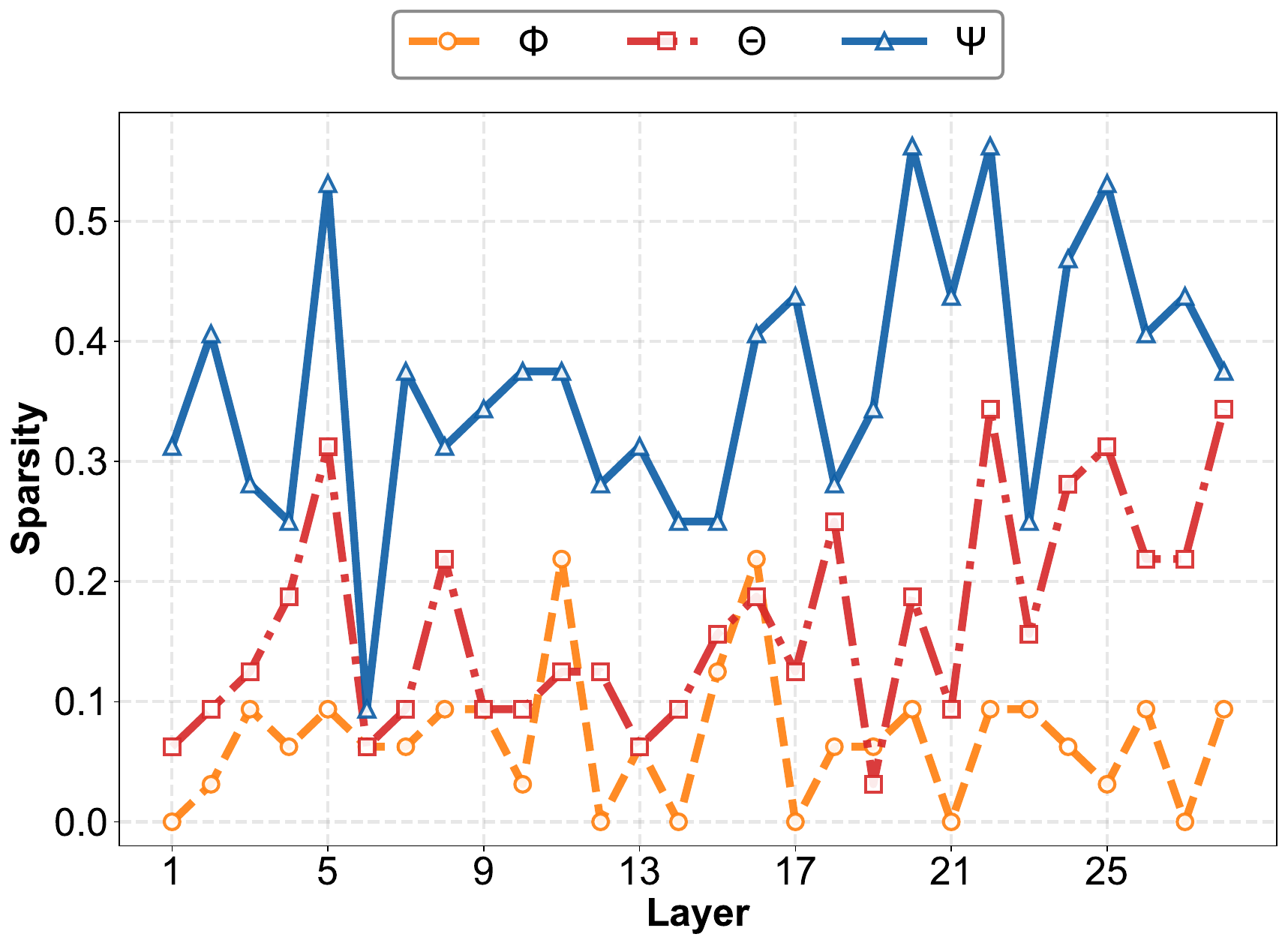}
    \label{appfig:sparsity_r32}}
\quad
\subfloat[$r=64$]{
    \includegraphics[width=0.22\linewidth]{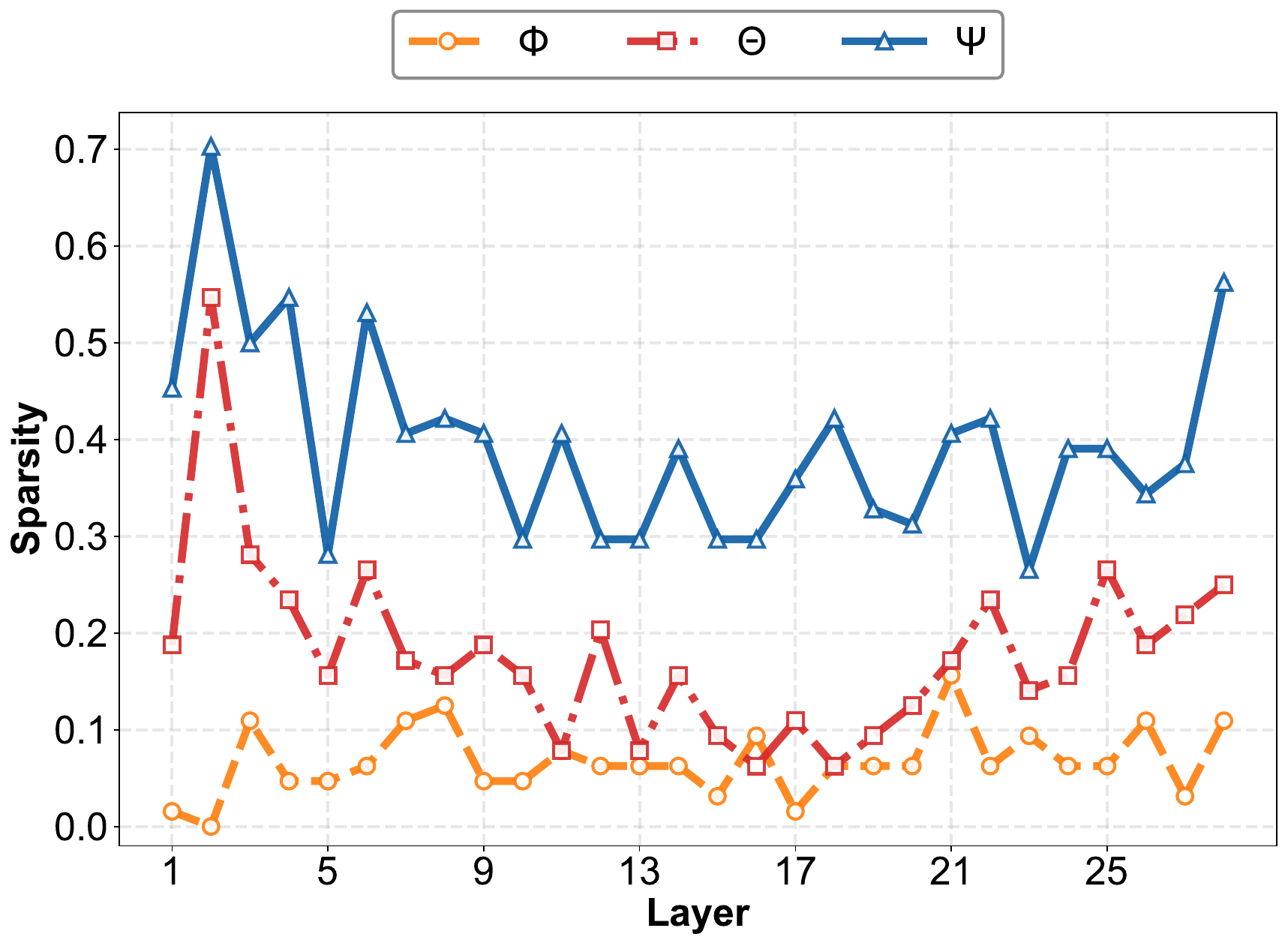}
    \label{appfig:sparsity_r64}}
\caption{Layer-wise sparsity distribution of the value projection module under
different rank configurations. Each subfigure shows the sparsity of the diagonal
modulation factors $\boldsymbol{\Phi}$, the pre-modulation latent activations
$\boldsymbol{\theta}$, and the final adapted activations $\boldsymbol{\Psi}$
across the 28 Transformer layers.}
\label{appfig:sparsity_2X2}
\end{figure*}

To provide a more detailed view, we visualize the layer-wise distribution of
$\boldsymbol{\Phi}$, $\boldsymbol{\theta}$, and $\boldsymbol{\Psi}$ across the
28 Transformer layers in Figure~\ref{appfig:sparsity_2X2}. The results reveal
that BaRA allocates sparsity heterogeneously across layers instead of enforcing
uniform compression. Layers that are less sensitive to sparsification receive
stronger modulation-level suppression, while layers that are more critical to
the model prediction preserve denser representations. This behavior suggests
that BaRA performs adaptive sparsity allocation conditioned on both the input
and the layer-wise functional role.

Figure~\ref{appfig:sparsity_2X2} further confirms the dual-filter effect of
BaRA. The final activation sparsity $\boldsymbol{\Psi}$ is consistently higher
than the sparsity of either individual component, demonstrating that the sparse
diagonal modulation and instance-dependent activation sparsification act
synergistically. Moreover, the close correlation between $\boldsymbol{\theta}$
and $\boldsymbol{\Psi}$ suggests that input-dependent activation sparsification
is the dominant source of effective compression, while the diagonal modulation
serves as an additional selective gate. This mechanism enables BaRA to suppress
redundant adaptation dimensions while preserving dense computation for
semantically important representations.

Overall, the layer-wise results demonstrate that BaRA adaptively balances
parameter-level and activation-level sparsity under different rank constraints.
By decoupling diagonal modulation from latent activation generation, BaRA
achieves higher effective sparsity than LoRA while maintaining a heterogeneous
allocation pattern across layers.

\subsubsection{Token-level Sparsity visualization Under Different Input Scenarios}

We next analyze token-level activation sparsity to examine whether BaRA exhibits
input-dependent sparsification across different semantic domains. The experiment
is conducted on Qwen2.5-7B by applying BaRA to the value projection modules in
all Transformer layers. We use English texts from diverse domains, including
daily life, scientific research, education, technology, and medical practice.
The sparsity threshold is set to $\tau=0.1$, the maximum sequence length is set
to 512, and all computations are performed using bfloat16 precision.

For each input text, we first extract the diagonal modulation factors and the
corresponding activation projections. We then perform 100 independent
reparameterization samples and compute the resulting adapted activations.
For each token, its sparsity score is obtained by averaging the proportion of
near-zero activation values over all samples and all layers. This procedure
allows us to visualize not only deterministic activation sparsity, but also the
expected sparsification behavior induced by the stochastic BaRA module.

In the visualization, the background intensity of each token indicates its
activation sparsity. Lighter colors correspond to higher sparsity, indicating
that the token activates fewer adaptation dimensions. Darker colors correspond
to lower sparsity, indicating denser adapted activations. Therefore, the
visualization reflects how BaRA allocates computational resources across tokens
according to their semantic roles and contextual importance.

\begin{figure*}[!t]
\centering

\subfloat[Water-Saving Irrigation\label{appfig:sparsity_water_saving}]{
    \includegraphics[width=0.3\linewidth]{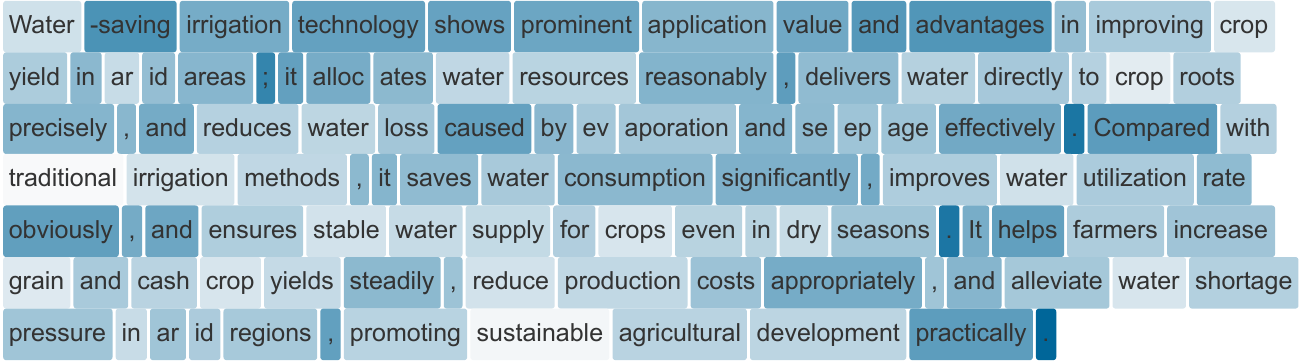}
}
\hfill
\subfloat[Intangible Heritage\label{appfig:sparsity_intangible_heritage}]{
    \includegraphics[width=0.3\linewidth]{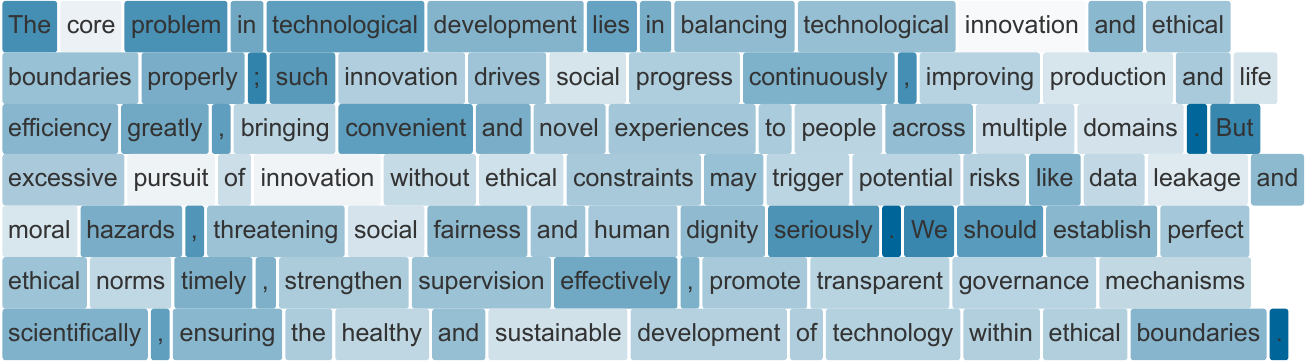}
}
\hfill
\subfloat[Tech Ethics\label{appfig:sparsity_tech_ethics}]{
    \includegraphics[width=0.3\linewidth]{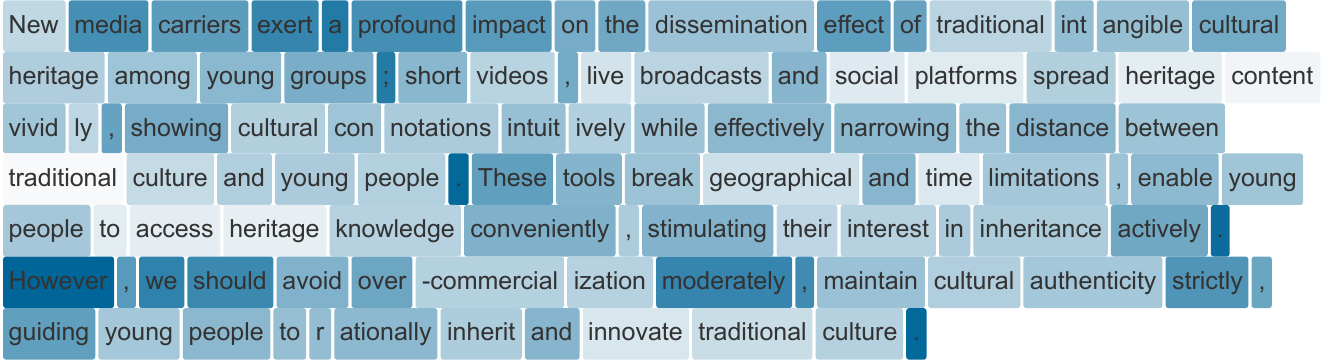}
}

\vspace{1em}

\subfloat[User Behavior\label{appfig:sparsity_user_behavior}]{
    \includegraphics[width=0.3\linewidth]{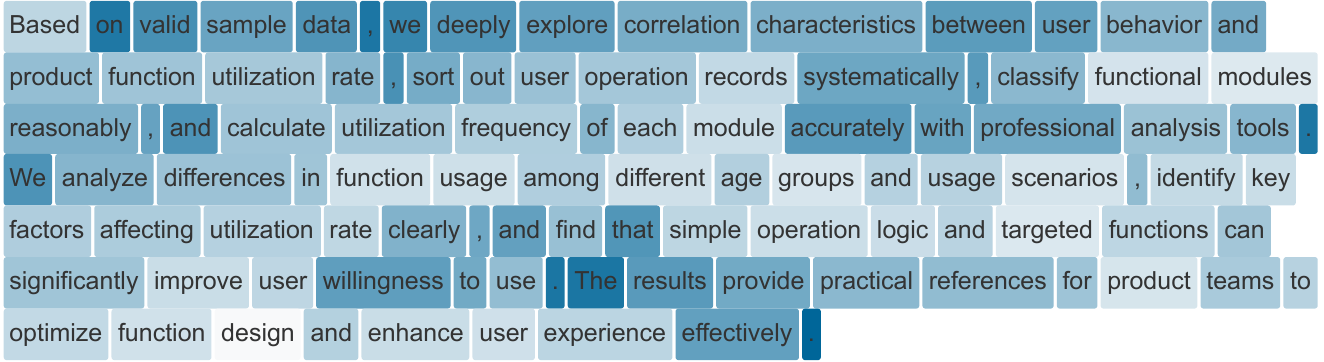}
}
\hfill
\subfloat[Respiratory Infections\label{appfig:sparsity_respiratory_infections}]{
    \includegraphics[width=0.3\linewidth]{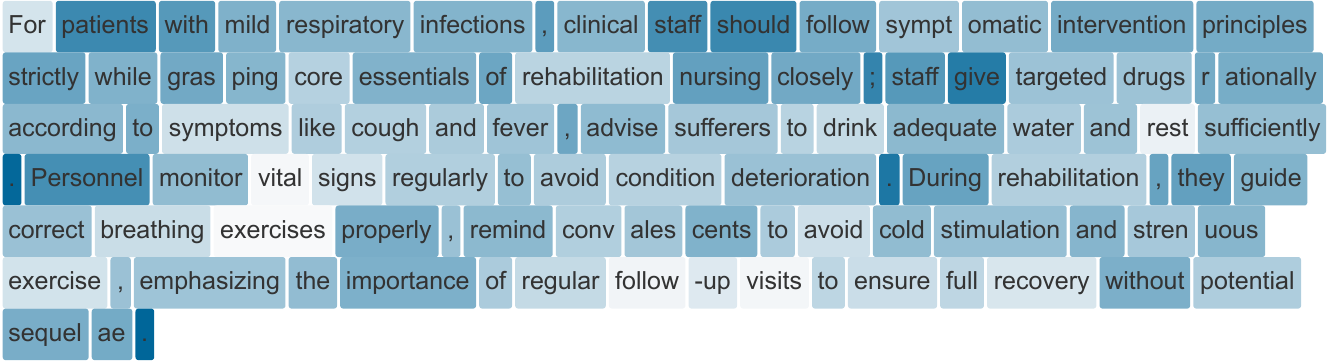}
}
\hfill
\subfloat[Flipped Classroom\label{appfig:sparsity_flipped_classroom}]{
    \includegraphics[width=0.3\linewidth]{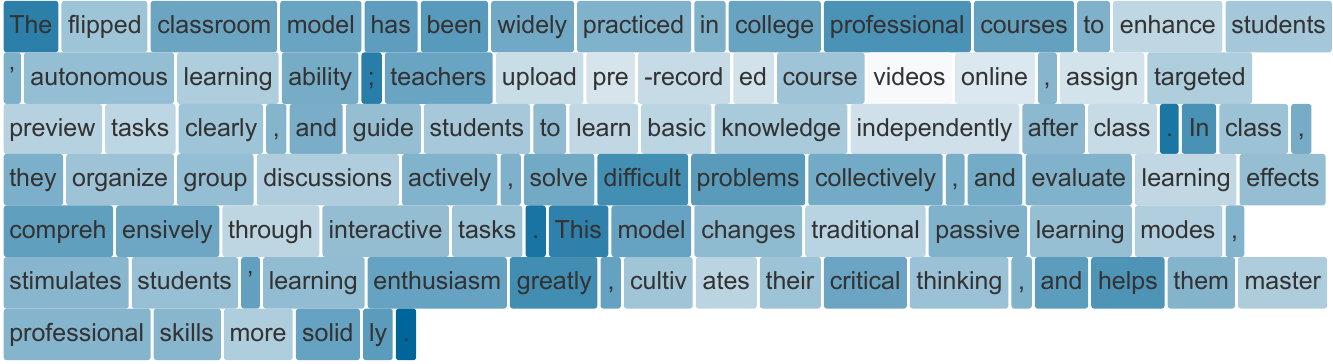}
}

\vspace{1em}

\subfloat[Urban Green Spaces\label{appfig:sparsity_urban_green_spaces}]{
    \includegraphics[width=0.3\linewidth]{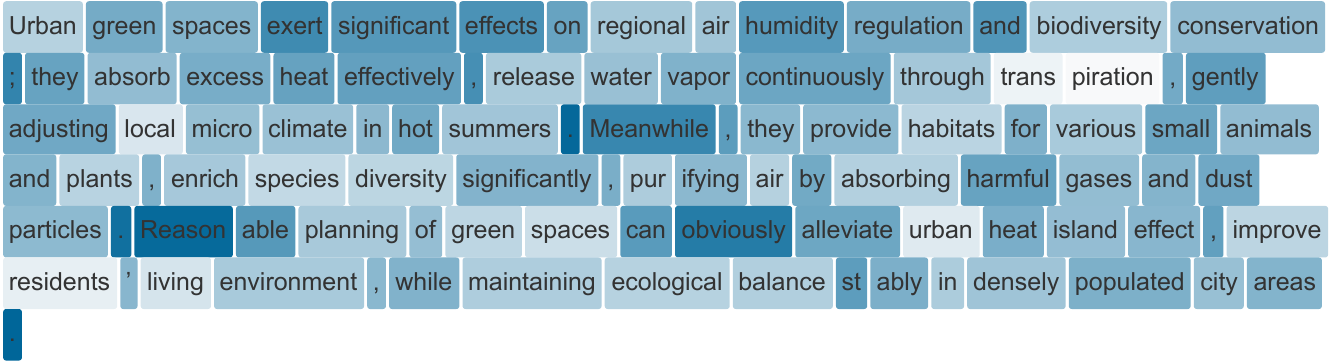}
}
\hfill
\subfloat[Intelligent Sensors\label{appfig:sparsity_intelligent_sensors}]{
    \includegraphics[width=0.3\linewidth]{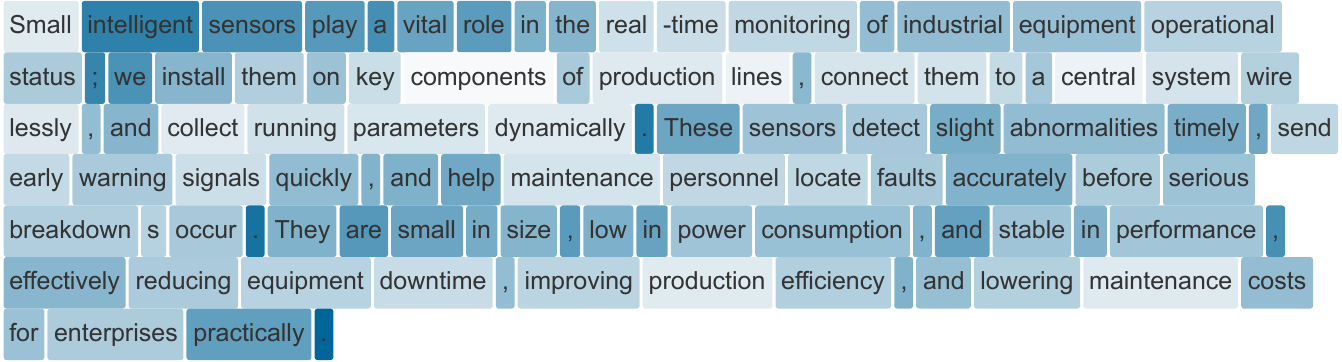}
}
\hfill
\subfloat[Medical Imaging\label{appfig:sparsity_medical_imaging}]{
    \includegraphics[width=0.3\linewidth]{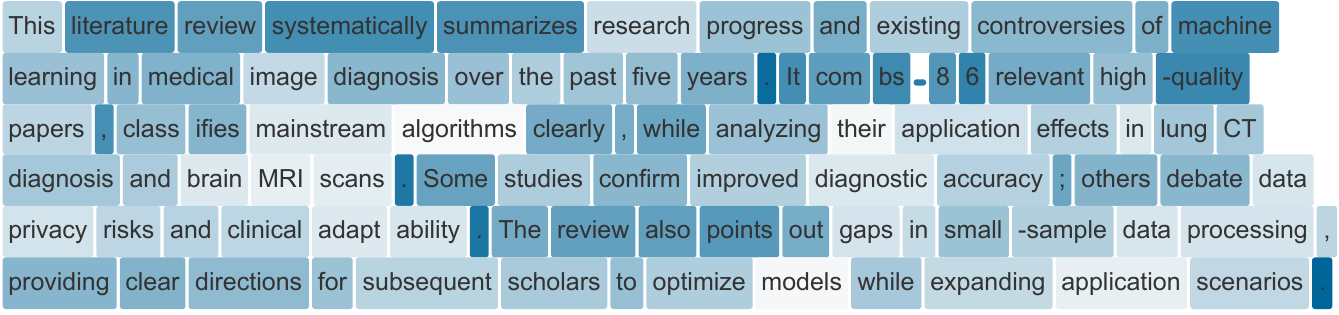}
}

\caption{Token-level Sparsity visualization under the proposed BaRA method.
Each subfigure corresponds to one input text from a different semantic domain.
The background intensity of each token represents its activation sparsity:
lighter colors indicate higher sparsity, while darker colors indicate denser
activations.}
\label{appfig:sparsity_all}
\end{figure*}

Figure~\ref{appfig:sparsity_all} shows that BaRA produces non-uniform sparsity
patterns across tokens and input domains. Tokens carrying domain-specific
semantics or key contextual information tend to preserve denser activations,
whereas less informative tokens are more likely to be sparsified. This
observation is consistent with the design motivation of BaRA: instead of using
a fixed low-rank adaptation pathway for all tokens, BaRA learns to allocate
adaptation capacity in an input-dependent manner.

These token-level visualizations provide qualitative evidence that BaRA can
perform context-aware sparsification. Together with the layer-wise results, they
suggest that BaRA improves adaptation efficiency through a dual mechanism:
heterogeneous sparsity allocation across layers and selective activation
suppression across tokens.

\section{Conclusion}
This paper addresses a fundamental limitation of existing LoRA-based parameter-efficient fine-tuning methods, namely their reliance on fixed and context-agnostic adaptation ranks. 
Specifically, we proposed \emph{Bayesian Adaptive Rank Allocation (BaRA)}, a sparsity-aware Bayesian fine-tuning framework that performs automatic and context-dependent rank allocation through structured global--local latent variables.
By formulating adaptation capacity as a sparse latent structure, BaRA enables the effective rank to vary across inputs, thereby allocating adaptation resources according to instance-specific complexity.
This mechanism allows the model to suppress redundant adaptation dimensions while preserving sufficient flexibility for informative or challenging inputs. As a result, BaRA achieves improved robustness, better uncertainty calibration, and more favorable test-time scaling behavior compared with fixed-rank adaptation methods.
Beyond the modeling contribution, we provided a complexity-theoretic analysis to explain the generalization advantage of BaRA. 
Overall, BaRA establishes a principled connection between probabilistic latent factor modeling and parameter-efficient fine-tuning.
The proposed framework suggests that sparsity-aware Bayesian inference can serve as an effective foundation for scalable, robust, and uncertainty-aware adaptation of large language models.

\bibliographystyle{IEEEtran}
\bibliography{ieee}

@inproceedings{DBLP:journals/corr/KingmaB14,
  author       = {Diederik P. Kingma and
                  Jimmy Ba},
  editor       = {Yoshua Bengio and
                  Yann LeCun},
  title        = {Adam: {A} Method for Stochastic Optimization},
  booktitle    = {3rd International Conference on Learning Representations, {ICLR} 2015,
                  San Diego, CA, USA, May 7-9, 2015, Conference Track Proceedings},
  year         = {2015},
  url          = {http://arxiv.org/abs/1412.6980},
  bibsource    = {dblp computer science bibliography, https://dblp.org}
}

@article{brown2020language,
  title={Language models are few-shot learners},
  author={Brown, Tom B},
  journal={arXiv preprint arXiv:2005.14165},
  year={2020}
}

@article{blei2003latent,
  title={Latent dirichlet allocation},
  author={Blei, David M and Ng, Andrew Y and Jordan, Michael I},
  journal={Journal of machine Learning research},
  volume={3},
  number={Jan},
  pages={993--1022},
  year={2003}
}

@inproceedings{zhou2012beta,
  title={Beta-negative binomial process and Poisson factor analysis},
  author={Zhou, Mingyuan and Hannah, Lauren and Dunson, David and Carin, Lawrence},
  booktitle={Artificial Intelligence and Statistics},
  pages={1462--1471},
  year={2012},
  organization={PMLR}
}

@article{xiong2023can,
  title={Can llms express their uncertainty? an empirical evaluation of confidence elicitation in llms},
  author={Xiong, Miao and Hu, Zhiyuan and Lu, Xinyang and Li, Yifei and Fu, Jie and He, Junxian and Hooi, Bryan},
  journal={arXiv preprint arXiv:2306.13063},
  year={2023}
}

@article{wang2023lora,
  title={LoRA ensembles for large language model fine-tuning},
  author={Wang, Xi and Aitchison, Laurence and Rudolph, Maja},
  journal={arXiv preprint arXiv:2310.00035},
  year={2023}
}

@article{kendall2017uncertainties,
  title={What uncertainties do we need in bayesian deep learning for computer vision?},
  author={Kendall, Alex and Gal, Yarin},
  journal={Advances in neural information processing systems},
  volume={30},
  year={2017}
}

@article{wang2024blob,
  title={Blob: Bayesian low-rank adaptation by backpropagation for large language models},
  author={Wang, Yibin and Shi, Haizhou and Han, Ligong and Metaxas, Dimitris and Wang, Hao},
  journal={Advances in Neural Information Processing Systems},
  volume={37},
  pages={67758--67794},
  year={2024}
}

@article{kim2024knowledge,
  title={Knowledge entropy decay during language model pretraining hinders new knowledge acquisition},
  author={Kim, Jiyeon and Lee, Hyunji and Cho, Hyowon and Jang, Joel and Hwang, Hyeonbin and Won, Seungpil and Ahn, Youbin and Lee, Dohaeng and Seo, Minjoon},
  journal={arXiv preprint arXiv:2410.01380},
  year={2024}
}

@article{ruiz2016generalized,
  title={The generalized reparameterization gradient},
  author={Ruiz, Francisco R and AUEB, Titsias RC and Blei, David and others},
  journal={Advances in neural information processing systems},
  volume={29},
  year={2016}
}

@inproceedings{naesseth2017reparameterization,
  title={Reparameterization gradients through acceptance-rejection sampling algorithms},
  author={Naesseth, Christian and Ruiz, Francisco and Linderman, Scott and Blei, David},
  booktitle={Artificial Intelligence and Statistics},
  pages={489--498},
  year={2017},
  organization={PMLR}
}

@inproceedings{lambert2025rewardbench,
  title={Rewardbench: Evaluating reward models for language modeling},
  author={Lambert, Nathan and Pyatkin, Valentina and Morrison, Jacob and Miranda, Lester James Validad and Lin, Bill Yuchen and Chandu, Khyathi and Dziri, Nouha and Kumar, Sachin and Zick, Tom and Choi, Yejin and others},
  booktitle={Findings of the Association for Computational Linguistics: NAACL 2025},
  pages={1755--1797},
  year={2025}
}

@article{chen2021evaluating,
  title={Evaluating large language models trained on code},
  author={Chen, Mark},
  journal={arXiv preprint arXiv:2107.03374},
  year={2021}
}

@misc{li2023alpacaeval,
  title={Alpacaeval: An automatic evaluator of instruction-following models},
  author={Li, Xuechen and Zhang, Tianyi and Dubois, Yann and Taori, Rohan and Gulrajani, Ishaan and Guestrin, Carlos and Liang, Percy and Hashimoto, Tatsunori B},
  year={2023}
}

@article{kumar2022fine,
  title={Fine-tuning can distort pretrained features and underperform out-of-distribution},
  author={Kumar, Ananya and Raghunathan, Aditi and Jones, Robbie and Ma, Tengyu and Liang, Percy},
  journal={arXiv preprint arXiv:2202.10054},
  year={2022}
}

@inproceedings{son2025not,
  title={Not All Adapters Matter: Selective Adapter Freezing for Memory-Efficient Fine-Tuning of Language Models},
  author={Son, Hyegang and Son, Yonglak and Kim, Changhoon and Kim, Young Geun},
  booktitle={Proceedings of the 2025 Conference of the Nations of the Americas Chapter of the Association for Computational Linguistics: Human Language Technologies (Volume 1: Long Papers)},
  pages={9479--9496},
  year={2025}
}

@article{kingma2013auto,
  title={Auto-encoding variational bayes},
  author={Kingma, Diederik P and Welling, Max},
  journal={arXiv preprint arXiv:1312.6114},
  year={2013}
}

@inproceedings{liu2023deja,
  title={Deja vu: Contextual sparsity for efficient llms at inference time},
  author={Liu, Zichang and Wang, Jue and Dao, Tri and Zhou, Tianyi and Yuan, Binhang and Song, Zhao and Shrivastava, Anshumali and Zhang, Ce and Tian, Yuandong and Re, Christopher and others},
  booktitle={International Conference on Machine Learning},
  pages={22137--22176},
  year={2023},
  organization={PMLR}
}

@article{yosinski2014transferable,
  title={How transferable are features in deep neural networks?},
  author={Yosinski, Jason and Clune, Jeff and Bengio, Yoshua and Lipson, Hod},
  journal={Advances in neural information processing systems},
  volume={27},
  year={2014}
}

@article{pan2024lisa,
  title={Lisa: Layerwise importance sampling for memory-efficient large language model fine-tuning},
  author={Pan, Rui and Liu, Xiang and Diao, Shizhe and Pi, Renjie and Zhang, Jipeng and Han, Chi and Zhang, Tong},
  journal={Advances in Neural Information Processing Systems},
  volume={37},
  pages={57018--57049},
  year={2024}
}

@article{wipf2011latent,
  title={Latent variable Bayesian models for promoting sparsity},
  author={Wipf, David P and Rao, Bhaskar D and Nagarajan, Srikantan},
  journal={IEEE Transactions on Information Theory},
  volume={57},
  number={9},
  pages={6236--6255},
  year={2011},
  publisher={IEEE}
}

@article{wipf2004sparse,
  title={Sparse Bayesian learning for basis selection},
  author={Wipf, David P and Rao, Bhaskar D},
  journal={IEEE Transactions on Signal processing},
  volume={52},
  number={8},
  pages={2153--2164},
  year={2004},
  publisher={IEEE}
}

@article{kotha2023understanding,
  title={Understanding catastrophic forgetting in language models via implicit inference},
  author={Kotha, Suhas and Springer, Jacob Mitchell and Raghunathan, Aditi},
  journal={arXiv preprint arXiv:2309.10105},
  year={2023}
}

@inproceedings{wang2023ensemble,
  title={Ensemble of low-rank adapters for large language model fine-tuning},
  author={Wang, Xi and Aitchison, Laurence and Rudolph, Maja},
  booktitle={NeurIPS Workshop on Efficient Natural Language and Speech Processing},
  year={2023}
}

@article{lakshminarayanan2017simple,
  title={Simple and scalable predictive uncertainty estimation using deep ensembles},
  author={Lakshminarayanan, Balaji and Pritzel, Alexander and Blundell, Charles},
  journal={Advances in neural information processing systems},
  volume={30},
  year={2017}
}

@article{yin2023large,
  title={Do large language models know what they don't know?},
  author={Yin, Zhangyue and Sun, Qiushi and Guo, Qipeng and Wu, Jiawen and Qiu, Xipeng and Huang, Xuanjing},
  journal={arXiv preprint arXiv:2305.18153},
  year={2023}
}

@inproceedings{liu2025uncertainty,
  title={Uncertainty quantification and confidence calibration in large language models: A survey},
  author={Liu, Xiaoou and Chen, Tiejin and Da, Longchao and Chen, Chacha and Lin, Zhen and Wei, Hua},
  booktitle={Proceedings of the 31st ACM SIGKDD Conference on Knowledge Discovery and Data Mining V. 2},
  pages={6107--6117},
  year={2025}
}

@inproceedings{guo2017calibration,
  title={On calibration of modern neural networks},
  author={Guo, Chuan and Pleiss, Geoff and Sun, Yu and Weinberger, Kilian Q},
  booktitle={International conference on machine learning},
  pages={1321--1330},
  year={2017},
  organization={PMLR}
}

@inproceedings{aghajanyan2021intrinsic,
  title={Intrinsic dimensionality explains the effectiveness of language model fine-tuning},
  author={Aghajanyan, Armen and Gupta, Sonal and Zettlemoyer, Luke},
  booktitle={Proceedings of the 59th annual meeting of the association for computational linguistics and the 11th international joint conference on natural language processing (volume 1: long papers)},
  pages={7319--7328},
  year={2021}
}

@article{li2018measuring,
  title={Measuring the intrinsic dimension of objective landscapes},
  author={Li, Chunyuan and Farkhoor, Heerad and Liu, Rosanne and Yosinski, Jason},
  journal={arXiv preprint arXiv:1804.08838},
  year={2018}
}

@article{kaplan2020scaling,
  title={Scaling laws for neural language models},
  author={Kaplan, Jared and McCandlish, Sam and Henighan, Tom and Brown, Tom B and Chess, Benjamin and Child, Rewon and Gray, Scott and Radford, Alec and Wu, Jeffrey and Amodei, Dario},
  journal={arXiv preprint arXiv:2001.08361},
  year={2020}
}

@article{li2024preserving,
  title={Preserving diversity in supervised fine-tuning of large language models},
  author={Li, Ziniu and Chen, Congliang and Xu, Tian and Qin, Zeyu and Xiao, Jiancong and Luo, Zhi-Quan and Sun, Ruoyu},
  journal={arXiv preprint arXiv:2408.16673},
  year={2024}
}

@article{cui2023ultrafeedback,
  title={Ultrafeedback: Boosting language models with scaled ai feedback},
  author={Cui, Ganqu and Yuan, Lifan and Ding, Ning and Yao, Guanming and He, Bingxiang and Zhu, Wei and Ni, Yuan and Xie, Guotong and Xie, Ruobing and Lin, Yankai and others},
  journal={arXiv preprint arXiv:2310.01377},
  year={2023}
}

@article{wang2016towards,
  title={Towards Bayesian deep learning: A framework and some existing methods},
  author={Wang, Hao and Yeung, Dit-Yan},
  journal={IEEE Transactions on Knowledge and Data Engineering},
  volume={28},
  number={12},
  pages={3395--3408},
  year={2016},
  publisher={IEEE}
}

@article{kumar2025latent,
  title={Latent Space Factorization in LoRA},
  author={Kumar, Shashi and Kaloga, Yacouba and Mitros, John and Motlicek, Petr and Kodrasi, Ina},
  journal={arXiv preprint arXiv:2510.19640},
  year={2025}
}

@article{han2024parameter,
  title={Parameter-efficient fine-tuning for large models: A comprehensive survey},
  author={Han, Zeyu and Gao, Chao and Liu, Jinyang and Zhang, Jeff and Zhang, Sai Qian},
  journal={arXiv preprint arXiv:2403.14608},
  year={2024}
}

@article{zhang2023lora,
  title={Lora-fa: Memory-efficient low-rank adaptation for large language models fine-tuning},
  author={Zhang, Longteng and Zhang, Lin and Shi, Shaohuai and Chu, Xiaowen and Li, Bo},
  journal={arXiv preprint arXiv:2308.03303},
  year={2023}
}

@inproceedings{he2022sparseadapter,
  title={Sparseadapter: An easy approach for improving the parameter-efficiency of adapters},
  author={He, Shwai and Ding, Liang and Dong, Daize and Zhang, Jeremy and Tao, Dacheng},
  booktitle={Findings of the Association for Computational Linguistics: EMNLP 2022},
  pages={2184--2190},
  year={2022}
}

@article{kopiczko2023vera,
  title={Vera: Vector-based random matrix adaptation},
  author={Kopiczko, Dawid J and Blankevoort, Tijmen and Asano, Yuki M},
  journal={arXiv preprint arXiv:2310.11454},
  year={2023}
}

@inproceedings{valipour2023dylora,
  title={DyLoRA: Parameter-efficient tuning of pre-trained models using dynamic search-free low-rank adaptation},
  author={Valipour, Mojtaba and Rezagholizadeh, Mehdi and Kobyzev, Ivan and Ghodsi, Ali},
  booktitle={Proceedings of the 17th Conference of the European Chapter of the Association for Computational Linguistics},
  pages={3274--3287},
  year={2023}
}

@article{lialin2023relora,
  title={Relora: High-rank training through low-rank updates},
  author={Lialin, Vladislav and Shivagunde, Namrata and Muckatira, Sherin and Rumshisky, Anna},
  journal={arXiv preprint arXiv:2307.05695},
  year={2023}
}

@article{rahmati2025c,
  title={C-LoRA: Contextual Low-Rank Adaptation for Uncertainty Estimation in Large Language Models},
  author={Rahmati, Amir Hossein and Jantre, Sanket and Zhang, Weifeng and Wang, Yucheng and Yoon, Byung-Jun and Urban, Nathan M and Qian, Xiaoning},
  journal={arXiv preprint arXiv:2505.17773},
  year={2025}
}

@article{onal2024gaussian,
  title={Gaussian stochastic weight averaging for Bayesian low-rank adaptation of large language models},
  author={Onal, Emre and Fl{\"o}ge, Klemens and Caldwell, Emma and Sheverdin, Arsen and Fortuin, Vincent},
  journal={arXiv preprint arXiv:2405.03425},
  year={2024}
}

@inproceedings{gal2016dropout,
  title={Dropout as a bayesian approximation: Representing model uncertainty in deep learning},
  author={Gal, Yarin and Ghahramani, Zoubin},
  booktitle={international conference on machine learning},
  pages={1050--1059},
  year={2016},
  organization={PMLR}
}

@inproceedings{NEURIPS2024_7d535754,
 author = {Wang, Yibin and Shi, Haizhou and Han, Ligong and Metaxas, Dimitris and Wang, Hao},
 booktitle = {Advances in Neural Information Processing Systems},
 doi = {10.52202/079017-2164},
 editor = {A. Globerson and L. Mackey and D. Belgrave and A. Fan and U. Paquet and J. Tomczak and C. Zhang},
 pages = {67758--67794},
 publisher = {Curran Associates, Inc.},
 title = {BLoB: Bayesian Low-Rank Adaptation by Backpropagation for Large Language Models},
 volume = {37},
 year = {2024}
}

@article{zhang2024autolora,
  title={Autolora: Automatically tuning matrix ranks in low-rank adaptation based on meta learning},
  author={Zhang, Ruiyi and Qiang, Rushi and Somayajula, Sai Ashish and Xie, Pengtao},
  journal={arXiv preprint arXiv:2403.09113},
  year={2024}
}

@inproceedings{wang2024roselora,
  title={Roselora: Row and column-wise sparse low-rank adaptation of pre-trained language model for knowledge editing and fine-tuning},
  author={Wang, Haoyu and Liu, Tianci and Li, Ruirui and Cheng, Monica Xiao and Zhao, Tuo and Gao, Jing},
  booktitle={Proceedings of the 2024 Conference on Empirical Methods in Natural Language Processing},
  pages={996--1008},
  year={2024}
}

@article{ding2023sparse,
  title={Sparse low-rank adaptation of pre-trained language models},
  author={Ding, Ning and Lv, Xingtai and Wang, Qiaosen and Chen, Yulin and Zhou, Bowen and Liu, Zhiyuan and Sun, Maosong},
  journal={arXiv preprint arXiv:2311.11696},
  year={2023}
}

@article{zhang2023adalora,
  title={Adalora: Adaptive budget allocation for parameter-efficient fine-tuning},
  author={Zhang, Qingru and Chen, Minshuo and Bukharin, Alexander and Karampatziakis, Nikos and He, Pengcheng and Cheng, Yu and Chen, Weizhu and Zhao, Tuo},
  journal={arXiv preprint arXiv:2303.10512},
  year={2023}
}

@article{zhang2020deep,
  title={Deep autoencoding topic model with scalable hybrid Bayesian inference},
  author={Zhang, Hao and Chen, Bo and Cong, Yulai and Guo, Dandan and Liu, Hongwei and Zhou, Mingyuan},
  journal={IEEE Transactions on Pattern Analysis and Machine Intelligence},
  volume={43},
  number={12},
  pages={4306--4322},
  year={2020},
  publisher={IEEE}
}

@article{yang2024bayesian,
  title={Bayesian reward models for LLM alignment},
  author={Yang, Adam X and Robeyns, Maxime and Coste, Thomas and Shi, Zhengyan and Wang, Jun and Bou-Ammar, Haitham and Aitchison, Laurence},
  journal={arXiv preprint arXiv:2402.13210},
  year={2024}
}

@article{hendrycks2021measuring,
  title={Measuring Massive Multitask Language Understanding}, 
  author={Dan Hendrycks and Collin Burns and Steven Basart and Andy Zou and Mantas Mazeika and Dawn Song and Jacob Steinhardt},
  journal={arXiv preprint arXiv:2009.03300},
  year={2021}
}

@inproceedings{houlsby2019parameter,
  title={Parameter-efficient transfer learning for NLP},
  author={Houlsby, Neil and Giurgiu, Andrei and Jastrzebski, Stanislaw and Morrone, Bruna and De Laroussilhe, Quentin and Gesmundo, Andrea and Attariyan, Mona and Gelly, Sylvain},
  booktitle={International conference on machine learning},
  pages={2790--2799},
  year={2019},
  organization={PMLR}
}

@article{hu2022lora,
  title={Lora: Low-rank adaptation of large language models.},
  author={Hu, Edward J and Shen, Yelong and Wallis, Phillip and Allen-Zhu, Zeyuan and Li, Yuanzhi and Wang, Shean and Wang, Lu and Chen, Weizhu and others},
  journal={ICLR},
  volume={1},
  number={2},
  pages={3},
  year={2022}
}

@article{samplawski2025scalable,
  title={Scalable Bayesian Low-Rank Adaptation of Large Language Models via Stochastic Variational Subspace Inference},
  author={Samplawski, Colin and Cobb, Adam D and Acharya, Manoj and Kaur, Ramneet and Jha, Susmit},
  journal={arXiv preprint arXiv:2506.21408},
  year={2025}
}

@article{qwen2,
  title={Qwen2 technical report},
  author={Yang, An and Yang, Baosong and Hui, Binyuan and Zheng, Bo and Yu, Bowen and Zhou, Chang and Li, Chengpeng and Li, Chengyuan and Liu, Dayiheng and Huang, Fei and others},
  journal={arXiv preprint arXiv:2407.10671},
  year={2024}
}

@article{yang2023bayesian,
  title={Bayesian low-rank adaptation for large language models},
  author={Yang, Adam X and Robeyns, Maxime and Wang, Xi and Aitchison, Laurence},
  journal={arXiv preprint arXiv:2308.13111},
  year={2023}
}

@article{sakaguchi2021winogrande,
  title={Winogrande: An adversarial winograd schema challenge at scale},
  author={Sakaguchi, Keisuke and Bras, Ronan Le and Bhagavatula, Chandra and Choi, Yejin},
  journal={Communications of the ACM},
  volume={64},
  number={9},
  pages={99--106},
  year={2021},
  publisher={ACM New York, NY, USA}
}

@article{clark2018think,
  title={Think you have solved question answering? try arc, the ai2 reasoning challenge},
  author={Clark, Peter and Cowhey, Isaac and Etzioni, Oren and Khot, Tushar and Sabharwal, Ashish and Schoenick, Carissa and Tafjord, Oyvind},
  journal={arXiv preprint arXiv:1803.05457},
  year={2018}
}

@misc{mihaylov1809can,
      title={Can a Suit of Armor Conduct Electricity? A New Dataset for Open Book Question Answering}, 
      author={Todor Mihaylov and Peter Clark and Tushar Khot and Ashish Sabharwal},
      year={2018},
      eprint={1809.02789},
      archivePrefix={arXiv},
      primaryClass={cs.CL},
      url={https://arxiv.org/abs/1809.02789}, 
}

@article{clark2019boolq,
  title={Boolq: Exploring the surprising difficulty of natural yes/no questions},
  author={Clark, Christopher and Lee, Kenton and Chang, Ming-Wei and Kwiatkowski, Tom and Collins, Michael and Toutanova, Kristina},
  journal={arXiv preprint arXiv:1905.10044},
  year={2019}
}

\section*{Appendix A: Proof of Complexity-Based Generalization Bound}

\appendices
\section{Proof of Complexity-Based Generalization Bound}
\label{app:proof_complexity_bound}

In this appendix, we provide the detailed proof of Theorem~\ref{thm:bara_generalization_main}. 
The proof is based on empirical Rademacher complexity and standard uniform convergence arguments.

\subsection{Empirical Rademacher Complexity}

Given a sample $\mathcal{D}=\{\xv_i\}_{i=1}^{n}$, the empirical Rademacher complexity of a real-valued hypothesis class $\mathcal{H}$ is defined as
\begin{equation}
\widehat{\mathfrak{R}}_{\mathcal{D}}(\mathcal{H})
=
\mathbb{E}_{\boldsymbol{\sigma}}
\left[
\sup_{h\in\mathcal{H}}
\frac{1}{n}
\sum_{i=1}^{n}
\sigma_i h(\xv_i)
\right],
\end{equation}
where $\sigma_1,\ldots,\sigma_n$ are independent Rademacher random variables taking values in $\{-1,+1\}$ with equal probability.

For BaRA, the hypothesis class is
\begin{equation}
\mathcal{H}_{\Phi,\theta}
=
\left\{
\xv\mapsto
B
\operatorname{diag}
\left(
\boldsymbol{\Phi}\odot\boldsymbol{\theta}(\xv)
\right)
A\xv
\right\}.
\end{equation}
For each input $\xv_i$, the prediction can be decomposed into rank-wise components:
\begin{equation}
h(\xv_i)
=
\sum_{k=1}^{r}
\Phi_k\theta_k(\xv_i)
\bv_k
\av_k^{\top}\xv_i,
\end{equation}
where $\av_k^{\top}$ is the $k$-th row of $A$ and $\bv_k$ is the $k$-th column of $B$.

Due to the joint sparse gate, only the rank components in 
$\mathcal{S}_{\Phi,\theta}(\xv_i)$ are effectively activated. 
Thus,
\begin{equation}
h(\xv_i)
=
\sum_{k\in \mathcal{S}_{\Phi,\theta}(\xv_i)}
\Phi_k\theta_k(\xv_i)
\bv_k
\av_k^{\top}\xv_i.
\end{equation}
Let
\begin{equation}
s_i
=
s_{\Phi,\theta}(\xv_i)
=
|\mathcal{S}_{\Phi,\theta}(\xv_i)|
\end{equation}
and
\begin{equation}
\bar{s}_{\Phi,\theta}
=
\frac{1}{n}
\sum_{i=1}^{n}
s_i.
\end{equation}

\subsection{Bounding the Contribution of Active Components}

Under Assumption~\ref{assump:bounded_lipschitz}, we have
\begin{equation}
\|\xv_i\|_2\le R_x,\quad
\|A\|_F\le R_A,\quad
\|B\|_F\le R_B,
\end{equation}
and
\begin{equation}
|\Phi_k|\le R_{\Phi},
\qquad
|\theta_k(\xv_i)|\le R_{\theta}.
\end{equation}
Therefore, for each active rank component,
\begin{equation}
\left|
\Phi_k\theta_k(\xv_i)
\bv_k
\av_k^{\top}\xv_i
\right|
\le
R_{\Phi}R_{\theta}
\|\bv_k\|_2
\|\av_k\|_2
\|\xv_i\|_2.
\end{equation}
Using the boundedness of $A$ and $B$, we further obtain
\begin{equation}
\left|
\Phi_k\theta_k(\xv_i)
\bv_k
\av_k^{\top}\xv_i
\right|
\le
R_{\Phi}R_{\theta}R_A R_B R_x.
\end{equation}
This shows that each active rank component has uniformly bounded contribution.

\subsection{Sparse Support Complexity}

For each input $\xv_i$, only $s_i$ out of $r$ rank components are active. 
The number of possible supports of size $s_i$ is bounded by
\begin{equation}
\binom{r}{s_i}
\le
\left(\frac{er}{s_i}\right)^{s_i}.
\end{equation}
This contributes a logarithmic complexity term of order
\begin{equation}
s_i\log\frac{er}{s_i}
\le
s_i\log r,
\end{equation}
where we use the fact that $s_i\le r$.

Averaging over all training samples yields the support complexity
\begin{equation}
\frac{1}{n}
\sum_{i=1}^{n}
s_i\log r
=
\bar{s}_{\Phi,\theta}\log r.
\end{equation}

Using standard Rademacher complexity bounds for sparse linear combinations, we obtain
\begin{equation}
\label{eq:rademacher_bound_app}
\widehat{\mathfrak{R}}_{\mathcal{D}}
(\mathcal{H}_{\Phi,\theta})
\le
C
R_A R_B R_{\Phi} R_{\theta} R_x
\sqrt{
\frac{
\bar{s}_{\Phi,\theta}\log r
}{n}
},
\end{equation}
where $C>0$ is an absolute constant.

\subsection{From Rademacher Complexity to Generalization Bound}

Let $\ell(h(\xv),y)$ be bounded by $M$ and $L$-Lipschitz with respect to $h(\xv)$. 
By the standard Rademacher generalization theorem, with probability at least $1-\delta$, the following holds uniformly for all $h\in\mathcal{H}_{\Phi,\theta}$:
\begin{equation}
\mathcal{R}(h)
\le
\widehat{\mathcal{R}}_{\mathcal{D}}(h)
+
2
\mathfrak{R}_{n}
(\ell\circ\mathcal{H}_{\Phi,\theta})
+
3M
\sqrt{
\frac{\log(2/\delta)}{2n}
}.
\end{equation}
By the Lipschitz contraction inequality,
\begin{equation}
\mathfrak{R}_{n}
(\ell\circ\mathcal{H}_{\Phi,\theta})
\le
L
\mathfrak{R}_{n}
(\mathcal{H}_{\Phi,\theta}).
\end{equation}
Substituting Eq.~\eqref{eq:rademacher_bound_app} gives
\begin{equation}
\begin{split}
\mathcal{R}(h)
\le
\widehat{\mathcal{R}}_{\mathcal{D}}(h)
+
&2LC
R_A R_B R_{\Phi} R_{\theta} R_x
\sqrt{
\frac{
\bar{s}_{\Phi,\theta}\log r
}{n}
}
\\
&+
3M
\sqrt{
\frac{\log(2/\delta)}{2n}
}.
\end{split}
\end{equation}
Absorbing universal constants into the big-$\mathcal{O}$ notation yields
\begin{equation}
\begin{split}
\mathcal{R}(h)
&\le
\widehat{\mathcal{R}}_{\mathcal{D}}(h)
+ \\
&\mathcal{O}
\left(
L R_A R_B R_{\Phi} R_{\theta} R_x
\sqrt{
\frac{
\bar{s}_{\Phi,\theta}\log r
}{n}
}
+
M
\sqrt{
\frac{
\log(1/\delta)
}{n}
}
\right).
\end{split}
\end{equation}
This proves Theorem~\ref{thm:bara_generalization_main}.

\subsection{Proof of Proposition~\ref{prop:joint_sparsity_main}}

We now prove the synergistic sparsity relation between the global and local latent variables. 
Recall that
\begin{equation}
\mathcal{S}_{\Phi}
=
\left\{
k:
|\Phi_k|>\tau_{\Phi}
\right\},
\end{equation}
and
\begin{equation}
\mathcal{S}_{\theta}(\xv)
=
\left\{
k:
|\theta_k(\xv)|>\tau_{\theta}
\right\}.
\end{equation}
The joint support is
\begin{equation}
\mathcal{S}_{\Phi,\theta}(\xv)
=
\left\{
k:
|\Phi_k\theta_k(\xv)|>\tau
\right\}.
\end{equation}
Choose $\tau=\tau_{\Phi}\tau_{\theta}$. 
If $k\notin \mathcal{S}_{\Phi}$, then $|\Phi_k|\le\tau_{\Phi}$. 
If $k\notin \mathcal{S}_{\theta}(\xv)$, then $|\theta_k(\xv)|\le\tau_{\theta}$. 
Therefore, whenever $k$ is not simultaneously active in both the global and local supports, the product gate cannot be strongly activated. 
Thus, the effective joint support satisfies
\begin{equation}
\mathcal{S}_{\Phi,\theta}(\xv)
\subseteq
\mathcal{S}_{\Phi}
\cap
\mathcal{S}_{\theta}(\xv),
\end{equation}
up to the threshold choice of the product gate. 
Taking cardinality gives
\begin{equation}
s_{\Phi,\theta}(\xv)
\le
\left|
\mathcal{S}_{\Phi}
\cap
\mathcal{S}_{\theta}(\xv)
\right|
\le
\min
\left\{
s_{\Phi},
s_{\theta}(\xv)
\right\}.
\end{equation}
Averaging over the training samples yields
\begin{equation}
\bar{s}_{\Phi,\theta}
=
\frac{1}{n}
\sum_{i=1}^{n}
s_{\Phi,\theta}(\xv_i)
\le
\frac{1}{n}
\sum_{i=1}^{n}
\min
\left\{
s_{\Phi},
s_{\theta}(\xv_i)
\right\}.
\end{equation}
Since $s_{\Phi}$ is independent of $\xv_i$, we have
\begin{equation}
\bar{s}_{\Phi,\theta}
\le
s_{\Phi}.
\end{equation}
Also,
\begin{equation}
s_{\Phi,\theta}(\xv_i)
\le
s_{\theta}(\xv_i),
\end{equation}
which gives
\begin{equation}
\bar{s}_{\Phi,\theta}
\le
\frac{1}{n}
\sum_{i=1}^{n}
s_{\theta}(\xv_i)
=
\bar{s}_{\theta}.
\end{equation}
Combining the two inequalities yields
\begin{equation}
\bar{s}_{\Phi,\theta}
\le
\min
\left\{
s_{\Phi},
\bar{s}_{\theta}
\right\}.
\end{equation}
This completes the proof.

\section{Biography Section}

\begin{IEEEbiographynophoto}{Zhibin Duan} received the Ph.D. degree in electronic engineering from Xidian University in  2024 and obtained the bachelor's degree from Xidian University in  2019.
He is currently an Assistant Professor at the School of Mathematics and Statistics, Xi'an Jiaotong University, since 2025.
His research interests include probabilistic machine learning, Bayesian deep learning, and topic modeling.
\end{IEEEbiographynophoto}

\begin{IEEEbiographynophoto}{Zongsheng Yue} (Member, IEEE) received his Ph.D.
degree from Xi’an Jiaotong University, Xi’an, China,
in 2021.
He was  a postdoctoral research
fellow with the College of Computing and Data
Science at Nanyang Technological University, from 2022 to 2025. From
September 2021 to March 2022, he was an associate
researcher in the Department of Computer Science
at Hong Kong University. He was a research assistant at the Department of Computing, Hong Kong
Polytechnic University, from October 2018 to June
2019, and at the Institute of Future Cities, The
Chinese University of Hong Kong, from February 2017 to September 2017,
respectively. 
He is currently an
 Professor with the School of Mathematics
and Statistics, Xi’an Jiaotong University.
His current research interests include noise modeling, image
restoration, and diffusion models.
\end{IEEEbiographynophoto}


\begin{IEEEbiographynophoto}{Bo Chen}
(Senior Member, IEEE) received the BS,
MS, and PhD degrees in electronic engineering from
Xidian University, Xi’an, China, in 2003, 2006, and
2008, respectively. He became a postdoctoral Fellow, a Research Scientist, and a Senior Research
Scientist with the Department of Electrical and Computer Engineering, Duke University, Durham, NC,
USA, from 2008 to 2012. From 2013, he has been a
Professor with the National Laboratory for Radar
Signal Processing, Xidian University. He was the
recipient of the Honorable Mention for 2010 National
Excellent Doctoral Dissertation Award and is selected into Overseas Talent
by Chinese Central Government, in 2014. His current research interests
include statistical machine learning, statistical signal processing, and radar
automatic target detection and recognition.
\end{IEEEbiographynophoto}

\begin{IEEEbiographynophoto}
    {Zongben Xu} received the PhD degree in mathematics from Xi’an Jiaotong University, Xi’an,
China, in 1987. He currently serves as the Academician of the Chinese Academy of Sciences,
the chief scientist of the National Basic Research
Program of China (973 Project), and the director
of the Institute for Information and System Sciences with Xi’an Jiaotong University. His current
research interests include nonlinear functional
analysis and intelligent information processing.
He was a recipient of the National Natural Science Award of China, in 2007, and the winner of the CSIAM Su Buchin
Applied Mathematics Prize, in 2008.
\end{IEEEbiographynophoto}

\vfill

\end{document}


\title{BaRA: Bayesian Adaptive Rank Allocation for Parameter-Efficient Fine-Tuning}

\author{
Zhibin~Duan, Yuhong~Wang, Jiahong~Fu,
Zongsheng~Yue\\ 
Bo~Chen,~\IEEEmembership{Senior Member,~IEEE,}
Zongben Xu,

\IEEEcompsocitemizethanks{\IEEEcompsocthanksitem Z. Duan, Y. Wang, B. Chen are with the National Key Lab of Radar Signal Processing, Xidian University, Xi'an, Shaanxi 710071, China；  
\protect 
E-mail: zbduan@xjtu.edu.cn; 
\IEEEcompsocthanksitem J.Fu, Z. Yue,  Z. Xu are with School of Mathematics and Statistics, Xi’an Jiaotong University, Xi’an, Shaanxi, 710049, China. (Corresponding author:  B. Chen)\protect
E-mail: bchen@mail.xidian.edu.cn;}
\thanks{This paper was produced by the IEEE Publication Technology Group. They are in Piscataway, NJ.}
\thanks{Manuscript received April 19, 2021; revised August 16, 2021.}}
\markboth{Journal of \LaTeX\ Class Files,~Vol.~14, No.~8, August~2021}%
{Shell \MakeLowercase{\textit{et al.}}: A Sample Article Using IEEEtran.cls for IEEE Journals}

\theoremstyle{plain}
\newtheorem{lemma}{Lemma}

\theoremstyle{definition}

\newcommand{\opnorm}[1]{\left\|#1\right\|_{op}}
\newcommand{\fnorm}[1]{\left\|#1\right\|_F}
\newcommand{\Rademacher}{\mathfrak{R}}
\newcommand{\Softplus}{\text{Softplus}}
\newcommand{\Softmax}{\text{Softmax}}

\begin{document}

\title{BaRA: Bayesian Adaptive Rank Allocation for Parameter-Efficient Fine-Tuning}

\author{
Zhibin~Duan, Yuhong~Wang, Jiahong~Fu,
Zongsheng~Yue\\ 
Bo~Chen,~\IEEEmembership{Senior Member,~IEEE,}
Zongben Xu,

\IEEEcompsocitemizethanks{\IEEEcompsocthanksitem Z. Duan, Y. Wang, B. Chen are with the National Key Lab of Radar Signal Processing, Xidian University, Xi'an, Shaanxi 710071, China;  (Corresponding author:  B. Chen)\protect
E-mail: bchen@mail.xidian.edu.cn;
\IEEEcompsocthanksitem J.Fu, Z. Yue, Z. Xu are with School of Mathematics and Statistics, Xi'an Jiaotong University, Xi'an, Shaanxi, 710049, China.} 
\thanks{This paper was produced by the IEEE Publication Technology Group. They are in Piscataway, NJ.}
\thanks{Manuscript received April 19, 2021; revised August 16, 2021.}}
\markboth{Journal of \LaTeX\ Class Files,~Vol.~14, No.~8, August~2021}%
{Shell \MakeLowercase{\textit{et al.}}: A Sample Article Using IEEEtran.cls for IEEE Journals}

\maketitle



\appendices
\section{Certified BaRA Generalization Bound}
\label{app:proof_complexity_bound}

This appendix gives the formal assumptions and proof for Theorem 1. The result is a deterministic Rademacher-complexity bound for a certified clipped and thresholded BaRA predictor. It is not a PAC-Bayesian posterior-predictive theorem for the raw soft Bayesian predictor.

\subsection{Certified Clipped and Thresholded Predictor}

For a scalar $z$, define
\begin{equation}
\operatorname{clip}_{R}(z)
=
\operatorname{sign}(z)\min\{|z|,R\},
\end{equation}
and apply this operator coordinatewise to vectors. Let $\boldsymbol{g}_{\rm BaRA}(\xv)\in\mathbb{R}^r$ be the gate vector induced by the variational global--local mechanism, and define the clipped induced gate
\begin{equation}
\bar{\boldsymbol{g}}_{\rm BaRA}(\xv)
=
\operatorname{clip}_{R_g}(\boldsymbol{g}_{\rm BaRA}(\xv)),
\end{equation}
so that $\|\bar{\boldsymbol{g}}_{\rm BaRA}(\xv)\|_{\infty}\le R_g$.
For $\tau>0$, define $T_{\tau}(z)=z\mathbf{1}\{|z|>\tau\}$ coordinatewise. Let $\operatorname{Top}_s$ keep the $s$ largest-magnitude coordinates of a vector and set the remaining coordinates to zero. The certified BaRA gate is
\begin{equation}
\boldsymbol{u}_{\tau,s}(\xv)
=
\operatorname{Top}_s
\left(
T_{\tau}(\bar{\boldsymbol{g}}_{\rm BaRA}(\xv))
\right),
\label{eq:certified_gate_app}
\end{equation}
and the corresponding predictor is
\begin{equation}
h_{\tau,s}(\xv)
=
h_0(\xv)
+
B\operatorname{diag}(\boldsymbol{u}_{\tau,s}(\xv))A\xv.
\label{eq:certified_predictor_app}
\end{equation}
By construction,
\begin{equation}
\|\boldsymbol{u}_{\tau,s}(\xv)\|_0\le s,
\qquad
\|\boldsymbol{u}_{\tau,s}(\xv)\|_{\infty}\le R_g.
\end{equation}
The empirical average active rank can be reported as a diagnostic, but the theorem uses the certified uniform active-rank budget above.

We assume
\begin{equation}
\|\xv\|_2\le R_x,
\qquad
\|A\|_F\le R_A,
\qquad
\|B\|_F\le R_B.
\label{eq:norm_bounds_app}
\end{equation}
The loss is assumed to be scalar, bounded, and Lipschitz:
\begin{equation}
0\le \ell(h(\xv),y)\le M,
\qquad
|\ell(u,y)-\ell(v,y)|\le L\|u-v\|_2.
\label{eq:loss_assumption_app}
\end{equation}
For classification with standard negative log-likelihood, the theorem applies directly to a bounded surrogate such as the clipped loss $\ell_M(u,y)=\min\{-\log p_u(y),M\}$, or under an explicit bounded-logit assumption. The proof below is stated for scalar predictions or fixed scalar margin/logit functionals of vector outputs; a fully vector-valued contraction can be used with the corresponding output-dimension constant.

Let $\mathcal{H}_{\tau,s}$ denote the class of certified predictors of the form in Eq.~\eqref{eq:certified_predictor_app} satisfying the norm bounds in Eq.~\eqref{eq:norm_bounds_app} and generated by the gate class in Eq.~\eqref{eq:certified_gate_app}.

For a sample $\mathcal{D}=\{(\xv_i,y_i)\}_{i=1}^{n}$, define the sample-wise gate-amplitude matrix
\begin{equation}
U_h(\mathcal{D})=(u_{ik})_{i,k},
\qquad
u_{ik}=u_{\tau,s,k}(\xv_i).
\end{equation}
Let $\mathcal{U}_s(\mathcal{D})$ be the set of all such matrices generated by the certified BaRA gate class. We assume that $\mathcal{U}_s(\mathcal{D})$ admits a proper empirical $\epsilon$-cover in entrywise sup-norm by matrices whose rows are $s$-sparse and whose entries are bounded by $R_g$, with
\begin{equation}
\log \mathcal{N}_{\infty}(\epsilon,\mathcal{U}_s(\mathcal{D}))
\le
s\log\frac{er}{s}
+
\Gamma_{\rm gate}(n,s,\epsilon),
\qquad
\epsilon\le \frac{R_g}{\sqrt n}.
\label{eq:gate_cover_main}
\end{equation}
The term $s\log(er/s)$ is the sparse rank-selection cost, while $\Gamma_{\rm gate}$ captures the remaining empirical log-cover complexity of the input-dependent gate-amplitude map. If the local gate is too flexible, $\Gamma_{\rm gate}$ may offset the benefit of small active rank.

\subsection{Fixed Gate-Amplitude Rademacher Bound}

Fix a sample and a gate-amplitude matrix $U=(u_{ik})\in\mathbb{R}^{n\times r}$ satisfying $|u_{ik}|\le R_g$ and $\|u_i\|_0\le s$ for every $i$. For a unit output direction $v$, consider the scalar projection class
\begin{equation}
f_{A,B,U,v}(\xv_i)
=
v^{\top}B\operatorname{diag}(u_i)A\xv_i,
\qquad
\|v\|_2\le1,
\end{equation}
with $\|A\|_F\le R_A$ and $\|B\|_F\le R_B$. Let $\mathcal{F}_U$ denote this class.

\begin{lemma}
\label{lem:fixed_gate_rademacher_app}
For every fixed $U$ satisfying the above bounds,
\begin{equation}
\widehat{\mathfrak{R}}_{\mathcal{D}}(\mathcal{F}_U)
\le
R_A R_B R_g R_x
\sqrt{\frac{s}{n}}.
\end{equation}
\end{lemma}

\begin{proof}
The empirical Rademacher complexity is
\begin{equation}
\widehat{\mathfrak{R}}_{\mathcal{D}}(\mathcal{F}_U)
=
\mathbb{E}_{\sigma}
\left[
\sup_{A,B,v}
\frac{1}{n}
\sum_{i=1}^{n}
\sigma_i v^{\top}B\operatorname{diag}(u_i)A\xv_i
\right].
\end{equation}
Let $\beta=B^{\top}v\in\mathbb{R}^{r}$. Since $\|v\|_2\le1$,
\begin{equation}
\|\beta\|_2
\le
\|B\|_F\|v\|_2
\le
R_B.
\end{equation}
Writing $a_k^{\top}$ for the $k$-th row of $A$,
\begin{equation}
v^{\top}B\operatorname{diag}(u_i)A\xv_i
=
\sum_{k=1}^{r}
\beta_k u_{ik}a_k^{\top}\xv_i.
\end{equation}
Define
\begin{equation}
z_k=
\frac{1}{n}
\sum_{i=1}^{n}
\sigma_i u_{ik}\xv_i.
\end{equation}
Then
\begin{equation}
\begin{split}
\sup_{A,\beta}
\sum_{k=1}^{r}\beta_k a_k^{\top}z_k
&\le
R_B
\left(\sum_{k=1}^{r}(a_k^{\top}z_k)^2\right)^{1/2}
\\
&\le
R_A R_B
\left(\sum_{k=1}^{r}\|z_k\|_2^2\right)^{1/2}.
\end{split}
\end{equation}
By Jensen's inequality and independence of the Rademacher signs,
\begin{equation}
\begin{split}
\mathbb{E}_{\sigma}
\left(\sum_{k=1}^{r}\|z_k\|_2^2\right)^{1/2}
&\le
\left(\sum_{k=1}^{r}\mathbb{E}_{\sigma}\|z_k\|_2^2\right)^{1/2}
\\
&=
\left(
\frac{1}{n^2}
\sum_{i=1}^{n}\sum_{k=1}^{r}u_{ik}^{2}\|\xv_i\|_2^2
\right)^{1/2}
\\
&\le
R_g R_x\sqrt{\frac{s}{n}}.
\end{split}
\end{equation}
This proves the lemma.
\end{proof}

\subsection{Gate Cover and Generalization Bound}

Let $N_{\epsilon}$ be the cover size in Eq.~\eqref{eq:gate_cover_main}. For any $U\in\mathcal{U}_s(\mathcal{D})$, choose a cover element $\widetilde U$ such that $\max_{i,k}|u_{ik}-\widetilde u_{ik}|\le\epsilon$. Then for every training input,
\begin{equation}
\left\|
B\operatorname{diag}(u_i-\widetilde u_i)A\xv_i
\right\|_2
\le
\epsilon R_A R_B R_x.
\end{equation}
Thus cover approximation changes the empirical Rademacher average by at most $\epsilon R_A R_B R_x$.

For each fixed cover element, Lemma~\ref{lem:fixed_gate_rademacher_app} gives a complexity bound of order $R_A R_B R_g R_x\sqrt{s/n}$. The Rademacher complexity of a finite union of $N_{\epsilon}$ uniformly bounded scalar classes satisfies
\begin{equation}
\widehat{\mathfrak{R}}_{\mathcal{D}}
\left(
\bigcup_{j=1}^{N_{\epsilon}}\mathcal{F}_{\widetilde U^{(j)}}
\right)
\le
\max_j \widehat{\mathfrak{R}}_{\mathcal{D}}(\mathcal{F}_{\widetilde U^{(j)}})
+
B_h\sqrt{\frac{2\log N_{\epsilon}}{n}},
\end{equation}
where $B_h\le R_A R_B R_g R_x$. Combining the fixed-gate term, the finite-union term, the cover approximation error, and Eq.~\eqref{eq:gate_cover_main}, and absorbing lower-order terms using $\epsilon\le R_g/\sqrt n$, yields
\begin{equation}
\widehat{\mathfrak{R}}_{\mathcal{D}}(\mathcal{H}_{\tau,s})
\le
C_1R_A R_B R_g R_x
\sqrt{
\frac{s\log(er/s)+\Gamma_{\rm gate}(n,s,\epsilon)}{n}
}.
\label{eq:certified_prediction_rad_app}
\end{equation}
By Lipschitz contraction for scalar losses or fixed scalar margin functionals,
\begin{IEEEeqnarray}{rCl}
\widehat{\mathfrak{R}}_{\mathcal{D}}
(\ell\circ\mathcal{H}_{\tau,s})
&\le&
C_2L R_A R_B R_g R_x
\nonumber\\
&&{}\times
\sqrt{
\frac{
s\log(er/s)+\Gamma_{\rm gate}(n,s,\epsilon)
}{n}
}.
\end{IEEEeqnarray}
Finally, the standard bounded-loss Rademacher generalization theorem gives, with probability at least $1-\delta$, uniformly over $h_{\tau,s}\in\mathcal{H}_{\tau,s}$,
\begin{IEEEeqnarray}{rCl}
\mathcal{R}(h_{\tau,s})
&\le&
\widehat{\mathcal{R}}_{\mathcal{D}}(h_{\tau,s})
\nonumber\\
&&{}+
C L R_A R_B R_g R_x
\nonumber\\
&&{}\times
\sqrt{
\frac{s\log(er/s)+\Gamma_{\rm gate}(n,s,\epsilon)}{n}
}
\nonumber\\
&&{}+
3M
\sqrt{
\frac{\log(2/\delta)}{2n}
}.
\end{IEEEeqnarray}
This gives Eq. 21 in the main text.

\subsection{Rank-Explicit LoRA Comparator}

For comparison, consider a rank-$r$ LoRA adapter
\begin{equation}
h_{\rm L}(\xv)
=
h_0(\xv)+B_{\rm L}A_{\rm L}\xv,
\end{equation}
with $\|A_{\rm L}\|_F\le R_A$ and $\|B_{\rm L}\|_F\le R_B$.
Applying the same scalar-projection proof to the fixed-gate case $u_{ik}=1$, $s=r$, and $R_g=1$ gives
\begin{equation}
\widehat{\mathfrak{R}}_{\mathcal{D}}(\mathcal{F}_{\rm LoRA})
\le
R_A R_B R_x\sqrt{\frac{r}{n}}.
\end{equation}
Consequently, with probability at least $1-\delta$,
\begin{equation}
\mathcal{R}(h_{\rm L})
\le
\widehat{\mathcal{R}}_{\mathcal{D}}(h_{\rm L})
+
C_L L R_A R_B R_x\sqrt{\frac{r}{n}}
+
3M\sqrt{\frac{\log(2/\delta)}{2n}}.
\label{eq:lora_comparator_bound_app}
\end{equation}
Comparing the leading terms in Eq.（21） and Eq.~\eqref{eq:lora_comparator_bound_app}, BaRA has a smaller rank-explicit upper bound whenever
\begin{equation}
R_g^2
\left[
s\log\frac{er}{s}
+
\Gamma_{\rm gate}(n,s,\epsilon)
\right]
<
r,
\end{equation}
up to universal constants. This is a sufficient comparison under the same rank-explicit decomposition and norm scale; it is not a claim that every possible LoRA bound must scale with $r$.

\subsection{Mildness of the Advantage Condition}

The comparison condition above should not be read as automatically true for arbitrary input-dependent gates. It is nevertheless mild in the sparse controlled-adaptivity regime targeted by BaRA. To see this, normalize the active rank and gate entropy by the LoRA rank:
\begin{equation}
\alpha=\frac{s}{r},
\qquad
\gamma=\frac{\Gamma_{\rm gate}(n,s,\epsilon)}{r}.
\end{equation}
The sufficient advantage condition becomes
\begin{equation}
R_g^2
\left[
\alpha\log\frac{e}{\alpha}
+
\gamma
\right]
<
1.
\label{eq:normalized_advantage_condition_app}
\end{equation}
Equivalently, the ratio between the leading BaRA and rank-explicit LoRA complexity terms satisfies, up to universal constants,
\begin{equation}
\frac{\mathrm{Rad}_{\rm BaRA}}{\mathrm{Rad}_{\rm LoRA}}
\lesssim
R_g
\sqrt{
\alpha\log\frac{e}{\alpha}
+
\gamma
}.
\label{eq:normalized_rad_ratio_app}
\end{equation}
Thus BaRA has a smaller leading complexity term whenever the active-rank fraction $\alpha$ is small, the clipped gate amplitude satisfies $R_g\le1$, and the normalized gate complexity $\gamma$ is below the available slack
\begin{equation}
\gamma
<
R_g^{-2}
-
\alpha\log\frac{e}{\alpha}.
\label{eq:gate_entropy_slack_app}
\end{equation}

In the purely global or zero-entropy case $\Gamma_{\rm gate}=0$, with normalized gates $R_g\le1$, the condition holds for every strictly sparse active rank $s<r$. Indeed, for $0<\alpha<1$, the function
\begin{equation}
f(\alpha)=\alpha\log\frac{e}{\alpha}
\end{equation}
satisfies
\begin{equation}
f'(\alpha)=\log\frac{1}{\alpha}>0,
\qquad
f(1)=1,
\end{equation}
and hence $f(\alpha)<1$ for every $0<\alpha<1$. Therefore
\begin{equation}
R_g^2s\log\frac{er}{s}
\le
r\alpha\log\frac{e}{\alpha}
<
r.
\end{equation}

When the local gate has nonzero entropy, the condition still permits a nontrivial gate-complexity budget. For example, if $R_g=1$, then
\begin{equation}
\frac{\Gamma_{\rm gate}(n,s,\epsilon)}{r}
<
1-\alpha\log\frac{e}{\alpha}.
\end{equation}
Thus $\Gamma_{\rm gate}$ need not vanish; it can be a constant fraction of $r$ when the active-rank fraction is small. This formalizes the intended regime of BaRA: sparse active rank plus controlled gate entropy. Conversely, if the local gate can assign essentially unrelated supports to different examples, $\Gamma_{\rm gate}$ may erase the sparsity gain.

\subsection{Relation to the Original Soft Gate}

The theorem is stated for the certified gate $\boldsymbol{u}_{\tau,s}$. For the bounded clipped soft-gate predictor
\begin{equation}
h_{\rm soft}(\xv)
=
h_0(\xv)
+
B\operatorname{diag}(\bar{\boldsymbol{g}}_{\rm BaRA}(\xv))A\xv,
\end{equation}
define the residual
\begin{equation}
\rho_{\tau,s}
=
\sup_{\xv}
\left\|
\bar{\boldsymbol{g}}_{\rm BaRA}(\xv)
-
\boldsymbol{u}_{\tau,s}(\xv)
\right\|_{\infty}.
\end{equation}
Then
\begin{equation}
\|h_{\rm soft}(\xv)-h_{\tau,s}(\xv)\|_2
\le
R_A R_B R_x\rho_{\tau,s}.
\end{equation}
If the loss is $L$-Lipschitz, the risk and empirical-risk differences each contribute at most $L R_A R_B R_x\rho_{\tau,s}$. Hence the certified bound transfers to the clipped soft predictor with the additional approximation term
\begin{equation}
2L R_A R_B R_x\rho_{\tau,s}.
\label{eq:soft_residual_app}
\end{equation}
For pure thresholding without a top-$s$ truncation removing larger coordinates, $\rho_{\tau,s}\le\tau$, recovering the simpler $2L\tau R_A R_B R_x$ term. With top-$s$ certification, the residual form in Eq.~\eqref{eq:soft_residual_app} is the appropriate safe statement.

\subsection{Random Gamma--Weibull Variables and Boundedness}

The Gamma prior and Weibull variational posterior have unbounded support, so the deterministic gate bounds are not automatic for raw samples. The theorem applies directly to clipped or projected gates as in Eq.~\eqref{eq:certified_gate_app}. Alternatively, if the required bounds hold on an event $\mathcal{E}_R$ with probability at least $1-\eta$, then the same conclusion holds with probability at least $1-\delta-\eta$ by conditioning on $\mathcal{E}_R$ and union bounding the failure events.

\subsection{Proof of Proposition 1}

We now prove the synergistic sparsity relation for a global--local decomposition of the induced gate. 
Recall that
\begin{equation}
\mathcal{S}_{\rm G}
=
\left\{
k:
|g_{{\rm G},k}|>\tau_{\rm G}
\right\},
\qquad
\mathcal{S}_{\rm L}(\xv)
=
\left\{
k:
|g_{{\rm L},k}(\xv)|>\tau_{\rm L}
\right\}.
\end{equation}
The joint thresholded support is
\begin{equation}
\mathcal{S}_{\tau}(\xv)
=
\left\{
k:
|g_{{\rm G},k}g_{{\rm L},k}(\xv)|>\tau
\right\}.
\end{equation}
Assume $|g_{{\rm G},k}|\le R_{\rm G}$ and $|g_{{\rm L},k}(\xv)|\le R_{\rm L}$, and choose
\begin{equation}
\tau\ge \max\{R_{\rm L}\tau_{\rm G},R_{\rm G}\tau_{\rm L}\}.
\end{equation}
If $k\notin \mathcal{S}_{\rm G}$, then $|g_{{\rm G},k}|\le\tau_{\rm G}$, and hence
\begin{equation}
|g_{{\rm G},k}g_{{\rm L},k}(\xv)|
\le
R_{\rm L}\tau_{\rm G}
\le
\tau.
\end{equation}
Similarly, if $k\notin \mathcal{S}_{\rm L}(\xv)$, then
\begin{equation}
|g_{{\rm G},k}g_{{\rm L},k}(\xv)|
\le
R_{\rm G}\tau_{\rm L}
\le
\tau.
\end{equation}
Therefore, whenever $|g_{{\rm G},k}g_{{\rm L},k}(\xv)|>\tau$, both $k\in\mathcal{S}_{\rm G}$ and $k\in\mathcal{S}_{\rm L}(\xv)$ must hold. 
Thus,
\begin{equation}
\mathcal{S}_{\tau}(\xv)
\subseteq
\mathcal{S}_{\rm G}
\cap
\mathcal{S}_{\rm L}(\xv).
\end{equation}
Taking cardinality gives
\begin{equation}
|\mathcal{S}_{\tau}(\xv)|
\le
\min
\left\{
s_{\rm G},
s_{\rm L}(\xv)
\right\}.
\end{equation}
Averaging over the training samples yields
\begin{equation}
\bar{s}_{\tau}
\le
\min
\left\{
s_{\rm G},
\bar{s}_{\rm L}
\right\},
\qquad
\bar{s}_{\rm L}
=
\frac{1}{n}\sum_{i=1}^{n}s_{\rm L}(\xv_i).
\end{equation}
This completes the proof.